\documentclass{article}

\usepackage[main, final]{neurips_2026}
\usepackage[table]{xcolor} 
\usepackage{ifthen}
\newboolean{showcomments}
\setboolean{showcomments}{true}
\usepackage{color}
\usepackage{todonotes}

\definecolor{bleudefrance}{rgb}{0.19, 0.55, 0.91}
\definecolor{ao(english)}{rgb}{0.0, 0.5, 0.0}

\newcommand{\addcite}[0]{\ifthenelse{\boolean{showcomments}}
{\textcolor{purple}{(add cite(s)) }}{}}%

\newcommand{\addref}[0]{\ifthenelse{\boolean{showcomments}}
{\textcolor{purple}{(add ref) }}{}}%

\newcommand{\myparagraph}[1]{\vspace{1mm}\noindent\textbf{#1.}}

\newcommand{\enrique}[1]{  \ifthenelse{\boolean{showcomments}}
{\todo[inline,color=bleudefrance]{Enrique: #1}}{}}
\newcommand{\rene}[1]{  \ifthenelse{\boolean{showcomments}}
{\todo[inline,color=cyan]{Ren\'e: #1}}{}}
\newcommand{\emmargin}[1]{\ifthenelse{\boolean{showcomments}}{\marginpar{\color{bleudefrance}\tiny EM: #1}}{}}
\newcommand{\hancheng}[1]{  \ifthenelse{\boolean{showcomments}}
{\todo[inline,color=orange]{Hancheng: #1}}{}}
\newcommand{\ziqing}[1]{  \ifthenelse{\boolean{showcomments}}
{\todo[inline,color=red]{Ziqing: #1}}{}}
\newcommand{\salma}[1]{  \ifthenelse{\boolean{showcomments}}
{\todo[inline,color=yellow]{Salma: #1}}{}}
\newcommand{\zxmargin}[1]{\ifthenelse{\boolean{showcomments}}{\marginpar{\color{purple}\tiny ZX: #1}}{}}
\newcommand{\stmargin}[1]{\ifthenelse{\boolean{showcomments}}{\marginpar{\color{red}\tiny ST: #1}}{}}

\newcommand{\hl}[1]{\ifthenelse{\boolean{showcomments}}
{\textcolor{red}{#1}}{#1}}

\newboolean{showedits}
\setboolean{showedits}{false}
\usepackage[markup=underlined]{changes}
\definechangesauthor[color=bleudefrance]{EM}
\newcommand{\aem}[1]{
\ifthenelse{\boolean{showedits}}
{\added[id=EM]{#1}}
{\!#1\hspace{-4.75pt}}
}
\newcommand{\repem}[2]{
\ifthenelse{\boolean{showedits}}
{\replaced[id=EM]{#1}{#2}}
{\!#1\hspace{-4.75pt}}
}
\newcommand{\dem}[1]{
\ifthenelse{\boolean{showedits}}
{\deleted[id=EM]{#1}}
{}
}

\usepackage[utf8]{inputenc} 
\usepackage[T1]{fontenc}    
\usepackage{hyperref}       
\usepackage{url}            
\usepackage{booktabs}       
\usepackage{nicefrac}       
\usepackage{microtype}      
\usepackage{enumitem}
\usepackage{float}
\usepackage{graphicx}
\usepackage{tabularx}
\usepackage{algorithm}
\usepackage{algpseudocode}

\usepackage{amsmath,amsfonts,bm}
\usepackage{amsthm}

\def\1{\bm{1}}

\DeclareMathAlphabet{\mathsfit}{\encodingdefault}{\sfdefault}{m}{sl}
\SetMathAlphabet{\mathsfit}{bold}{\encodingdefault}{\sfdefault}{bx}{n}

\newcommand{\R}{\mathbb{R}}

\DeclareMathOperator*{\argmax}{arg\,max}
\DeclareMathOperator*{\argmin}{arg\,min}

\DeclareMathOperator{\Tr}{Tr}

\newtheorem{thm}{Theorem}[section]
\newtheorem{lem}{Lemma}[section]
\newtheorem{cor}{Corollary}[section]
\newtheorem{prop}{Proposition}[section]

\newtheorem{asmp}{Assumption}[section]
\newtheorem{defn}{Definition}[section]

\newtheorem{rem}{Remark}[section]

\usepackage{amssymb}

\title{
A Margin Perspective on LoRA: Robustness to Catastrophic Forgetting and Adapter Merging (MaLoRA) 
}

\author{
  \begin{minipage}[t]{0.96\textwidth}
  \centering
  \normalfont
  \textbf{Ziqing Xu}$^{1,4}$\thanks{Correspondence: \texttt{ziqingxu@wharton.upenn.edu}} \hspace{1em}
  \textbf{Hancheng Min}$^{7}$ \hspace{1em}
  \textbf{Lachlan Ewen MacDonald}$^{3,4}$ \hspace{1em} \\
  \textbf{Salma Tarmoun}$^{2,4}$ \hspace{1em}
  \textbf{Enrique Mallada}$^{8}$ \hspace{1em}
  \textbf{Weijie Su}$^{1}$ \hspace{1em}
  \textbf{Ren\'e Vidal}$^{2,3,4,5,6}$ \\[0.35em]
  {\small
  $^{1}$Statistics and Data Science \quad
  $^{2}$AMCS \quad
  $^{3}$ESE \quad
  $^{4}$IDEAS \quad
  $^{5}$Radiology \quad
  $^{6}$CIS \\
  University of Pennsylvania \\
  $^{7}$INS \& SMS, Shanghai Jiao Tong University \quad
  $^{8}$ECE, Johns Hopkins University
  }
  \end{minipage}
}

\begin{document}

\maketitle

\begin{abstract}
Low-rank adaptation (LoRA) is the de facto method for parameter-efficient fine-tuning of large neural networks, achieving strong task adaptation with minimal trainable parameters and low optimization cost. Beyond this efficiency, practitioners have observed two striking properties: \textit{robustness to catastrophic forgetting} and the ability to \textit{merge independently trained adapters} into a single model that performs competitively across multiple tasks. These properties are central to continual and multi-task learning, yet remain poorly understood. In this work, we provide a theoretical explanation through a unified \textit{margin-based perspective}. We analyze LoRA in multiclass linear classification under a near-orthogonal task regime with $\ell_2$-regularization, and characterize the optimal LoRA adapter across regularization regimes. Our analysis shows that, in an intermediate regime of regularization parameters, the optimal adapter aligns with the \emph{max-margin solution} on the fine-tuning data. Building on this characterization, we derive two key consequences. First, we obtain closed-form expressions for the margins on pre-training and fine-tuning data, revealing a precise margin trade-off: the regularization parameter controls the balance between retention and adaptation. Second, we analyze adapter merging, proving that merged models achieve positive margin on each task and deriving optimal mixing coefficients that balance margins across tasks and maximize the margin over their union. These results lead to a simple, training-free merging rule, which we term \textit{margin-based LoRA merging} (MaLoRA). Experiments on modern architectures and real datasets validate our theoretical predictions, showing that MaLoRA matches or outperforms several adapter-merging baselines across a range of vision and language classification tasks.
\end{abstract}


\section{Introduction}

Foundation models have transformed artificial intelligence across numerous domains. However, deploying these models in real-world applications typically requires fine-tuning on specialized datasets to meet accuracy, safety, and alignment requirements. Traditional full fine-tuning, which updates all parameters of a pre-trained model, incurs substantial computational cost, including large memory footprints, long training times, and significant storage demands. Low-Rank Adaptation (LoRA)~\citep{hu2022lora} addresses this challenge by augmenting pre-trained weight matrices $W$ with low-rank adapters $B, A$ while keeping $W$ frozen. Empirically, LoRA matches the performance of full fine-tuning while training only a small fraction of the parameters.

While LoRA was originally introduced for its computational efficiency, practitioners have since identified two striking empirical properties.
\begin{enumerate}[leftmargin=*]
    \item \textit{Robustness to catastrophic forgetting.} In fine-tuning and continual learning, the degradation of performance on previously learned tasks after adapting to new ones (a.k.a. catastrophic forgetting) is a fundamental challenge~\citep{mccloskey1989catastrophic, kirkpatrick2017overcoming, ramasesh2021effect}. Surprisingly, recent studies show that LoRA exhibits strong robustness to catastrophic forgetting, even after substantial adaptation~\citep{wistuba2023continual, biderman2024lora, qiao2024learn}. This robustness makes LoRA particularly appealing for continual learning.
    
    \item \textit{Effective adapter merging.} In multi-task settings, maintaining separate fine-tuned models for each task is storage-intensive. Remarkably, LoRA adapters trained independently on different tasks can be merged (e.g., via simple weighted averaging) into a single adapter that performs well across all tasks~\citep{huang2023lorahub, yadav2023resolving, wang2024lora, yu2024language, zhao2024loraretriever}. This property enables flexible and efficient deployment.
\end{enumerate}

Despite these empirical benefits, the theoretical mechanisms underlying LoRA's \emph{robustness to catastrophic forgetting} and its \emph{adapter merging} capabilities remain poorly understood. In this work, we provide a unified theoretical explanation for both phenomena through a margin-based perspective.

\myparagraph{Paper contributions}
We study LoRA in multiclass linear classification under a near-orthogonal-task regime, in which fine-tuning samples are nearly orthogonal to one another and to the pre-training samples. In this setting, we characterize the global minimizer of the LoRA objective and use this characterization to analyze forgetting and adapter merging. Compared with prior theoretical work on LoRA (see Appendix~\ref{app:related_work}), our work makes the following contributions:
\begin{itemize}[leftmargin=*]
    \item \emph{Characterization of the optimal solution.}
    We characterize the global minimizer of LoRA across different regularization regimes (Theorem~\ref{thm:main}). In the high-penalty regime, the optimal adapter is zero. In the intermediate regime, the adapter aligns with the \emph{max-margin} direction on the fine-tuning data. In the low-penalty regime, the adapter aligns with a non-max-margin classifier whose direction generally does not admit a simple closed form. Nevertheless, we show that this direction converges to an interpretable limit as the regularization parameter tends to zero.

    \item \emph{Theory for reduced forgetting.}
    We define forgetting as the reduction in margin on the pre-training data after fine-tuning, and derive closed-form expressions for the normalized margins on the pre-training data, the fine-tuning data, and their union (Theorem~\ref{thm:lora_forget_less}). Our analysis shows that these margins are governed by the regularization parameter and satisfy a precise trade-off:  
    decreasing the regularization parameter increases the fine-tuning margin but also increases the Frobenius-norm ratio of the adapter to the pre-trained weights, leading to greater forgetting.
    We further identify the regularization parameter that maximizes the margin on the union of pre-training and fine-tuning tasks, thereby optimally balancing retention and adaptation.


    \item \emph{Theory for adapter merging.} Under the intermediate-regime assumption for each task, we derive closed-form expressions for the margin of the merged model on every task (Theorem~\ref{thm:lora_merge}). We show that the merged model attains strictly positive margin on each task, providing a theoretical explanation for the effectiveness of adapter merging. We also characterize, in closed form, the mixing coefficients that maximize the margin of the merged model on the union of all tasks, which naturally leads to \emph{margin-based LoRA merging} (MaLoRA), a theoretically motivated adapter-merging method. This contrasts with many existing model-merging approaches, which rely primarily on heuristic prescriptions for choosing how to combine task-specific models.

    \item \emph{Numerical validation.} Experiments on real datasets and modern architectures support the theory in three ways: \emph{(i)} in the linear probing setting, the empirically optimal regularization parameter agrees with the value predicted by our analysis; \emph{(ii)} in the same setting, the empirically optimal adapter mixing weights agree with our closed-form coefficients; and \emph{(iii)} in LoRA adapter merging, MaLoRA achieves performance comparable to, and in several cases better than, strong heuristic adapter-merging baselines across vision and language classification tasks.
\end{itemize}

\section{Preliminaries}\label{sec:setup}
In this section, we first introduce the multiclass linear classification problem and its corresponding hard-margin SVM formulation. We then describe the specific learning problem considered in this work, which consists of a pre-training and fine-tuning phase. Finally, we state the key assumptions that underpin our analysis, and discuss their motivation and implications.

\subsection{Background on multiclass linear classification}\label{sec:background}

We first review the standard $K$-class linear classification problem. Given a dataset $\mathcal{D}$, the goal is to learn a weight matrix $W \in \mathbb{R}^{K \times d}$ that minimizes the mean empirical cross-entropy loss:
\begin{align}\label{eqn:ce_obj}
    \textstyle\min_{W \in \mathbb{R}^{K \times d}} L(W; \mathcal{D}) := \frac{1}{|\mathcal{D}|}\sum_{(\boldsymbol{x}, \boldsymbol{y}) \in \mathcal{D}} \mathcal{L}_{\mathrm{CE}}(\boldsymbol{y}, W \boldsymbol{x}),
\end{align}
where $\boldsymbol{x} \in \mathbb{R}^d$ is the input and $\boldsymbol{y} \in \mathbb{R}^K$ is a one-hot label vector. The cross-entropy loss is given by:
\begin{align}
    \mathcal{L}_{\mathrm{CE}}(\boldsymbol{y}, \hat{\boldsymbol{y}}):= -\sum_{c=1}^K y_c \log\!\left(\frac{\exp(\hat{y}_c)}{\sum_{i=1}^K \exp(\hat{y}_i)}\right),\qquad \hat{\boldsymbol{y}} = W \boldsymbol{x}\,.
\end{align}
Recent studies have shown that when the data are linearly separable, gradient descent (GD) implicitly biases the solution of~\eqref{eqn:ce_obj} toward the max-margin classifier~\citep{soudry2018implicit, lyu2019gradient, ravi2024implicit}. Specifically, the iterates of GD converge in direction to the max-margin solution of the following hard-margin SVM problem\footnote{Whenever \eqref{eqn:svm} is feasible, its solution is unique since the objective $\|W\|_F^2$ is strongly convex and the feasible set is convex.}:
\begin{align}\label{eqn:svm}
    W^{\mathrm{SVM}}(\mathcal{D}) := \argmin_{W \in \mathbb{R}^{K \times d}} \|W\|_F^2 
    \quad \text{subject to} \quad 
    C(W; \mathcal{D}) \geq 1,
\end{align}
where $C(W; \mathcal{D})$ is the minimum margin over the dataset:
\begin{align}
    \textstyle C(W; \mathcal{D}) := \min_{(\boldsymbol{x}, \boldsymbol{y}) \in \mathcal{D}} \; 
    \min\limits_{k \ne c, y_c = 1}
    \left( \boldsymbol{w}_c^\top \boldsymbol{x} - \boldsymbol{w}_k^\top \boldsymbol{x} \right),
    \quad \text{with} \quad W = [\boldsymbol{w}_1; \dots; \boldsymbol{w}_K]^\top.
\end{align}
The hard-margin SVM problem in \eqref{eqn:svm} seeks a weight matrix of minimum norm that separates all examples with at least unit margin between the correct class and the nearest competing class. 

Intuitively, classification decisions depend primarily on the direction of the weight matrix rather than its scale. To enable comparisons across different scales of $W$, we adopt the \emph{normalized margin} used in~\citep{crammer2001algorithmic}, i.e., $\textstyle \gamma(W; \mathcal{D}) \!:=\! \frac{C(W; \mathcal{D})}{\|W\|_F}$ for $W\ne0$. This definition has been used in recent theoretical studies~\citep{lyu2019gradient, ravi2024implicit}. We additionally set $\gamma(0;\mathcal D)\!:=\!0$ by convention.

\subsection{Pre-training and fine-tuning setup}\label{sec:pre_ft_setup}

We now describe the specific pre-training and fine-tuning framework used in our theoretical analysis.

\myparagraph{Pre-training stage}
Let $\mathcal{D}_{\mathrm{pre}}$ be a labeled dataset with $K$ classes, and let $W_{\mathrm{pre}}\!\in\!\mathbb{R}^{K \times d}$ denote a pre-trained linear classifier obtained by GD minimizing the cross-entropy loss $L(W;\mathcal{D}_{\mathrm{pre}})$. When $\mathcal{D}_{\mathrm{pre}}$ is linearly separable, this objective admits no finite minimizer. The GD iterates instead diverge in norm while converging in direction to the max-margin solution~\citep{lyu2019gradient,soudry2018implicit}.
We formalize this directional structure in Assumption~\ref{asmp:pretrain_w}.

\myparagraph{Fine-tuning stage}
We adapt the pre-trained model to a downstream dataset $\mathcal{D}_{\mathrm{ft}}$ consisting of $n$ additional samples from each of $2\leq\bar K\leq K$ classes selected from the original $K$ pre-training classes. This restriction applies throughout the paper and appendix, including to every task in adapter merging.
This setup captures practical scenarios such as \emph{domain-incremental learning}~\citep{esaki2024one}, \emph{domain adaptation}~\citep{zohrizadeh2019class, zhang2022few}, and \emph{bias-rebalancing fine-tuning}~\citep{li2021discover, valizadehaslani2024two}.

When LoRA is applied to adapt the pre-trained model to the fine-tuning task, we introduce a low-rank update to the weights, parameterized by matrices $B \in \mathbb{R}^{K \times r}$ and $A \in \mathbb{R}^{r \times d}$:
\begin{align}\label{eqn:ft_obj}
    \textstyle\min_{B, A} \;
    L(W_{\mathrm{pre}} + BA; \mathcal{D}_{\mathrm{ft}}) + 
    \frac{\lambda}{2}(\|B\|_F^2 + \|A\|_F^2).
\end{align}
Here, the Frobenius-norm penalties explicitly constrain the adapters in weight space, limiting the deviation of the fine-tuned model from the pretrained initialization.

Although some empirical studies of LoRA do not include this penalty explicitly in the loss, they train the adapters using the AdamW optimizer~\citep{loshchilov2017decoupled}, whose decoupled weight decay induces an effect analogous to $L_2$ regularization on $B$ and $A$. Thus, the objective in~\eqref{eqn:ft_obj} captures the regularization that is, explicitly or implicitly, used in standard LoRA fine-tuning practice~\citep{hu2022lora, wistuba2023continual, biderman2024lora}.

Our goal is to understand the structure of the optimal solution to~\eqref{eqn:ft_obj}, and how it explains both LoRA's robustness to catastrophic forgetting and its effectiveness in enabling adapter merging.

\subsection{Assumptions and their implications}
Our theoretical analysis uses the following assumptions.
Let $N_{\mathrm{ft}}\!=\!\bar K n$ be the number of fine-tuning samples,
write $X_{\mathrm{ft}}\!=\![\boldsymbol{\bar x}_1,\ldots,\boldsymbol{\bar x}_{N_{\mathrm{ft}}}]$,
and let $X_{\mathrm{pre}}$ collect the pre-training feature vectors as columns.

\begin{asmp}[Task geometry]\label{asmp:data}
The combined pre-training and fine-tuning datasets are linearly separable.
There exist $\epsilon_1,\epsilon_2\!\in\![0,1)$ such that, for all $i,j\!\in\![N_{\mathrm{ft}}]$ and $\boldsymbol{x}\!\in\!\mathcal D_{\mathrm{pre}}$,
\begin{align}
    (a)\ \|\boldsymbol{\bar x}_i\|_2\!=\!\|\boldsymbol{x}\|_2\!=\!1,\quad
    (b)\ |\boldsymbol{\bar x}_i^{\top}\boldsymbol{\bar x}_j|\!\le\!\epsilon_1\ (i\ne j),\quad
    (c)\ |\boldsymbol{\bar x}_i^{\top}\boldsymbol{x}|\!\le\!\epsilon_2.
\end{align}
\end{asmp}

\begin{asmp}[Dimension]\label{asmp:dim}
In the strict-orthogonal case $\epsilon_1\!=\!\epsilon_2\!=\!0$, the input dimension satisfies
$d\!\ge\!N_{\mathrm{ft}}\!+\!\operatorname{rank}(X_{\mathrm{pre}})$.
For near-orthogonal data, we assume $(N_{\mathrm{ft}}\!-\!1)\epsilon_1\!<\!1$.
\end{asmp}


\begin{asmp}[LoRA rank]\label{asmp:lora_rank}
The LoRA rank parameter satisfies $r\!\ge\!\bar K$, where $\bar K$ is the number of fine-tuning classes.
\end{asmp}

\begin{asmp}[Pre-trained classifier]\label{asmp:pretrain_w}
The pre-trained classifier $W_{\mathrm{pre}}$ is a scaled version of the hard-margin SVM solution on $\mathcal{D}_{\mathrm{pre}}$:
$W_{\mathrm{pre}}\!=\!\rho_{\mathrm{pre}}\!\cdot\!W_{\mathrm{pre}}^{\mathrm{SVM}}/\|W_{\mathrm{pre}}^{\mathrm{SVM}}\|_F$,
for some scalar $\rho_{\mathrm{pre}} \!>\! 0$, where $W_{\mathrm{pre}}^{\mathrm{SVM}} \!=\! W^{\mathrm{SVM}}(\mathcal{D}_{\mathrm{pre}})$.
\end{asmp}

\myparagraph{Assumption justification}
The near-orthogonal task Assumption~\ref{asmp:data} is arguably a strong assumption in practice, but reflects the realistic geometric structure of fine-tuning data, and similar orthogonality-type assumptions are commonly used in theoretical analyses to enable closed-form characterizations~\citep{bui2021inductive, boursier2022gradient, frei2022implicit, kou2023implicit}. 
The feature-correlation analyses on CIFAR-100 and six multi-domain benchmarks in Appendices~\ref{appsec:validation_asmp_orthogonal_data} and~\ref{appsec:multidomain_feature_correlations} provide empirical motivation for studying this reference geometry.
Assumption~\ref{asmp:dim} makes the geometric feasibility and conditioning requirements explicit.
The strict-case bound accounts for both the fine-tuning sample count and the pre-training feature span, specifying a high-dimensional, sample-limited setting.
\textit{Assumption~\ref{asmp:lora_rank}} requires only $r \!\geq\! \bar K$, meaning the LoRA rank needs only to be at least the number of fine-tuning classes.
Finally, \textit{Assumption~\ref{asmp:pretrain_w}} is motivated by recent theoretical results on the implicit bias of GD in multiclass classification~\citep{lyu2021gradient, ravi2024implicit}, as discussed in \S\ref{sec:background}. $\rho_{\mathrm{pre}}$ captures the (unspecified) scale at which pre-training was stopped, which arises naturally in implicit-bias analyses where the GD iterate diverges along the SVM direction.

For clarity of exposition, all theorems in the main paper are stated under the strict-orthogonal special case ($\epsilon_1\!=\!\epsilon_2\!=\!0$) of Assumption~\ref{asmp:data}.
The general versions of these results for $\epsilon_1,\epsilon_2 \!>\! 0$, together with their proofs, are deferred to the Appendix~\ref{app:proof_main_theorem}--\ref{appsec:proof_merge}.

\myparagraph{Implications}  
The assumptions introduced above lead to a clean closed-form expression for the $\bar{K}$-class hard-margin SVM solution for the fine-tuning task, and also imply \emph{near-orthogonality} between the pre-trained weights $W_{\mathrm{pre}}$ and the fine-tuning dataset $\mathcal{D}_{\mathrm{ft}}$, 
which underpins our subsequent theoretical analysis in \S\ref{sec:main}.
To formalize these implications, we begin by introducing the concept of a \emph{simplex equiangular tight frame} (simplex ETF), which characterizes the geometry of the hard-margin SVM solution in our setting
\begin{defn}[Simplex ETF]\label{def:etf}
    For $m\!\geq\!2$, an $m$-simplex ETF is a collection of $m$ vectors in $\mathbb{R}^p$ for some $p\!\geq\!m$, given by the columns of $M_m \!:=\! \sqrt{\frac{m}{m-1}}\, P\left(\boldsymbol{I}_m \!-\! \frac{1}{m} \boldsymbol{1}_m \boldsymbol{1}_m^\top\right)$,
    where $P \in \mathbb{R}^{p \times m}$ satisfies $P^\top P \!=\! \boldsymbol{I}_m$. 
\end{defn}
Throughout, we use the canonical choice $p\!=\!m$ and $P\!=\!\boldsymbol I_m$, so $M_m\!\in\!\mathbb{R}^{m\times m}$.
Using this definition, we now present a proposition that, under the strict-orthogonal specialization, characterizes the structure of the hard-margin SVM solution for the fine-tuning data ($\mathcal{D}_{\mathrm{ft}}$) and the orthogonality between the pre-trained weights ($W_{\mathrm{pre}}$) and the fine-tuning data.
\begin{samepage}
\begin{prop}\label{prop:mm_ft}
Let $\bar K\!\ge\!2$. Under Assumption~\ref{asmp:data} with $\epsilon_1\!=\!\epsilon_2\!=\!0$ and Assumption~\ref{asmp:pretrain_w}, the following hold:
\begin{enumerate}[label=(\roman*),leftmargin=*]
    \item The minimum-norm SVM solution $W_{\mathrm{ft}}^{\mathrm{SVM}}\!\in\!\mathbb R^{\bar K\times d}$ for $\mathcal D_{\mathrm{ft}}$, defined by~\eqref{eqn:svm} with $\bar K$ output classes, admits the closed-form solution $W_{\mathrm{ft}}^{\mathrm{SVM}} \!=\! \sqrt{\frac{\bar K\!-\!1}{\bar K}}\left(M_{\bar K} \!\otimes\! \boldsymbol{1}_n^\top\right) X_{\mathrm{ft}}^\top$, where $\boldsymbol{x}_{i,j}$ is the $j$-th sample from class $i$, and $X_{\mathrm{ft}} \!=\! [\boldsymbol{x}_{1,1}, \dots, \boldsymbol{x}_{1,n}, \dots, \boldsymbol{x}_{\bar{K},n}]$ is the fine-tuning data matrix.

    \item The pre-trained weights are orthogonal to the fine-tuning data: $W_{\mathrm{pre}}\boldsymbol{\bar x} = 0$ for all $\boldsymbol{\bar x}\in\mathcal{D}_{\mathrm{ft}}$.
\end{enumerate}
\end{prop}
\end{samepage}

%
\myparagraph{Remark on Proposition~\ref{prop:mm_ft}} 
Proposition~\ref{prop:mm_ft} reveals that the hard-margin SVM solution for the fine-tuning task admits a compact closed-form expression: $W_{\mathrm{ft}}^{\mathrm{SVM}} \!=\! \sqrt{\frac{\bar K-1}{\bar K}}\left(M_{\bar K} \otimes \boldsymbol{1}_n^\top\right) X_{\mathrm{ft}}^\top$.
However, this form does not immediately reveal how the classifier aggregates information from the fine-tuning data. 
To provide insight, we explicitly derive the expression for the \emph{first row} of $W_{\mathrm{ft}}^{\mathrm{SVM}}$ (the remaining rows follow symmetrically), yielding $\textstyle\left(\frac{\bar K-1}{\bar K} \sum_{i=1}^n \boldsymbol{\bar x}_{1,i} - \frac{1}{\bar K} \sum_{k=2}^{\bar K} \sum_{j=1}^{n} \boldsymbol{\bar x}_{k,j} \right)^\top.$

Intuitively, each row of $W_{\mathrm{ft}}^{\mathrm{SVM}}$ encodes a direction that emphasizes its corresponding class mean while uniformly suppressing the influence of all other classes, leading to class separation in max-margin classification. The second result in Proposition~\ref{prop:mm_ft} follows from the representer theorem for SVMs, which implies that the hard-margin SVM solution on pre-training data lies in the span of $\mathcal{D}_{\mathrm{pre}}$. By Assumption~\ref{asmp:pretrain_w}, $W_{\mathrm{pre}}$ inherits this property; combined with condition (c) of Assumption~\ref{asmp:data} at $\epsilon_2\!=\!0$, each row of $W_{\mathrm{pre}}$ is therefore orthogonal to the fine-tuning data.

Building upon these assumptions and propositions, we develop a rigorous theoretical framework that provides a principled explanation for LoRA's empirical advantages in the subsequent section. Our analysis reveals properties that enable precise quantification of LoRA's benefits in terms of robustness to catastrophic forgetting and adapter merging performance.


\section{Main Results}\label{sec:main}
In this section, we first analyze the optimization landscape of the LoRA objective and characterize its global minimum in \S\ref{sec:thm_landscape}.
We then derive closed-form expressions for the margins on the union of pre-training and fine-tuning datasets, explaining why LoRA mitigates \textit{catastrophic forgetting} in \S\ref{sec:thm_forgetting}.
Finally, in \S\ref{sec:thm_adapter_merging}, we extend our margin-based analysis to the multi-task setting, explaining the effectiveness of \textit{adapter merging} and deriving optimal mixing coefficients that maximize the margin on the union of all tasks. Moreover, building on these closed-form coefficients, we translate them into a simple training-free merging rule, \emph{margin-based LoRA merging} (MaLoRA, Algorithm~\ref{alg:margin_lora_merge}).

\subsection{Global minimum of the LoRA objective}\label{sec:thm_landscape}
In this subsection, we present our main theoretical result characterizing the global minimizer of the LoRA objective in~\eqref{eqn:ft_obj}. Interestingly, under suitable regularization, we find that part of the optimal LoRA adapter aligns exactly with the $\bar K$-class hard-margin SVM solution ($W_{\mathrm{ft}}^{\mathrm{SVM}}$) for the fine-tuning data (see Proposition~\ref{prop:mm_ft} for its definition). While our analysis neither assumes nor relies on this alignment, its emergence reveals a geometric structure that partially explains LoRA's effectiveness.
\begin{thm}\label{thm:main}
Let $K$ and $\bar K \leq K$ be the number of pre-training and fine-tuning classes, respectively.
There exist scalar functions
$a_{\lambda},\,b_{\lambda},\,c_{\lambda},\Theta_{\lambda}$ of $\lambda$ such that for any global minimizer \(\bigl(B_{\lambda}^{\ast},A_{\lambda}^{\ast}\bigr)\) of
\eqref{eqn:ft_obj} with $\lambda\!>\!0$, we have
\begin{align}\label{app:lora_unified_form}
    \textstyle B_{\lambda}^{\ast}A_{\lambda}^{\ast}=\left(\begin{pmatrix}
        \bigl(a_{\lambda}+b_{\lambda}\bigr)\,\mathbf I_{\bar K}
        \;-\;
        b_{\lambda}\,\mathbf 1_{\bar K}\mathbf 1_{\bar K}^{\!\top}\\
        -c_{\lambda}\,\mathbf 1_{K-\bar K}\mathbf 1_{\bar K}^{\!\top}
    \end{pmatrix}
    \otimes \mathbf 1_n^\top\right) X_{\mathrm{ft}}^\top\,, \quad\|B_{\lambda}^{\ast}A_{\lambda}^{\ast}\|_F=\Theta_{\lambda}\,.
\end{align}
When \(\bar K\!\ge\! 2\), there exists a critical regularization weight $\lambda_{\mathrm{crit}}\!\in\![0,\frac{1}{\bar{K}\sqrt{n}})$ such that the scalar functions $a_{\lambda},\,b_{\lambda},\,c_{\lambda}, \Theta_{\lambda}$ are characterized as follows:
\begin{enumerate}[label=(\roman*),leftmargin=*]
    \item \textit{High-penalty regime} \(\bigl(\lambda\!\ge\! \frac{1}{\bar{K}\sqrt{n}}\bigr)\): $a_{\lambda}\!=\!b_{\lambda}\!=\!c_{\lambda}\!=\!\Theta_{\lambda}\!=\!0$, thus $B_{\lambda}^{\ast}A_{\lambda}^{\ast}\!=\!\mathbf0_{K\times d}\,.$

    \item \textit{Intermediate regime}
        \(\bigl(\lambda_{\mathrm{crit}}\!<\!\lambda\!<\!\frac{1}{\bar{K}\sqrt{n}}\bigr)\): $a_{\lambda}\!=\!\frac{\Theta_{\lambda}\sqrt{\bar{K}-1}}{\bar{K}\sqrt{n}},
            b_{\lambda}\!=\!\frac{\Theta_{\lambda}}{\bar{K}\sqrt{n(\bar{K}-1)}}, c_{\lambda}\!=\!0,$
        and $\Theta_{\lambda}$ is the unique root of a nonlinear equation (see \eqref{appeqn:grad_psi} in Appendix~\ref{app:proof_main_theorem}), thus the minimizer is
        \begin{align}\label{eqn:structure_lora_intermediate}
            \textstyle B_{\lambda}^{\ast}A_{\lambda}^{\ast}\;=\;\frac{\Theta_{\lambda}}{\sqrt{\bar{K}n}}\,\left(\begin{pmatrix}M_{\bar{K}}\\ \boldsymbol 0_{(K-\bar{K})\times \bar{K}}\end{pmatrix}\otimes \boldsymbol 1_n^\top\right) X_{\mathrm{ft}}^\top.
        \end{align}

    \item \textit{Low-penalty regime} \(\bigl(0\!<\!\lambda\!\leq\!\lambda_{\mathrm{crit}}\bigr)\):
        $a_{\lambda},b_{\lambda},c_{\lambda},\Theta_{\lambda}$ are non-negative.
\end{enumerate}
Moreover, the small-regularization limit satisfies
\begin{align}\label{eqn:structure_lora_low}
    \textstyle\lim_{\lambda\to 0^{+}}\Theta_{\lambda}\!=\!\infty\,, \qquad
    \lim_{\lambda\to 0^{+}}
    \frac{B_{\lambda}^{\ast}A_{\lambda}^{\ast}}{\lVert B_{\lambda}^{\ast}A_{\lambda}^{\ast}\rVert_{F}}
    \!=\!
    \sqrt{\frac{1}{n\bar{K}}} \left(M_K^{(\bar{K})}\otimes \boldsymbol 1_n^\top \right)X_{\mathrm{ft}}^\top\,,
\end{align}
where $M_K^{(\bar{K})}$ is the first $\bar{K}$ columns of the K-ETF matrix $M_K$.
\end{thm}

\begin{rem}
For $\bar K\!\ge\!2$, the critical weight satisfies $\lambda_{\mathrm{crit}}\!=\!0$ if and only if $\bar K\!=\!K$. In this case, the intermediate regime covers all $0\!<\!\lambda\!<\!1/(K\sqrt n)$, and the low-penalty regime contains no positive regularization weights. The limits in~\eqref{eqn:structure_lora_low} remain valid as $\lambda\!\to\!0^+$, approached through the intermediate regime. 
\end{rem}

The proof of Theorem~\ref{thm:main} is provided in Appendix~\ref{app:proof_main_theorem}. The theorem shows that the optimal LoRA update $B_\lambda^{\ast}A_\lambda^{\ast}$ has the unified form in~\eqref{app:lora_unified_form}, and identifies up to three regimes based on the regularization parameter $\lambda$. We make the following remarks:

\myparagraph{Effect of regularization and connection with max-margin classifier}
The theorem shows that the regularization parameter governs the structure of the optimal LoRA adapters. In the \textit{high-penalty regime}, the regularization term dominates and drives the optimal adapters to zero. In the \textit{intermediate regime}, the adapters balance cross-entropy and regularization, yielding the explicit form in \eqref{eqn:structure_lora_intermediate}. In particular, the first $\bar K$ rows of $B_\lambda^\ast A_\lambda^\ast$ align with the $\bar K$-class hard-margin SVM solution on the fine-tuning data (Proposition~\ref{prop:mm_ft}), which gives a clear max-margin interpretation of the learned solution. In the \textit{low-penalty regime}, explicit formulas for $a_\lambda$ and $b_\lambda$ are no longer available. Nevertheless, as $\lambda\to 0^+$, the cross-entropy term dominates and $\Theta_\lambda\to\infty$, while our asymptotic analysis still characterizes the limiting direction of $B_\lambda^\ast A_\lambda^\ast$ in \eqref{eqn:structure_lora_low}. This direction coincides with the hard-margin SVM solution only when $\bar K = K$.

\myparagraph{Implication for the fine-tuned model}  
The structured form of the optimal LoRA adapters in \eqref{app:lora_unified_form} naturally enables $W_{\mathrm{LoRA}}^{\lambda} \!:=\!W_{\mathrm{pre}} \!+\! B_{\lambda}^{\ast} A_{\lambda}^{\ast}$ to perform well on both pre-training and fine-tuning data. Under Assumptions~\ref{asmp:data} and~\ref{asmp:pretrain_w}, Proposition~\ref{prop:mm_ft} and Theorem~\ref{thm:main} imply that $W_{\mathrm{pre}} \boldsymbol{\bar x} = 0$ for all $\boldsymbol{\bar x} \in \mathcal{D}_{\mathrm{ft}}$ and $B_{\lambda}^{\ast} A_{\lambda}^{\ast} \boldsymbol{x} = 0$ for all $\boldsymbol{x} \in \mathcal{D}_{\mathrm{pre}}$. That is, the two components operate independently: $W_{\mathrm{pre}}$ classifies the pre-training data, while $B_{\lambda}^{\ast} A_{\lambda}^{\ast}$ adapts to the fine-tuning data:
\begin{align}
\forall\boldsymbol{x}\in\mathcal{D}_{\mathrm{pre}}\cup\mathcal{D}_{\mathrm{ft}}\,,\quad 
W_{\mathrm{LoRA}}^{\lambda} \boldsymbol{x} = 
\begin{cases}
    W_{\mathrm{pre}} \boldsymbol{x} & \text{if } \boldsymbol{x} \in \mathcal{D}_{\mathrm{pre}}, \\
    B_{\lambda}^{\ast} A_{\lambda}^{\ast} \boldsymbol{x} & \text{if } \boldsymbol{x} \in \mathcal{D}_{\mathrm{ft}}.
\end{cases} \quad \text{holds}\quad \forall \lambda>0\,.
\end{align}
This clean separation lets us decompose the normalized margin of $W_{\mathrm{LoRA}}^{\lambda}$ on each component dataset into a contribution from the relevant component (either $W_{\mathrm{pre}}$ or $B_{\lambda}^{\ast}A_{\lambda}^{\ast}$) and a relative-norm factor:
%
\begin{align}\label{eqn:margin_expression}
\gamma(W_{\mathrm{LoRA}}^{\lambda};\mathcal D)=\gamma(W;\mathcal D)\frac{\|W\|_F}{\|W_{\mathrm{LoRA}}^{\lambda}\|_F},\qquad
(W,\mathcal D)\in\left\{(W_{\mathrm{pre}},\mathcal D_{\mathrm{pre}}),(B_\lambda^\ast A_\lambda^\ast,\mathcal D_{\mathrm{ft}})\right\}.
\end{align}
%
The expression in \eqref{eqn:margin_expression} shows that the normalized margin on the combined dataset depends not only on the individual normalized margins of $W_{\mathrm{pre}}$ and $B_{\lambda}^{\ast} A_{\lambda}^{\ast}$ on the $\mathcal{D}_{\mathrm{pre}}$ and $\mathcal{D}_{\mathrm{ft}}$ respectively, but also on the relative Frobenius norms of these components. In the following sections, we use \eqref{eqn:margin_expression} to derive closed-form expressions for the normalized margin on the combined dataset and to compute the optimal mixing coefficients for \emph{adapter merging}. For clarity of presentation, we focus on the case $\bar{K} = K$ and refer the reader to Appendix~\ref{appsec:proof_forget} and Appendix~\ref{appsec:proof_merge} for the full version of our results.

\subsection{Margin trade-off between forgetting and fine-tuning under LoRA adaptation}\label{sec:thm_forgetting}
We now analyze catastrophic forgetting through the normalized margin. Our goal is to characterize how the LoRA fine-tuned model trades off performance on the pre-training data and the fine-tuning data as the regularization parameter varies, and to identify the regularization level that maximizes the margin on their union. For convenience, define $\gamma_{\mathrm{pre}}\!:=\!\gamma(W_{\mathrm{pre}};\mathcal D_{\mathrm{pre}}),\gamma_{\mathrm{ft},\lambda}\!:=\!\gamma(B_\lambda^\ast A_\lambda^\ast;\mathcal D_{\mathrm{ft}})$. Moreover, we define the adaptation to the fine-tuning data and forgetting on the pre-training data by
\begin{align}
    \mathrm{Adapt}(\lambda)\!=\!\gamma\bigl(W_{\mathrm{LoRA}}^\lambda;\mathcal D_{\mathrm{ft}}\bigr)\!-\!\gamma(W_{\mathrm{pre}},\mathcal D_{\mathrm{ft}}), \quad \mathrm{Forget}(\lambda)\!:=\!\gamma_{\mathrm{pre}}\!-\!\gamma\bigl(W_{\mathrm{LoRA}}^\lambda;\mathcal D_{\mathrm{pre}}\bigr)\,.
\end{align}
Based on these definitions, we now present the theorem.
\begin{thm}\label{thm:lora_forget_less}
  Adopt the setup of Theorem~\ref{thm:main} under strict orthogonality, and assume $\bar K=K\!\ge\!2$.
  Let $\gamma_{\mathrm{ft}}^{\ast}$ be the max normalized margin any linear classifier can obtain on the fine-tuning data $\mathcal{D}_{\mathrm{ft}}$, and recall the scalar $\rho_{\mathrm{pre}}$ from Assumption~\ref{asmp:pretrain_w}.
  Then, $\mathrm{Adapt}(\lambda)$ and $\mathrm{Forget}(\lambda)$ can be characterized uniformly over all $\lambda$ as follows:
  \begin{align}\label{eqn:margin_combined}
      \mathrm{Adapt}(\lambda)
       \!=\! \gamma_{\mathrm{ft,\lambda}}\,
          \frac{\Theta_{\lambda}}{\sqrt{\Theta_{\lambda}^2+\rho_{\mathrm{pre}}^{2}}}\,, \quad \mathrm{Forget}(\lambda)=\gamma_{\mathrm{pre}}\biggl(1-\frac{\rho_{\mathrm{pre}}}{\sqrt{\Theta_\lambda^2+\rho_{\mathrm{pre}}^2}}\biggr)\,.
  \end{align}
    Moreover, the effect of the regularization parameter $\lambda$ is characterized by two regimes:
    \begin{enumerate}[label=(\roman*),leftmargin=*]
        \item \textit{High-penalty regime:} if $\lambda \!\ge\! \frac{1}{\bar K\sqrt n}$, then $\Theta_\lambda\!=\!0$, and hence $\mathrm{Forget}(\lambda)\!=\!\mathrm{Adapt}(\lambda)\!=\!0$.
        \item \textit{Intermediate and low-penalty regimes:} if $\lambda \!<\! \frac{1}{\bar K\sqrt n}$, then $\gamma_{\mathrm{ft},\lambda}\!=\!\gamma_{\mathrm{ft}}^\ast$, and $\Theta_\lambda$ is decreasing in $\lambda$. Therefore, $\mathrm{Adapt}(\lambda)$ and $\mathrm{Forget}(\lambda)$ are both decreasing in $\lambda$. In particular, 
        \begin{align}
            \lim_{\lambda\to 0^+}
            \bigl(\mathrm{Adapt}(\lambda),\mathrm{Forget}(\lambda)\bigr)
            \!=\!
            \bigl(\gamma_{\mathrm{ft}}^\ast,\gamma_{\mathrm{pre}}\bigr), \quad
            \lim_{\lambda\to \frac{1}{\bar K\sqrt n}}
            \bigl(\mathrm{Adapt}(\lambda),\mathrm{Forget}(\lambda)\bigr)
            \!=\!
            (0,0).
        \end{align}
    \end{enumerate}
\end{thm}
\begin{rem}Notably, the intermediate and low-penalty regimes coincide when $\bar{K} = K$. This merging is specific to the case $\bar{K} \!=\! K$; when $\bar{K} \!<\! K$, the two regimes remain distinct.
\end{rem}
%
\myparagraph{Balancing adaptation and forgetting}
Theorem~\ref{thm:lora_forget_less} shows that the same quantity 
$\Theta_\lambda$ controls both adaptation and forgetting. As $\lambda$ decreases, the adapter scale $\Theta_\lambda$ increases, improving 
$\mathrm{Adapt}(\lambda)$ on the fine-tuning data while also increasing 
$\mathrm{Forget}(\lambda)$ on the pre-training data. Thus, LoRA fine-tuning induces a direct margin trade-off: stronger adaptation comes at the cost of larger pre-training margin degradation. The natural question is how to choose $\lambda$ to balance these effects. 
In the following Corollary~\ref{cor:optimal_lambda_tradeoff}, we quantify this trade-off by studying 
the normalized margin on the union of the pre-training and fine-tuning datasets, which leads to an explicit optimal regularization level.

\begin{cor}[Optimal balance between adaptation and forgetting]
\label{cor:optimal_lambda_tradeoff}
Under the conditions of Theorem~\ref{thm:lora_forget_less}, with $\bar K\!=\!K$, the normalized margin of $W_{\mathrm{LoRA}}^\lambda$ on the union $\mathcal D_{\mathrm{pre}}\cup\mathcal D_{\mathrm{ft}}$ is
\begin{align}
    \gamma(W_{\mathrm{LoRA}}^\lambda;
    \mathcal D_{\mathrm{pre}}\cup\mathcal D_{\mathrm{ft}})
    =
    \min\bigl\{
        \gamma_{\mathrm{pre}}-\mathrm{Forget}(\lambda),
        \mathrm{Adapt}(\lambda)
    \bigr\}\,.
\end{align}
This margin is uniquely maximized at the balancing point $\lambda^\ast\!\in\!(0,1/(K\sqrt n))$, with
\begin{align}
    \max_{\lambda>0}
    \gamma(W_{\mathrm{LoRA}}^{\lambda};
    \mathcal D_{\mathrm{pre}}\cup\mathcal D_{\mathrm{ft}})
    =
    \left(
    \frac{1}{(\gamma_{\mathrm{ft}}^{\ast})^{2}}
    +
    \frac{1}{\gamma_{\mathrm{pre}}^{2}}
    \right)^{-1/2}\,.
\end{align}
\end{cor}

\subsection{Margin characterization of merged LoRA models and optimal mixing coefficients}\label{sec:thm_adapter_merging}
We now extend the margin-based analysis to the multi-task setting, where a single pre-trained model is independently adapted by $T$ LoRA adapters that we wish to merge into a unified model. 
Through the same margin lens, we derive closed-form expressions for the merged model's margin on each task and identify mixing coefficients that maximize the margin on the union of all tasks (Theorem~\ref{thm:lora_merge}). 
Building on this characterization, we design a practical LoRA merging algorithm \emph{margin-based LoRA merging} (MaLoRA, Algorithm~\ref{alg:margin_lora_merge}).

Let $\bigl(B_i^{\ast},A_i^{\ast}\bigr)$ denote the LoRA adapter trained on dataset $\mathcal D_{i}$, for $i\in[T]$. We merge the adapters with the pre-trained weight matrix as follows:
\begin{align}\label{eqn:unified_model}
    \textstyle W_{\mathrm{LoRA}}(\boldsymbol \alpha)
  \;:=\;
  W_{\mathrm{pre}}
  \;+\;
  \sum_{i=1}^{T}\alpha_i\,B_i^{\ast}A_i^{\ast},
  \qquad
  \alpha_1,\dots,\alpha_T\geq0,
\end{align}
where \(\boldsymbol \alpha=(\alpha_1,\dots,\alpha_T)\) are mixing coefficients.
Moreover, we generalize Assumption~\ref{asmp:data} to the setting of learning multiple tasks as follows. 
\begin{asmp}\label{asmp:merge} 
There exist $\epsilon_1,\epsilon_2\in[0,1)$ such that 
\begin{enumerate}[label=(\alph*),leftmargin=*] 
    \item Every sample has unit norm: $\|\boldsymbol{x}\|_2=1$ for every $\boldsymbol{x}\in\bigcup_{i=1}^T\mathcal D_i\cup \mathcal{D}_{\mathrm{pre}}$. 
    \item \emph{Within-task:} for each $i\in[T]$ and each pair of distinct $\boldsymbol{\bar x},\boldsymbol{\bar x'}\in\mathcal D_i$, $|\boldsymbol{\bar x}^{\top}\boldsymbol{\bar x'}|\le\epsilon_1$. 
    \item \emph{Cross-task and pre-training:} $\forall \boldsymbol{\bar x}\in\mathcal D_i$, $\boldsymbol{\bar x'}\in\mathcal D_j, i\neq j$, or $x\in\mathcal D_{\mathrm{pre}}$, then $|\boldsymbol{\bar x}^{\top}\boldsymbol{\bar x'}|,\, |\boldsymbol{\bar x}^{\top}\boldsymbol{x}|\le \epsilon_2$.
\end{enumerate} 
\end{asmp}
The following theorem characterizes the optimal mixing coefficients in the strict orthogonal case ($\epsilon_1\!=\!\epsilon_2\!=\!0$). We refer the reader to Appendix~\ref{appsec:proof_merge} for the proof.
\begin{thm}\label{thm:lora_merge}
  Under Assumption~\ref{asmp:merge}, suppose each fine–tuning task \(\mathcal D_{i}\,(i=1,\dots ,T)\) is trained
  with a regularization parameter in the intermediate or low-penalty regime.  
  Let $\Theta_{\lambda, i}\!=\!\|B_i^{\ast}A_i^{\ast}\|_F$ and $\gamma_i\!=\!\gamma(B_i^{\ast}A_i^{\ast};\mathcal{D}_i)$.
  For any nonnegative coefficient vector
  \(\boldsymbol\alpha=(\alpha_1,\dots ,\alpha_T)\in\mathbb R_{\geq0}^T\), the merged model
  \eqref{eqn:unified_model} achieves the normalized margins on each task as follows
  \begin{align}\label{eq:merge_margin_task}
  \!
  \begin{split}
    \textstyle \gamma\bigl(W_{\mathrm{LoRA}}(\boldsymbol\alpha);\mathcal D_{\mathrm{pre}}\bigr)
      \!=\!\;
        \frac{\gamma_{\mathrm{pre}}\rho_{\mathrm{pre}}}
             {\sqrt{\rho_{\mathrm{pre}}^{2}
                    +\sum_{j=1}^{T}\alpha_j^2\Theta_{\lambda, j}^2}}\,,\quad 
    \gamma\bigl(W_{\mathrm{LoRA}}(\boldsymbol\alpha);\mathcal D_{i}\bigr)
      \!=\!\;
        \frac{\gamma_{i}\alpha_i \Theta_{\lambda, i}}{\sqrt{\rho_{\mathrm{pre}}^{2}+\sum_{j=1}^{T}\alpha_j^2\Theta_{\lambda, j}^2}},\\
        \textstyle\gamma\bigl(W_{\mathrm{LoRA}}(\boldsymbol\alpha);\mathcal{D}_{\mathrm{pre}}\!\cup\! \{D_i\}_{i=1}^T\bigr)\!=\!\min\left\{\gamma\bigl(W_{\mathrm{LoRA}}(\boldsymbol\alpha);\mathcal D_{\mathrm{pre}}\bigr), \gamma\bigl(W_{\mathrm{LoRA}}(\boldsymbol\alpha);\mathcal D_{i}\bigr),i\in [T]\right\}\,.
  \end{split}
  \!
   \end{align}
  Choosing the weights
  \(
    \textstyle\alpha_i = \frac{C(W_{\mathrm{pre}}; \mathcal{D}_{\mathrm{pre}})}{C(B_iA_i; \mathcal{D}_i)},
    \; i\in[T],
  \)
  maximizes the margin on the union of all tasks:
  \begin{align*}
  \textstyle \max_{\boldsymbol\alpha\in\mathbb R_{\geq0}^T}
      \gamma\bigl(W_{\mathrm{LoRA}}(\boldsymbol\alpha);\mathcal{D}_{\mathrm{pre}}\cup \{D_i\}_{i=1}^T\bigr)\;=\; \left(\frac{1}{\gamma_{\mathrm{pre}}^{2}}
                  +\sum_{j=1}^{T}\frac{1}{(\gamma_{j})^{2}}\right)^{-1/2}\,.
  \end{align*}
\end{thm}
\begin{algorithm}[t]
\caption{Margin-based LoRA Merging (MaLoRA)}\label{alg:margin_lora_merge}
\begin{algorithmic}[1]
\Require $W_{\mathrm{pre}}$, $\{(B_i^\ast,A_i^\ast)\}_{i=1}^T$, $\mathcal D_{\mathrm{pre}}^{\mathrm{val}},\{\mathcal D_i^{\mathrm{val}}\}_{i=1}^T$, $S\subset\mathbb R_{>0}$
\State $\bar C_{\mathrm{pre}} \gets \bar C(W_{\mathrm{pre}};\mathcal D_{\mathrm{pre}}^{\mathrm{val}})$
\For{$i=1,\dots,T$}
    \State $\bar C_i \gets \bar C(W_{\mathrm{pre}}+B_i^\ast A_i^\ast;\mathcal D_i^{\mathrm{val}})$, \;
    $\beta_i \gets \begin{cases} \bar C_{\mathrm{pre}}/\bar C_i, & \bar C_i>0,\\ 0, & \text{otherwise} \end{cases}$
\EndFor
\State $s^\ast \in \arg\max_{s\in S}\mathrm{ValAcc}\!\left(W_{\mathrm{pre}}+s\sum_{i=1}^T\beta_i B_i^\ast A_i^\ast\right)$
\State \Return $W_{\mathrm{pre}}+s^\ast\sum_{i=1}^T\beta_i B_i^\ast A_i^\ast$
\end{algorithmic}
\end{algorithm}
Theorem~\ref{thm:lora_merge} characterizes the margins on each task as well as on the union of tasks as functions of the mixing coefficients. In particular, it shows that the optimal mixing coefficients are given by the ratio between the unnormalized margin of the pre-trained model on $\mathcal{D}_{\mathrm{pre}}$ and that of each fine-tuned adapter on its own task. The unnormalized margin measures how confidently a classifier separates classes on its dataset, with smaller values indicating more challenging classification problems. This formula admits a direct interpretation: an adapter that achieves a smaller margin on its task is assigned a larger mixing weight, increasing its contribution so as to preserve a positive margin for that task in the merged model.

\paragraph{Towards practical margin-based adapter mixing}
The preceding analysis suggests choosing adapter mixing coefficients according to how confidently each adapter separates its task data, as measured by the unnormalized margin. In practice, however, datasets are often not perfectly separable, so the minimum margin $C(\cdot;\cdot)$ can be negative and overly sensitive to hard examples. We therefore use the average margin $\bar C(\cdot;\mathcal D)$ as a more stable proxy for adapter performance.
Under the strict-orthogonal assumptions of Theorem~\ref{thm:lora_merge}, the per-sample margins of each optimal adapter are equal on its own fine-tuning dataset. Consequently, minimum and average margins coincide on these fine-tuning datasets.
In our practical algorithm, we therefore use average margins as empirical proxies for the minimum margins appearing in the theoretical mixing rule.

\paragraph{Algorithm~\ref{alg:margin_lora_merge}: Margin-based LoRA merging (MaLoRA).}
Motivated by Theorem~\ref{thm:lora_merge}, we propose a practical LoRA merging algorithm using the average-margin proxies $\bar C(W;\mathcal D)$.
When a task has non-positive average margin $\bar C_i$, the corresponding coefficient would be negative and no longer consistent with the regime motivated by the theorem. Algorithm~\ref{alg:margin_lora_merge} therefore drops such tasks by setting their coefficients to zero. We also tune a global scaling factor $s^\ast$ on a held-out validation set, following \cite{ilharco2022editing}, and find that this further improves performance. In \S\ref{sec:lora_merging_comparison}, we compare the resulting method with other popular adapter-merging baselines.

\section{Experiments}\label{sec:experiment}
Our experiments aim to validate the theoretical predictions in Section~\ref{sec:main} and assess how well they transfer beyond the linear setting, rather than to establish state-of-the-art performance. Since our analysis gives closed-form predictions for optimal regularization and adapter mixing coefficients, we test whether these quantities match empirical behavior in modern architectures.
Specifically, we focus on two questions: \emph{(i)} whether the mixing coefficients derived from our margin analysis agree with empirically optimal coefficients, and \emph{(ii)} whether the resulting theory-driven merging rule, MaLoRA, remains competitive with adapter-merging baselines in nonlinear models. Experiments on the effect of regularization are deferred to Appendix~\ref{sec:margin}.

\myparagraph{Setup and metric}
We evaluate our theory and method in two complementary settings. First, in linear probing on CIFAR-100~\citep{krizhevsky2009learning} (\S\ref{sec:adapter_merging}), we use four frozen pre-trained backbones (\texttt{ResNet-50}, \texttt{ViT-B/16}, \texttt{ConvNeXt}, and \texttt{CLIP}) as feature extractors~\citep{He_2016_CVPR,dosovitskiy2021vit,radford2021learning,liu2022convnet}, apply LoRA only to the final linear classifier~\citep{kornblith2019better}, and construct tasks from the CIFAR-100 superclass hierarchy. Second, in multi-domain adapter merging (\S\ref{sec:lora_merging_comparison}), we evaluate ViT-B/16 on PACS, VLCS, and TerraIncognita~\citep{fang2013unbiased,li2017deeperbroaderartierdomain,beery2018recognition}, and XLM-RoBERTa Base on MNLI, Amazon Reviews, and CivilComments~\citep{zhang2015character,N18-1101,DBLP:journals/corr/abs-1903-04561,conneau2020unsupervised}. Although our theory is stated in terms of normalized margin, we report classification accuracy in experiments, since real-world classification tasks often do not achieve perfect separation and can have negative normalized margins. Additional details are given in Appendix~\ref{appsec:experiment}.

\subsection{Optimal mixing coefficients for adapter merging: theory vs. grid search}\label{sec:adapter_merging}

\begin{figure}[t]
    \centering
    \includegraphics[width=1\textwidth]{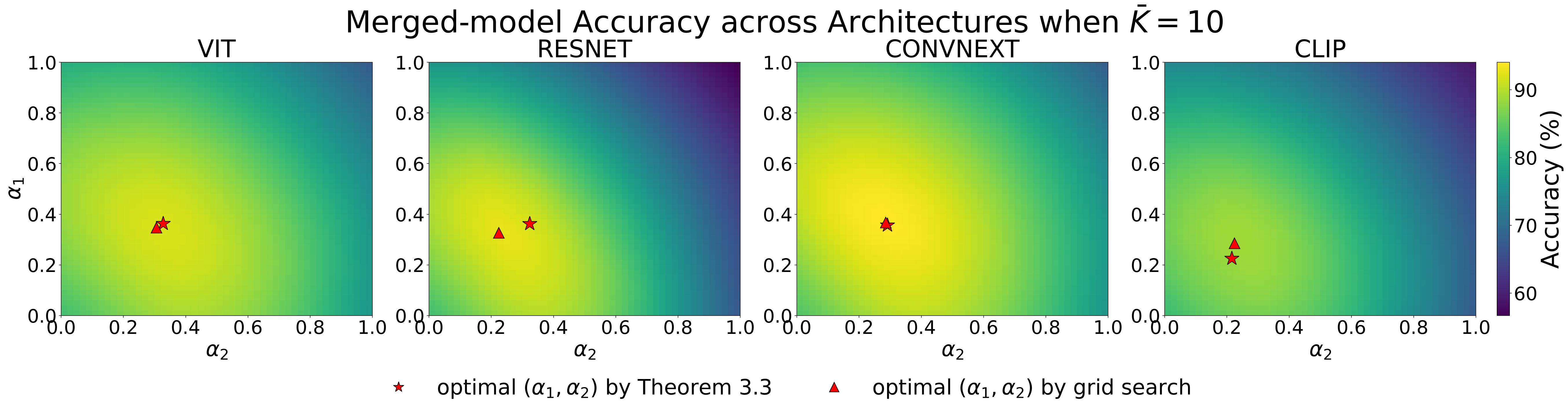}
    \caption{Merged-model accuracy across architectures. Each panel shows the accuracy of the merged model evaluated on the combined dataset, across a $50 \times 50$ grid of mixing coefficients $(\alpha_1, \alpha_2) \in (0,1)^2$. For each architecture, the red star indicates the theoretically predicted optimal coefficients from Theorem~\ref{thm:lora_merge}, while the red triangle marks the empirically optimal coefficients.}
    \label{fig:adapter_merging}
\end{figure}

We fine-tune LoRA adapters on two disjoint tasks ($\mathcal{D}_1$, $\mathcal{D}_2$), then merge them with the pre-trained weights using mixing coefficients $(\alpha_1, \alpha_2)$. We evaluate the merged model’s performance on the combined dataset $\mathcal{D}_{\mathrm{pre}} \cup \mathcal{D}_1 \cup \mathcal{D}_2$. As shown in Figure~\ref{fig:adapter_merging}, our predicted mixing coefficients closely match the grid-searched optima, confirming the practical accuracy of our merging theory. 
%
\subsection{Comparison between MaLoRA and other training-free baselines}\label{sec:lora_merging_comparison}
Following \cite{ilharco2022editing}, we report \emph{average normalized accuracy} (per-task merged accuracy divided by per-task finetuned accuracy, averaged across tasks) and its complement, \emph{relative accuracy drop}, in Table~\ref{tab:norm_acc_6cell}. 
The aggregate performance of the methods is comparable, with average normalized accuracy ranging from $92.87\%$ to $93.32\%$. MaLoRA lies at the top of this range, achieving an average normalized accuracy of $93.32\%$ and the smallest relative accuracy drop of $6.68\%$. 
We emphasize that although MaLoRA is competitive with baselines, our purpose here is not to achieve state-of-the-art results but to demonstrate the efficacy of our theoretically-grounded method.

%
\begin{table}[htbp]
\centering
\caption{Average normalized accuracy and relative accuracy drop, following~\cite{ilharco2022editing}. Results are averaged over 10 seeds. \textbf{Bold} indicates the best result in each column. 
}
\label{tab:norm_acc_6cell}
\footnotesize
\setlength{\tabcolsep}{4pt}
\begin{tabular}{l|cccccc|cc}
\toprule
\multicolumn{1}{c|}{}
& \multicolumn{6}{c|}{Per-benchmark normalized accuracy}
& \multicolumn{2}{c}{Aggregate metrics} \\
\cmidrule(lr){2-7}
\cmidrule(lr){8-9}
Method & PACS & VLCS & Terra Inc. & MNLI & Amazon & CivilComm. & Avg.\ norm.\ acc & Rel.\ drop \\
\midrule
Soup        & 96.38 & 96.89 & 64.31 & 101.16 & 100.64 & \textbf{98.27} & 92.94 & 7.06 \\
TA          & 97.52 & 97.19 & 64.46 & \textbf{101.27} & 100.68 & 97.80 & 93.15 & 6.85 \\
TIES        & 97.52 & 96.95 & 63.78 & 100.87 & 100.64 & 98.00 & 92.96 & 7.04 \\
DARE-TIES   & \textbf{97.57} & 97.23 & 64.23 & 100.90 & 100.45 & 97.92 & 93.05 & 6.95 \\
DARE-Linear & 96.48 & 96.77 & 64.73 & 101.12 & 100.50 & 97.61 & 92.87 & 7.13 \\
\rowcolor{gray!15}
\textbf{MaLoRA} & 96.69 & \textbf{97.54} & \textbf{65.75} & 101.23 & \textbf{100.69} & 98.00 & \textbf{93.32} & \textbf{6.68} \\
\bottomrule
\end{tabular}
\end{table}

\myparagraph{Computational overhead}
MaLoRA computes task margins once and reuses them throughout the scale search.
Under a separate matched $49$-configuration timing protocol, the mean runtime is $972.6$~s versus $950.3$~s for TA, approximately $2.35\%$ overhead across the six benchmarks.
Detailed timings and the measurement protocol are in Appendix~\ref{appsec:computational_overhead}.

\section{Conclusion, Limitations, and Future Work}\label{sec:conclusion}
In this work, we study LoRA fine-tuning through the normalized margin under a near-orthogonal task regime. We characterize the global minimizer of the regularized LoRA objective across three regularization regimes (Theorem~\ref{thm:main}), derive closed-form expressions for the margins on the pre-training and fine-tuning data, and identify the regularization level that maximizes the margin on their union (Theorem~\ref{thm:lora_forget_less}). We further extend the analysis to multi-task settings by deriving closed-form optimal mixing coefficients for adapter merging (Theorem~\ref{thm:lora_merge}). Based on this theory, we propose \emph{margin-based LoRA merging} (MaLoRA, Algorithm~\ref{alg:margin_lora_merge}), which matches or outperforms several adapter-merging baselines on a range of vision and language benchmarks. Across architectures and tasks, the empirical results closely align with the theoretical predictions.

\myparagraph{Limitations and future work}
First, our analysis relies on simplifying assumptions, most notably near-orthogonal task structure (Assumption~\ref{asmp:data}) and balanced class sizes in the fine-tuning data.
These assumptions enable a sharp closed-form characterization, but they also limit the scope of the theory.
An important direction for future work is to relax them and study LoRA under more general data distributions and pre-training conditions.
Second, experiments applying \textit{MaLoRA} to nonlinear models provide empirical support for the practical usefulness of our merging rule beyond multiclass linear classification.
These findings motivate extending the margin-based theoretical analysis to nonlinear adaptation.
Third, our theory characterizes margins on the training datasets; establishing corresponding generalization or test-risk guarantees requires further analysis.

\begin{ack}
Ziqing Xu, Lachlan Ewen MacDonald, Salma Tarmoun, and Ren\'e Vidal acknowledge support from NSF grant 2031985, Simons Foundation grant 814201 (Theorinet), ONR MURI 503405-78051, and University of Pennsylvania startup funds.
Enrique Mallada acknowledges support from NSF Global Centers grant 2330450 and DOE Office of Science (ASCR) award 826565.
Hancheng Min acknowledges support from the Science and Technology Commission of Shanghai Municipality through the Shanghai Pujiang Programme (grant 25PJA070) and Shanghai Jiao Tong University startup funds.
\end{ack}

\bibliographystyle{plainnat}
\bibliography{references}


\appendix
\clearpage
\tableofcontents
\clearpage

\section{LLM Usage}\label{sec:llm}
In this work, we use LLM to revise the writing, and help build standard training baselines of the experiments, such as dataset downloading, data processing, and pre-trained model loading. 

\section{Related work}\label{app:related_work}
\myparagraph{Theory of LoRA} 
There are many works theoretically studying the expressiveness, characterizing the loss landscape, and understanding the learning dynamics of LoRA.
\cite{zeng2023expressive} prove that, under mild assumptions, LoRA can approximate any deep linear, feed-forward, or transformer network. Within the NTK regime, \cite{malladi2023kernel} characterize the conditions under which one can study LoRA in the NTK regime, while \cite{jang2024lora} show that when the LoRA rank is $r_i\gtrsim\sqrt{N}$, where $N$ is the number of samples, the optimization landscape of LoRA has no spurious local minima, and GD can find $\mathcal{O}(\sqrt{N})$-rank solutions that generalize well. Moreover, \cite{xu2025understanding} study the learning dynamics of LoRA in the context of matrix factorization, and show that smaller initialization leads to longer training time and lower training error. However, none of these studies theoretically characterize why LoRA reduces \textit{catastrophic forgetting} and support effective \textit{adapter merging}.

\myparagraph{LoRA merging}
A variety of techniques have been proposed to \emph{merge} task-specific LoRA adapters by forming a weighted average of their parameters. For example, \textit{LoRA-Hub}~\citep{huang2023lorahub} learns per-task mixing coefficients first, and then applies a weighted average of LoRA weight matrices $B$ and $A$ separately. \textit{LoRA-Flow}~\citep{wang2024lora} introduces token-level gates that dynamically assign merging coefficients to each LoRA adapter before taking weighted averaging. \textit{LoRA-Retriever}~\citep{zhao2024loraretriever} first retrieves the most relevant adapters for each input, and then averages the selected LoRA adapters. Beyond weighted averaging, another line of work attempts to address redundancy and conflicting updates directly. \textit{TIES} ~\citep{yadav2023resolving} proceeds in three steps: trimming redundant parameters, resolving sign conflicts into an aggregate vector, and averaging only the parameters consistent with the aggregate sign, thereby mitigating degradation from redundant or conflicting updates. \textit{DARE}~\citep{yu2024language} can be used as a preprocessing step for other merging methods, where parameters are randomly dropped according to a specified rate and the remaining ones are rescaled, reducing redundancy and potential interference among merged adapters. However, all these methods remain largely heuristic and lack theoretical guarantees.

\myparagraph{Regularization and task orthogonality help reduce forgetting}
A large body of work in fine-tuning and continual learning shows that \emph{controlling deviation of the fine-tuned model from the pretrained model} reduces catastrophic forgetting. Concretely, these approaches impose \emph{weight-space} penalties that restrict the magnitude and direction of parameter updates away from the pretrained solution. Weight-anchoring and importance-aware penalties exemplify this idea: \cite{kirkpatrick2017overcoming} penalize movement along directions identified by the Fisher information matrix, while \cite{zenke2017continual} accumulate path-wise importance during training. \cite{schwarz2018progress} extend this idea to long sequences of tasks. Related approaches estimate parameter sensitivity \citep{aljundi2018mas}, apply output-level distillation to preserve prior behavior \citep{li2016lwf}, or explicitly regularize parameters toward the pretrained anchor \citep{li2018l2sp}. Although these approaches differ in how importance or anchoring is computed, the common mechanism is the same: \emph{weight-space regularizers constrain updates during fine-tuning relative to the pretrained model to mitigate catastrophic forgetting}.  
Another complementary perspective highlights the role of task relations. When task representations are sufficiently decorrelated, interference between tasks is naturally reduced and forgetting is alleviated. In linear and kernelized models, it has been shown that when tasks are orthogonal or nearly orthogonal, cross-task gradient interference vanishes in expectation, implying negligible forgetting \citep{doan2021theoretical,evron2022catastrophic}. These insights resonate with our analysis of LoRA, where we explicitly study a regularized objective on low-rank adapters and make the simplifying assumption that the fine-tuning data are orthogonal to the pretraining data. This assumption directly connects to prior findings on task orthogonality and provides a tractable setting in which we can isolate and analyze the role of regularization in mitigating forgetting.


\section{Characterization of Hard-margin SVM Solution for Orthogonal data}\label{appsec:svm_orthogonal}
In this section, we characterize the hard-margin SVM solution for $\bar{K}$-class classification problem for orthogonal data. 
We show the solution is closely related to the $\bar{K}$-ETF simplex.

\myparagraph{Problem Formulation}
Consider a $\bar{K}$-class classification problem with data $\{(x_{i,j}, y_i)\}_{i=1,j=1}^{\bar{K},n}$ where:
\begin{itemize}
    \item $\bar{K}\!\ge\!2$ is the number of classes in the dataset
    \item $n$ is the number of samples per class
    \item $y_i \in [\bar{K}]$ denotes the class label for samples $x_{i,j}$
    \item All data vectors $x_{i,j}$ have unit norm: $\|x_{i,j}\| = 1$
    \item All data vectors are mutually orthogonal: $x_{i,j}^T x_{k,l} = 0$ for $(i,j) \neq (k,l)$
\end{itemize}

Let $W \in \mathbb{R}^{\bar{K} \times d}$ be a weight matrix for a linear classifier, where the predicted class for input $x$ is $\argmax_k (W_k x)$. For simplicity, we use the following notation to denote all the data points: $X=[x_{1,1},\cdots,x_{\bar{K},n}]$.
We use the canonical ETF $M_{\bar K}\!:=\!\sqrt{\bar K/(\bar K-1)}\bigl(\boldsymbol I_{\bar K}\!-\!\boldsymbol 1_{\bar K}\boldsymbol 1_{\bar K}^{\top}/\bar K\bigr)$, corresponding to $P\!=\!\boldsymbol I_{\bar K}$ in Definition~\ref{def:etf}.

Now, we are ready to present the main result.

\begin{thm}[Maximum Margin with Frobenius Norm]\label{appthm:max_margin_frobenius}
Under the above data assumptions, the largest normalized margin any nonzero linear classifier can achieve is:
\begin{equation}
\gamma^* = \frac{1}{\sqrt{n(\bar{K}-1)}}
\end{equation}

The unique unit-margin minimum-Frobenius-norm SVM solution is
\begin{align*}
    W^{\mathrm{SVM}}=\sqrt{\frac{\bar K-1}{\bar K}}\left(M_{\bar K}\otimes\boldsymbol 1_n^\top\right)X^\top\,.
\end{align*}
The normalized margin is maximized precisely by $W\!=\!cW^{\mathrm{SVM}}$ with $c\!>\!0$.
\end{thm}

\medskip
\begin{proof}
We first solve the unit-margin minimum-norm SVM problem~\eqref{eqn:svm} with $\bar K$ output classes. By Proposition~\ref{prop:max_margin}, its solution has the form $W^{\mathrm{SVM}}\!=\!A^{\ast}X^\top$ for some $A^{\ast}\!\in\!\R^{\bar K\times\bar Kn}$. Since $X^\top X\!=\!\boldsymbol I_{\bar Kn}$, the logits satisfy $W^{\mathrm{SVM}}X\!=\!A^{\ast}$ and $\|W^{\mathrm{SVM}}\|_F\!=\!\|A^{\ast}\|_F$. Thus, $A^{\ast}$ solves
\begin{align}\label{appeqn:svm_obj}
    A^{\ast}&=\argmin_{A\in\R^{\bar K\times\bar Kn}}\frac12\|A\|_F^2
    \quad\text{subject to}\nonumber\\
    &A_{i,(i-1)n+j}-A_{k,(i-1)n+j}\ge1,
    \qquad i,k\in[\bar K],\ i\ne k,\ j\in[n].
\end{align}
This feasible problem has a strictly convex objective and a convex constraint set, so its solution is unique~\citep{burges1999uniqueness}.
We use the permutation invariances of this fixed-scale problem to characterize $A^{\ast}$.
We first decompose $A^{\ast}$ into block matrices, i.e., $A^{\ast}=\begin{bmatrix}
    A^{\ast}_1 & \cdots &A^{\ast}_{\bar{K}}
\end{bmatrix}$ where $A^{\ast}_i\in\R^{\bar{K}\times n}, i\in[\bar{K}]$.

\myparagraph{Invariance under column permutation within block matrices} Permuting columns within a class block preserves both the constraints and the objective in~\eqref{appeqn:svm_obj}. Thus, for any permutation matrices $\Pi_i\!\in\!\R^{n\times n}$, the matrix $\begin{bmatrix}
    A^{\ast}_1 \Pi_1 & \cdots &A^{\ast}_{\bar{K}} \Pi_{\bar{K}}
\end{bmatrix}$ is also optimal. Uniqueness gives $A^{\ast}_i\Pi_i\!=\!A^{\ast}_i$ for every $\Pi_i$, so all columns of each block are equal: $A^{\ast}_i\!=\!\boldsymbol v_i\boldsymbol 1_n^\top$ for some $\boldsymbol v_i\!\in\!\R^{\bar K}$.

\myparagraph{Invariance under permutation across block matrices} Let $V\!:=\![\boldsymbol v_1,\ldots,\boldsymbol v_{\bar K}]$. Simultaneously permuting output rows and the corresponding class blocks preserves the constraints and objective in~\eqref{appeqn:svm_obj}. Uniqueness therefore gives $\Pi V\Pi^\top\!=\!V$ for every permutation matrix $\Pi\!\in\!\R^{\bar K\times\bar K}$. By Lemma~\ref{applem:permutation}, $V\!=\!a\boldsymbol I_{\bar K}\!-\!b\boldsymbol 1_{\bar K}\boldsymbol 1_{\bar K}^\top$ for some $a,b\!\in\!\R$.

\myparagraph{Scalar optimization} The unit-margin constraints reduce to $a\!\ge\!1$, and
\begin{align*}
    \|A^{\ast}\|_F^2
    &=n\bar K(a-b)^2+n\bar K(\bar K-1)b^2\\
    &=n\left[(\bar K-1)a^2+\bar K^2\left(b-\frac{a}{\bar K}\right)^2\right].
\end{align*}
The unique minimum is attained at $a\!=\!1$ and $b\!=\!1/\bar K$. Hence,
\begin{align*}
    W^{\mathrm{SVM}}
    =\left[\left(\boldsymbol I_{\bar K}-\frac1{\bar K}\boldsymbol 1_{\bar K}\boldsymbol 1_{\bar K}^{\top}\right)\otimes\boldsymbol 1_n^\top\right]X^\top,
    \qquad \|W^{\mathrm{SVM}}\|_F=\sqrt{n(\bar K-1)},
\end{align*}
which gives the stated ETF formula and has $C(W^{\mathrm{SVM}};\mathcal D)\!=\!1$.

Finally, any classifier $W$ with $C(W;\mathcal D)\!>\!0$ can be rescaled to the feasible unit-margin classifier $W/C(W;\mathcal D)$. Minimum-norm optimality yields
\begin{align*}
    \gamma(W;\mathcal D)
    =\frac{1}{\|W/C(W;\mathcal D)\|_F}
    \le\frac{1}{\|W^{\mathrm{SVM}}\|_F}
    =\frac{1}{\sqrt{n(\bar K-1)}}.
\end{align*}
Equality holds precisely when $W/C(W;\mathcal D)\!=\!W^{\mathrm{SVM}}$, by uniqueness of the unit-margin minimum-norm solution. Thus all normalized-margin maximizers are exactly the positive multiples of $W^{\mathrm{SVM}}$. Classifiers with nonpositive margin cannot attain this positive optimum.
\end{proof}


\section{Proof of Theorem~\ref{thm:main}}\label{app:proof_main_theorem}
In this section, we first present the proof of Theorem~\ref{thm:main}.
Then, under additional assumptions on $\epsilon_1,\epsilon_2$, we show that the same conclusion still holds, with additional error terms that depend on $\epsilon_1,\epsilon_2$.

\subsection{Proof of Theorem~\ref{thm:main}}
For the case $2\!\le\!\bar K\!<\!K$, we first define the following functions used in the proof of Theorem~\ref{thm:main}. The balanced case $\bar K\!=\!K$ is treated separately below.
\begin{align}\label{appeqn:def_g}
    g_N(x)&:=\sqrt{\frac{\rho(K-\bar{K})}{\bar{K}(K-\bar{K})x^2+\bar{K}K(1-x)^2}}\nonumber\\
    g_a(x)&:=\exp\biggl(\frac{g_N(x)\sqrt{\frac{\rho}{\bar{K}n}}}{\sqrt{x^2+\frac{K}{K-\bar{K}}(1-x)^2}}\bigl[x^2+\sqrt{\frac{K}{K-\bar{K}}}(1-x)^2\bigr]\biggr)\nonumber\\
    g_b(x)&:=\exp\biggl(\frac{g_N(x)\sqrt{\frac{\rho}{\bar{K}n}}}{\sqrt{x^2+\frac{K}{K-\bar{K}}(1-x)^2}}\bigl[-\frac{x^2}{\bar{K}-1}+\sqrt{\frac{K}{K-\bar{K}}}(1-x)^2\bigr]\biggr)\nonumber\\
    g_c(x)&:=\exp\biggl(\frac{g_N(x)\sqrt{\frac{\rho}{\bar{K}n}}}{\sqrt{x^2+\frac{K}{K-\bar{K}}(1-x)^2}}\bigl[-\frac{\bar{K}}{K-\bar{K}}\cdot\sqrt{\frac{K}{K-\bar{K}}}(1-x)^2\bigr]\biggr)\,.
\end{align}
Moreover, let $x_0\in[0,1]$ be the root of the equation
\begin{align*}
    \frac{g_b(x)}{g_c(x)}=\frac{1}{\sqrt{\frac{K}{K-\bar{K}}}+1}\,.
\end{align*}
\begin{rem}
    Our proof technique follows \cite{fang2021exploring}, and the readers can see the same definition of $g_a(x), g_b(x), g_c(x)$ and $x_0$ in \cite{fang2021exploring}.
    Moreover, the authors prove the existence of $x_0$ under certain conditions. 
\end{rem}

Now, we begin the proof of Theorem~\ref{thm:main}.
\begin{thm}[Restatement of Theorem~\ref{thm:main}]
Let \(\bigl(B_{\lambda}^{\ast},A_{\lambda}^{\ast}\bigr)\) be any minimizer of \eqref{eqn:ft_obj}.  
There exist scalar functions
$a_{\lambda},\,b_{\lambda},\,c_{\lambda},\Theta_{\lambda}$ of $\lambda$ such that for any global minimizer \(\bigl(B_{\lambda}^{\ast},A_{\lambda}^{\ast}\bigr)\) of
\eqref{eqn:ft_obj} with $\lambda\!>\!0$, we have
\begin{align}
    B_{\lambda}^{\ast}A_{\lambda}^{\ast}=\left(\begin{pmatrix}
        \bigl(a_{\lambda}+b_{\lambda}\bigr)\,\mathbf I_{\bar K}
        \;-\;
        b_{\lambda}\,\mathbf 1_{\bar K}\mathbf 1_{\bar K}^{\!\top}\\
        -c_{\lambda}\,\mathbf 1_{K-\bar K}\mathbf 1_{\bar K}^{\!\top}
    \end{pmatrix}
    \otimes \mathbf 1_n^\top\right) X_{\mathrm{ft}}^\top\,, \quad\|B_{\lambda}^{\ast}A_{\lambda}^{\ast}\|_F=\Theta_{\lambda}\,.
\end{align}
When \(\bar K\!\ge\! 2\), there exists a critical regularization weight $\lambda_{\mathrm{crit}}\!\in\![0,\frac{1}{\bar{K}\sqrt{n}})$, with $\lambda_{\mathrm{crit}}\!=\!0$ if and only if $\bar K\!=\!K$, such that the scalar functions $a_{\lambda},\,b_{\lambda},\,c_{\lambda}, \Theta_{\lambda}$ are characterized as follows:
\begin{enumerate}[label=(\roman*),leftmargin=*]
    \item \textit{High-penalty regime} \(\bigl(\lambda\!\ge\! \frac{1}{\bar{K}\sqrt{n}}\bigr)\): $a_{\lambda}\!=\!b_{\lambda}\!=\!c_{\lambda}\!=\!\Theta_{\lambda}\!=\!0$, thus $B_{\lambda}^{\ast}A_{\lambda}^{\ast}\!=\!\mathbf0_{K\times d}\,.$

    \item \textit{Intermediate regime}
        \(\bigl(\lambda_{\mathrm{crit}}\!<\!\lambda\!<\!\frac{1}{\bar{K}\sqrt{n}}\bigr)\): $a_{\lambda}\!=\!\frac{\Theta_{\lambda}\sqrt{\bar{K}-1}}{\bar{K}\sqrt{n}},
            b_{\lambda}\!=\!\frac{\Theta_{\lambda}}{\bar{K}\sqrt{n(\bar{K}-1)}}, c_{\lambda}\!=\!0,$
        and $\Theta_{\lambda}$ is the unique root of a nonlinear equation (see \eqref{appeqn:grad_psi} in Appendix~\ref{app:proof_main_theorem}), thus the minimizer is
        \begin{align}
        B_{\lambda}^{\ast}A_{\lambda}^{\ast}\;=\;\frac{\Theta_{\lambda}}{\sqrt{\bar{K}n}}\,\left(\begin{pmatrix}M_{\bar{K}}\\ \boldsymbol 0_{(K-\bar{K})\times \bar{K}}\end{pmatrix}\otimes \boldsymbol 1_n^\top\right) X_{\mathrm{ft}}^\top.
        \end{align}

    \item \textit{Low-penalty regime} \(\bigl(0\!<\!\lambda\!\leq\!\lambda_{\mathrm{crit}}\bigr)\):
        $a_{\lambda},b_{\lambda},c_{\lambda},\Theta_{\lambda}$ are non-negative.
\end{enumerate}
Moreover, the small-regularization limit satisfies
\begin{align}
    \lim_{\lambda\to 0^{+}}\Theta_{\lambda}\!=\!\infty\,, \qquad
    \lim_{\lambda\to 0^{+}}
    \frac{B_{\lambda}^{\ast}A_{\lambda}^{\ast}}{\lVert B_{\lambda}^{\ast}A_{\lambda}^{\ast}\rVert_{F}}
    \!=\!
    \sqrt{\frac{1}{n\bar{K}}} \left(M_K^{(\bar{K})}\otimes \boldsymbol 1_n^\top \right)X_{\mathrm{ft}}^\top\,,
\end{align}
where $M_K^{(\bar{K})}$ is the first $\bar{K}$ columns of the K-ETF matrix $M_K$.
\end{thm}

\begin{proof}
We first show that one can simplify the objective in \eqref{eqn:ft_obj} based on the following Lemma.
\begin{lem}\label{applem:nc_obj}
Under Assumptions~\ref{asmp:data} and~\ref{asmp:pretrain_w} with $\epsilon_1=\epsilon_2=0$, the original fine‐tuning problem in \eqref{eqn:ft_obj} is equivalent to the following symmetric formulation:
\begin{align}\label{appeqn:obj_relax}
    \min_{B,\tilde A}\; 
    \mathcal{L}(B,\tilde A)
    &:=
    \frac{1}{\bar{K}n}\sum_{c=1}^{\bar{K}} \sum_{j=1}^n \mathcal{L}_{\mathrm{CE}}\bigl(y_c,\,B\tilde A e_{n(c-1)+j}\bigr)
    + \frac{\lambda}{2}\bigl(\|B\|_F^2 + \|\tilde A\|_F^2\bigr)\,,
\end{align}
where $X_{\mathrm{ft}} = [\bar x_{1,1},\dots,\bar x_{1,n},\cdots,\bar x_{\bar{K},1},\cdots \bar x_{\bar{K},n}], B\in\R^{K\times r}, \tilde A=A X_{\mathrm{ft}}\in\R^{r\times \bar{K}n}$.

Let \(X_{\mathrm{ft}, \perp}\) be an orthonormal complement of \(X_{\mathrm{ft}}\).  Then:
\begin{enumerate}
    \item If \((B^{\ast},A^{\ast})\) is any global minimizer of \eqref{eqn:ft_obj}, then 
    \[
    (B^{\ast},\,\tilde A^{\ast}) 
    \quad\text{with}\quad
    \tilde A^{\ast} = A^{\ast}X_{\mathrm{ft}}
    \]
    is a global minimizer of \eqref{appeqn:obj_relax}.
    
    \item Conversely, if \((B^{\ast},\tilde A^{\ast})\) is any global minimizer of \eqref{appeqn:obj_relax}, then $(B^{\ast},\tilde A^{\ast} X_{\mathrm{ft}}^\top)$
    yields a global minimizer of the original problem in \eqref{eqn:ft_obj}.
\end{enumerate}
\end{lem}
We refer the readers to Appendix~\ref{app_proof:nc_obj} for the proof of Lemma~\ref{applem:nc_obj}.
Lemma~\ref{applem:nc_obj} indicates that to characterize the global minimizer of the objective in~\eqref{eqn:ft_obj}, one can instead characterize the global minimizer of the objective in~\eqref{appeqn:obj_relax}.

We start with the following equations that characterize the global minimizer of \eqref{appeqn:obj_relax}
\begin{align*}
    \frac{\partial \mathcal{L}(B, \tilde A)}{\partial B} &=\lambda B + \frac{1}{\bar{K}n}\sum_{c=1}^{\bar{K}} \sum_{j=1}^n\frac{\partial}{\partial B\tilde A}\biggl(\mathcal{L}_{\mathrm{CE}}\left(y_c, B\tilde Ae_{(c-1)n+j}\right)\biggl)e_{(c-1)n+j}^\top \tilde A^\top = 0\\
    \frac{\partial \mathcal{L}(B, \tilde A)}{\partial \tilde A} &=\lambda \tilde A + B^\top \frac{1}{\bar{K}n}\sum_{c=1}^{\bar{K}} \sum_{j=1}^n\frac{\partial}{\partial B\tilde A}\biggl(\mathcal{L}_{\mathrm{CE}}\left(y_c, B\tilde Ae_{(c-1)n+j}\right)\biggl)e_{(c-1)n+j}^\top = 0\,.
\end{align*}
Based on the above equation, we can conclude that any minimizer $(B, \tilde A)$ of \eqref{appeqn:obj_relax} must satisfy
\begin{align}\label{eqn:balance}
    \lambda B^\top B=\lambda \tilde A \tilde A^\top \Rightarrow \|B\|_F^2 = \|\tilde A\|_F^2\,.
\end{align}
Our second step is to converge the problem into a constraint optimization problem.
The following lemma characterizes this.
\begin{lem}\label{applem:unconstraint_constraint_nonconvex}
    Let $B\in\R^{K\times r}, \tilde A\in\R^{r\times \bar{K}n}$, and assume $r\geq\bar K$ as in Assumption~\ref{asmp:lora_rank}.
    For each $\lambda \!>\! 0$, there exists a unique non-negative value $\rho_{\lambda}$ such that the solution set for the following optimization problems are equivalent
    \begin{align}
        &\min_{B,\tilde A}\; 
    \frac{1}{\bar{K}n}\sum_{c=1}^{\bar{K}} \sum_{j=1}^n \mathcal{L}_{\mathrm{CE}}\bigl(y_c,\,B\tilde A e_{(c-1)n+j}\bigr)
    + \frac{\lambda}{2}\bigl(\|B\|_F^2 + \|\tilde A\|_F^2\bigr)\,, \tag{Problem one} \label{appeqn:problem_one}\\
    &\min_{B,\tilde A}\; 
    \frac{1}{\bar{K}n}\sum_{c=1}^{\bar{K}} \sum_{j=1}^n \mathcal{L}_{\mathrm{CE}}\bigl(y_c,\,B\tilde A e_{(c-1)n+j}\bigr) \quad \text{s.t.} \|B\|_F^2\leq \rho_{\lambda}, \|\tilde A\|_F^2\leq \rho_{\lambda}\,. \tag{Problem two}\label{appeqn:problem_two}
    \end{align}
    Moreover, the map $\lambda\mapsto\rho_{\lambda}$ enjoys the following properties:
    \begin{enumerate}[label=(\roman*)]
        \item \textbf{Monotonicity and continuity.}
              $\rho_{\lambda}$ is \emph{continuous} and \emph{non-increasing} on
              $(0,\infty)$.
        \item \textbf{Asymptotic behaviour.}
              \[
                 \lim_{\lambda\to 0^{+}}\rho_{\lambda}=\infty,
                 \qquad
                 \lim_{\lambda\to\infty}\rho_{\lambda}=0.
              \]
        \item \textbf{Flat regions.}
              For any $0<\lambda_1<\lambda_2$,
              \(
                \rho(\lambda_1)=\rho(\lambda_2)
                \iff
                \rho(\lambda_1)=\rho(\lambda_2)=0.
              \)
    \end{enumerate}
\end{lem}
We refer the readers to Appendix~\ref{app_proof:unconstraint_constraint_nonconvex} for the proof. 

\myparagraph{Balanced case $\bar K\!=\!K\!\ge\!2$}
By Lemmas~\ref{applem:nc_obj} and~\ref{applem:unconstraint_constraint_nonconvex}, it suffices to solve the convex problem~\eqref{appeqn:problem_three}. Let $\ell(Z)$ denote its mean cross-entropy term and set $C\!:=\!(\boldsymbol I_K-\boldsymbol 1_K\boldsymbol 1_K^\top/K)\otimes\boldsymbol 1_n^\top$. The $K\!-\!1$ nonzero singular values of $C$ equal $\sqrt n$, so for $t\!>\!0$,
\begin{align*}
    \nabla\ell(tC)=-\frac{C}{n(e^t+K-1)}\,,\qquad
    \frac{C}{\sqrt n}\in\partial\|tC\|_*\,.
\end{align*}
Thus the global optimality condition $0\!\in\!\nabla\ell(Z)+\lambda\partial\|Z\|_*$ holds at $Z\!=\!tC$ exactly when $e^t\!+\!K\!-\!1\!=\!1/(\lambda\sqrt n)$. At zero, $\|\nabla\ell(0)\|_{\mathrm{op}}\!=\!1/(K\sqrt n)$, so zero is optimal exactly when $\lambda\!\ge\!1/(K\sqrt n)$. Uniqueness (Lemma~\ref{applem:ce_unique_solution}) therefore gives
\begin{align*}
    Z_\lambda^*=t_\lambda C\,,\qquad
    t_\lambda=
    \begin{cases}
        \log\!\bigl(\frac{1}{\lambda\sqrt n}-(K-1)\bigr), & 0<\lambda<\frac{1}{K\sqrt n},\\
        0, & \lambda\ge\frac{1}{K\sqrt n}.
    \end{cases}
\end{align*}
Since $\mathrm{rank}(Z_\lambda^*)\!\le\!K\!-\!1\!\le\!r$, this solution is feasible for the factorized problem. Writing $\bar A_\lambda^*\!:=\!A_\lambda^*X_{\mathrm{ft}}$, every global minimizer satisfies $B_\lambda^*\bar A_\lambda^*\!=\!Z_\lambda^*$ and $B_\lambda^*A_\lambda^*\!=\!t_\lambda CX_{\mathrm{ft}}^\top$. Consequently, $\rho_\lambda\!=\!\sqrt n(K-1)t_\lambda$ and $\Theta_\lambda\!=\!\sqrt{n(K-1)}t_\lambda$. These formulas prove the intermediate ETF form and both limits in~\eqref{eqn:structure_lora_low}. We set $\lambda_{\mathrm{crit}}\!=\!0$ and $\rho_{\mathrm{crit}}\!=\!+\infty$ as limiting boundaries; no minimizer at $\lambda\!=\!0$ is asserted.

For the remainder of this strict-orthogonal proof, assume $2\!\le\!\bar K\!<\!K$.
For simplicity, we will use $\rho$ to denote $\rho_{\lambda}$ for convenience.

\myparagraph{Characterization to solution of \eqref{appeqn:problem_two}} 
\eqref{appeqn:problem_two} can be viewed as a special case of the neural collapse phenomenon in the extreme setting of an imbalanced dataset: for the first $\bar{K}$ classes we observe $n$ data points per class, while all remaining classes have zero samples.  
In this context, we directly invoke the following result from Fang et al.~\cite{fang2021exploring} (see p. 26 in .~\cite{fang2021exploring}).

\begin{lem}[Neural collapse under extreme imbalance; {\cite[Page 26]{fang2021exploring}}]\label{lem:imbalanced_neural_collapse}
Assume $2\!\le\!\bar K\!<\!K$. For any constants $c_1,c_2,c_3, \rho>0$, define
\begin{align*}
    c_1' &=\frac{c_1}{c_1+(\bar{K}-1)c_2+(K-\bar{K})c_3} \\
    c_2' &= \frac{c_2}{c_1+(\bar{K}-1)c_2+(K-\bar{K})c_3} \\
    c_3' &= \frac{c_3}{c_1+(\bar{K}-1)c_2+(K-\bar{K})c_3}\\
    c_4 &=-c_1'\log c_1' - c_2'(\bar{K}-1)\log c_2' - (K-\bar{K})c_3'\log(c_3') \\
    c_5 &= \frac{\bar{K}c_2}{\bar{K}c_2+(K-\bar{K})c_3} \\
    c_6 &= \frac{(K-\bar{K})c_3}{\bar{K} c_2+(K-\bar{K})c_3}\\
    c_7 &= \frac{\bar{K}c_2+(K-\bar{K})c_3}{c_1+(\bar{K}-1)c_2+(K-\bar{K})c_3}
\end{align*}
For any feasible solution $(B, \tilde A)$ of \eqref{appeqn:problem_two}, the objective value of \eqref{appeqn:problem_two} can be simplified to
\begin{align*}
    \frac{1}{\bar{K}n}\sum_{c=1}^{\bar{K}} \sum_{j=1}^n \mathcal{L}_{\mathrm{CE}}\bigl(y_c,\,B\tilde A e_{(c-1)n+j}\bigr)
    \overset{a}{\geq} -\frac{c_7}{\bar{K}}\sqrt{\frac{\rho}{n}}\sqrt{\sum_{i=1}^{\bar{K}} \|c_5w_1+c_6 w_2-b_i\|^2}+c_4
\end{align*}
where $w_1=\frac{1}{\bar{K}}\sum_{i=1}^{\bar{K}} b_i, w_2 = \frac{1}{K-\bar{K}}\sum_{i=\bar{K}+1}^K b_i$, and $\overset{a}{\geq}$ becomes equality under certain choices of $c_1,c_2,c_3$.
\end{lem}

\noindent
Readers are referred to \cite{fang2021exploring} for the full proof.  
Note the following distinction between Lemma~\ref{lem:imbalanced_neural_collapse} and its analogue in \cite{fang2021exploring}: the inequality in \cite{fang2021exploring} becomes tight only when all minority‐class features are zero and certain choices of $c_1,c_2,c_3$, whereas in our lemma, we do not require that all minority‐class features are zero.  
This difference is resolved by our change of variables \(\tilde A = A X_{\mathrm{ft}}\), which effectively confines \(\tilde A\) to the subspace of observed (majority‐class) features and thus guarantees equality in Lemma~\ref{lem:imbalanced_neural_collapse}.

Now, we carefull pick the $B, \tilde A$ and $c_1,c_2,c_3$ to achieve the global minimum of the objective in \eqref{appeqn:problem_two}.
The following lemma characterizes this exactly.

\begin{lem}[Lemma 5 in \cite{fang2021exploring}]\label{applem:sol_step_one}
Under the assumptions of Theorem~\ref{thm:main}, with $2\!\le\!\bar K\!<\!K$, define
\begin{align*}
    \rho_{\mathrm{crit}}\!=\!\sqrt{n}\bar{K}(\bar{K}\!-\!1)\log\biggl(\sqrt{\frac{K}{K-\bar{K}}}\!+\!1\biggr)
\end{align*}
 such that the optimal value to \eqref{appeqn:problem_two} is as follows
\begin{itemize}
    \item when $\bar{K}\geq 2$ and $\rho_{\lambda} \!<\! \rho_{\mathrm{crit}}$, we choose $c_1\!=\!\exp\biggl(\frac{\rho_{\lambda}}{\bar{K}\sqrt{n}}\biggl), c_2\!=\!\exp\biggl(\frac{-\rho_{\lambda}}{\sqrt{n}\bar{K}(\bar{K}-1)}\biggl),c_3\!=\!1$, and \eqref{appeqn:problem_two} attain its minimum
    \begin{align*}
        -\frac{\rho_{\lambda}}{\sqrt{n}\bar{K}}\!+\!\log\biggl(K\!-\!\bar{K}\!+\!(\bar{K}-1)\exp\bigl(-\frac{\rho_{\lambda}}{\sqrt{n}\bar{K}(\bar{K}-1)}\bigl)\!+\!\exp\bigl(\frac{\rho_{\lambda}}{\sqrt{n}\bar{K}}\bigl)\biggl)\,,
    \end{align*}
    where $B^{\ast},\tilde A^{\ast}$ takes the following form
    \begin{align*}
      \bigl[b_1^{\ast},b_2^{\ast},\dots,b_{\bar{K}}^{\ast}\bigr]
        &= \sqrt{n}\,\bigl[\bar a_1^{\ast},\bar a_2^{\ast},\dots,\bar a_{\bar{K}}^{\ast}\bigr]
         = \sqrt{\frac{\rho}{\bar{K}}}\;P M_{\bar{K}}, \\[6pt]
      a_i^{\ast} &= \bar a_{\lceil i/n\rceil}^{\ast},
        \quad i=1,\dots ,\bar{K}n, \\
      b_j^{\ast} &= 0,
        \qquad j>\bar{K}.
    \end{align*}
    Here $P\in\R^{r\times\bar K}$ satisfies $P^\top P=\mathbf I_{\bar K}$, $a_i^\ast$ is the $i$-th column of $\tilde A^\ast$, and $\bar a_c^\ast$ is the common column for samples in class $c$.
    Writing $\bar A^{\ast}\!:=\!\tilde A^{\ast}\!=\!A^{\ast}X_{\mathrm{ft}}$ for the data-coordinate factor, the product takes the following form
    \begin{align}
        B^{\ast}\bar A^{\ast}
        \;=\;
        \frac{\rho_{\lambda}}{\sqrt{\bar{K}n(\bar{K}-1)}}\,
        \begin{pmatrix}
    M_{\bar{K}}\\
    \mathbf 0_{(K-\bar{K})\times \bar{K}}
    \end{pmatrix}
    \otimes \boldsymbol 1_n^\top\,.
    \end{align}
    \item when $\bar{K}\geq 2$ and $\rho \geq \rho_{\mathrm{crit}}$, one choose 
    \begin{align*}
        c_1=g_a(x_0), \quad c_2=g_b(x_0), \quad c_3=g_c(x_0)\,,
    \end{align*}
    where $g_a(x), g_b(x), g_c(x)$ and $x_0$ are defined in \eqref{appeqn:def_g}.
    Then, \eqref{appeqn:problem_two} attain its minimum
    \begin{align*}
        \log\biggl(\frac{g_a(x_0)+(\bar{K}-1)g_b(x_0)+(K-\bar{K})g_c(x_0)}{g_a(x_0)}\biggr)\,,
    \end{align*}
    where
    \begin{align*}
    b_i^{\ast} &\;=\;
    \begin{cases}
    \displaystyle
    g_N(x_0)\,P\!\left[
        \dfrac{\bar{K}\,x_0}{\sqrt{\bar{K}(\bar{K}-1)}}\,\bar y_i
        +\Bigl(\dfrac{1-x_0}{\sqrt{\bar{K}}}
                -\dfrac{x_0}{\sqrt{\bar{K}(\bar{K}-1)}}\Bigr)\mathbf 1_{\bar{K}}
    \right],
    & i\le \bar{K},\\[8pt]
    \displaystyle
    -\dfrac{g_N(x_0)\,\sqrt{\bar{K}}\,(1-x_0)}{K-\bar{K}}\;P\mathbf 1_{\bar{K}},
    & i>{\bar{K}}.
    \end{cases} \\
    \bar a_c^{\ast} &\;=\;
    \frac{g_N(x_0)}{\sqrt{n}} P\biggl[x_0\sqrt{\frac{\bar{K}}{\bar{K}\!-\!1}}\bar y_c
    \!+\!\bigl(\frac{(1\!-\!x_0)\sqrt{\frac{K}{K\!-\!\bar{K}}}}{\sqrt{\bar{K}}}\!-\!\frac{x_0}{\sqrt{\bar{K}(\bar{K}\!-\!1)}}\bigr)\mathbf 1_{\bar{K}}\biggr],
    \end{align*}
    where $\bar y_c\in\R^{\bar{K}}$ is a one-hot vector with the $c$-th entry equal to one (and $\bar y_i$ is defined analogously with the $i$-th entry equal to one), and $P\in\R^{r\times \bar{K}}$ is a partial orthogonal matrix such that $P^\top P=\mathbf I_{\bar{K}}$.
    The sample columns satisfy $a_{(c-1)n+j}^{\ast}=\bar a_c^{\ast}$ for $c\in[\bar K]$ and $j\in[n]$.
    Moreover, $B^{\ast}\tilde A^{\ast}$ takes the following form
    \begin{align*}
        B^{\ast}\tilde A^{\ast}=\begin{pmatrix}
            (\log g_a(x_0)-\log g_b(x_0))\mathbf I_{\bar{K}} + \log g_b(x_0)\mathbf 1_{\bar{K}}\mathbf 1_{\bar{K}}^\top \\
            \log g_c(x_0) \mathbf 1_{K-\bar{K}}\mathbf 1_{\bar{K}}^\top
        \end{pmatrix} \otimes \boldsymbol 1_n^\top\,.
    \end{align*}
\end{itemize}
\end{lem}
Lemma~\ref{applem:sol_step_one} is  Lemma 5 in \cite{fang2021exploring}. 
We refer the readers to \cite{fang2021exploring} for the proof.
Lemma~\ref{applem:sol_step_one} exactly characterize the optimal solution and value of the objective in \eqref{appeqn:problem_two}. 

\myparagraph{Computation of the product of LoRA adapters}
Based on Lemma~\ref{applem:sol_step_one}, we can characterize the solution to the LoRA objective in \eqref{eqn:ft_obj}, then it suffices to compute the product of the minimizer to prove Theorem~\ref{thm:main}.

In the \textbf{Intermediate regime}, restricting $B^{\ast}$ to its first $\bar K$ rows gives
\begin{align*}
    \bigl(B^{\ast}\bigr)_{1:\bar K,:}\,\bigl[\bar a_1^{\ast},\bar a_2^{\ast},\dots,\bar a_{\bar{K}}^{\ast}\bigr]&=\frac{\rho_{\lambda}}{\bar{K}\sqrt{n}} M_{\bar{K}}^\top P^\top P M_{\bar{K}} \\
    &=\frac{\rho_{\lambda}}{\bar{K}\sqrt{n}} \frac{\bar{K}}{\bar{K}-1} \bigl(\boldsymbol I_{\bar{K}}-\frac{1}{\bar{K}}\boldsymbol 1_{\bar{K}} \boldsymbol 1_{\bar{K}}^\top\bigr) \\
    &= \frac{\rho_{\lambda}}{\sqrt{\bar{K}n(\bar{K}-1)}} M_{\bar{K}}\,,
\end{align*}
and it leads to the product of final solution
\begin{align}
    B^{\ast}\tilde A^{\ast}
    \;=\;
    \frac{\rho_{\lambda}}{\sqrt{\bar{K}n(\bar{K}-1)}}\,
    \begin{pmatrix}
M_{\bar{K}}\\
\mathbf 0_{(K-\bar{K})\times \bar{K}}
\end{pmatrix}
\otimes \boldsymbol 1_n^\top \,.
\end{align}

In the \textbf{Low penalty regime}, in \cite{fang2021exploring}, the authors further show that 
\begin{align*}
    (b_i^{\ast})^\top \bar a_j^{\ast} = \begin{cases}
        \log g_a(x_0) &i=j\leq \bar{K}\\
        \log g_b(x_0) &i\not=j, i\leq \bar{K} \\
        \log g_c(x_0) & \bar{K}<i\leq K
    \end{cases}
    \qquad i\in[K],\ j\in[\bar K].
\end{align*}
Thus, one can show that 
\begin{align*}
    B^{\ast}\tilde A^{\ast}=\begin{pmatrix}
        (\log g_a(x_0)-\log g_b(x_0))\mathbf I_{\bar{K}} + \log g_b(x_0)\mathbf 1_{\bar{K}}\mathbf 1_{\bar{K}}^\top \\
        \log g_c(x_0) \mathbf 1_{K-\bar{K}}\mathbf 1_{\bar{K}}^\top
    \end{pmatrix} \otimes \boldsymbol 1_n^\top\,.
\end{align*}

\myparagraph{Characterization of $\lambda_{\mathrm{crit}}$} 
In Lemma~\ref{applem:sol_step_one}, when $\bar{K}\geq 2$, we see that the solution takes different form based on whether $\rho_{\lambda}\!<\!\rho_{\mathrm{crit}}$ or $\rho_{\lambda}\!\geq\!\rho_{\mathrm{crit}}$.
Moreover, based on Lemma~\ref{applem:unconstraint_constraint_nonconvex}, there must exist a unique $\lambda_{\mathrm{crit}}$ such that $\rho_{\mathrm{crit}}=\rho(\lambda_{\mathrm{crit}})$.
Now, we characterize the range of $\lambda_{\mathrm{crit}}$ here.

Due to Lemma~\ref{applem:sol_step_one}, we know when $\lambda>\lambda_{\mathrm{crit}}$, the solution takes the form
\begin{align*}
    B^{\ast}\bar A^{\ast}=\frac{\rho}{\sqrt n(\bar K-1)}\begin{pmatrix}
        \mathbf I_{\bar{K}}-\frac{1}{\bar K}\mathbf 1_{\bar K}\mathbf 1_{\bar K}^\top \\
        \mathbf 0_{(K-\bar{K})\times \bar{K}}
    \end{pmatrix} \otimes \boldsymbol 1_n^\top\,,
\end{align*}
and adding the penalty $\lambda\rho$ to this constrained minimum cross entropy gives the scalar penalized objective
\begin{align*}
    \psi(\rho):=\lambda\rho\!-\!\frac{\rho}{\sqrt{n}\bar{K}}\!+\!\log\biggl(K\!-\!\bar{K}\!+\!(\bar{K}-1)\exp\bigl(-\frac{\rho}{\sqrt{n}\bar{K}(\bar{K}-1)}\bigl)\!+\!\exp\bigl(\frac{\rho}{\sqrt{n}\bar{K}}\bigl)\biggl)\,,
\end{align*}
For this ETF-form update, $\rho=\sqrt{\bar K-1}\Theta$. Differentiating $\psi(\rho)$ and substituting this relation into the stationarity condition gives
\begin{align}\label{appeqn:grad_psi}
    &\frac{d \psi(\rho)}{d\rho} = 0\nonumber\\
    \iff& (\lambda-\frac{1}{\sqrt{n}\bar{K}})+\frac{1}{\sqrt{n}\bar{K}}\cdot\frac{\exp\bigl(\frac{\rho}{\sqrt{n}\bar{K}}\bigl)-\exp\bigl(-\frac{\rho}{\sqrt{n}\bar{K}(\bar{K}-1)}\bigl)}{K\!-\!\bar{K}\!+\!(\bar{K}-1)\exp\bigl(-\frac{\rho}{\sqrt{n}\bar{K}(\bar{K}-1)}\bigl)\!+\!\exp\bigl(\frac{\rho}{\sqrt{n}\bar{K}}\bigl)}=0 \nonumber\\
    \iff& (\lambda-\frac{1}{\sqrt{n}\bar{K}})+\frac{1}{\sqrt{n}\bar{K}}\cdot\frac{\exp\bigl(\frac{\Theta}{\sqrt{n(\bar{K}-1)}}\bigl)-1}{(K\!-\!\bar{K})\exp\bigl(\frac{\Theta}{\bar{K}\sqrt{n(\bar{K}-1)}}\bigl)\!+\!(\bar{K}-1)\!+\!\exp\bigl(\frac{\Theta}{\sqrt{n(\bar{K}-1)}}\bigl)}=0
\end{align}
We argue by contradiction. Suppose $\lambda_{\mathrm{crit}}\!\ge\!1/(\sqrt n\bar K)$. For every $\lambda\!>\!\lambda_{\mathrm{crit}}$, the preceding ETF form (including the zero product) applies and~\eqref{appeqn:grad_psi} gives $d\psi/d\rho\!>\!0$ for $\rho\!>\!0$, so $\rho_\lambda\!=\!0$. Letting $\lambda\!\downarrow\!\lambda_{\mathrm{crit}}$ and using continuity from Lemma~\ref{applem:unconstraint_constraint_nonconvex} yields $\rho(\lambda_{\mathrm{crit}})\!=\!0$, contradicting its defining value $\rho_{\mathrm{crit}}\!>\!0$. This covers both equality and strict inequality in the contradiction hypothesis. Therefore
\begin{align*}
    \lambda_{\mathrm{crit}} < \frac{1}{\sqrt{n}\bar{K}}\,.
\end{align*}

\myparagraph{Characterization of $\rho_{\lambda}$} 
In this part, we characterize the $\rho_{\lambda}$ when $\lambda \!>\!\lambda_{\mathrm{crit}}$.
By Lemma~\ref{applem:sol_step_one}, in this ETF regime $\psi(\rho)$ equals the constrained minimum cross entropy at budget $\rho$, plus the penalty $\lambda\rho$.
Thus, $\rho_{\lambda}$ must attain the minimum of $\psi(\rho)$.
Based on ~\eqref{appeqn:grad_psi}, one can first see that if $\lambda\!\geq\!\frac{1}{\sqrt{n}\bar{K}}$, $\frac{d \psi(\rho)}{d\rho}\!>\!0$, and the minimum is attained when $\rho_{\lambda}\!=\!0$.
When $\lambda_{\mathrm{crit}}<\lambda<\frac{1}{\sqrt{n}\bar{K}}$, solve~\eqref{appeqn:grad_psi} for $\Theta_\lambda$ and recover $\rho_\lambda=\sqrt{\bar K-1}\Theta_\lambda$.

\myparagraph{Asymptotic behaviour of $\rho_{\lambda}$}
On one hand, as $\lambda\!\to\! 0+$, one obviously has $\lim_{\lambda\to0+}\|B^{\ast}\tilde A^{\ast}\|_F=\infty$, since by Lemma~\ref{applem:unconstraint_constraint_nonconvex} we have $Z_{\lambda}^{\ast}=B^{\ast}\tilde A^{\ast}$ where $Z_{\lambda}^{\ast}$ is the unique minimizer of \eqref{appeqn:problem_three} and $\|Z_{\lambda}^{\ast}\|_*=\rho_{\lambda}\to\infty$. Then, we characterize its asymptotic direction, which is equivalent to study the limit of $\frac{\log g_a(x_0)}{\log g_b(x_0)}$ and $\frac{\log g_a(x_0)}{\log g_c(x_0)}$ as $\lambda\to 0+$. We will use $x_{0,\lambda}$ denote the choice of $x_0$ for fix regularization effect $\lambda$.

First, the equation to derive for $x_0$ is as follows
\begin{align*}
    &\frac{g_b(x)}{g_c(x)}=\frac{1}{1+\sqrt{\frac{K}{K-\bar{K}}}} \\
    \iff & \frac{\rho_{\lambda}}{\bar{K}\sqrt{n}}\cdot\frac{-\frac{x^2}{\bar{K}-1}+(\frac{K}{K-\bar{K}})^{3/2}(1-x)^2}{x^2+\frac{K}{K-\bar{K}}(1-x)^2} = -\log\bigl(1+\sqrt{\frac{K}{K-\bar{K}}}\bigr)
\end{align*}
Since $\lim_{\lambda\to 0+}\rho_{\lambda}=\infty$, one must have $x_{0,\lambda}$ approaches the solution of the following quadratic equation
\begin{align*}
    \frac{x_{0,\lambda}^2}{\bar{K}-1}=\bigl(\frac{K}{K-\bar{K}}\bigr)^{3/2}(1-x_{0,\lambda})^2\,.
\end{align*}
Moreover, one can see that the expression for $\frac{\log g_a(x_{0,\lambda})}{\log g_b(x_{0,\lambda})}$ and $\frac{\log g_a(x_{0,\lambda})}{\log g_c(x_{0,\lambda})}$ takes the following form
\begin{align*}
    \frac{\log g_a(x_{0,\lambda})}{\log g_b(x_{0,\lambda})} &= \frac{x_{0,\lambda}^2 + \sqrt{\frac{K}{K-\bar{K}}}(1-x_{0,\lambda})^2}{-\frac{x_{0,\lambda}^2}{\bar{K}-1}+\sqrt{\frac{K}{K-\bar{K}}}(1-x_{0,\lambda})^2} \\
    \frac{\log g_a(x_{0,\lambda})}{\log g_c(x_{0,\lambda})} &= \frac{x_{0,\lambda}^2 + \sqrt{\frac{K}{K-\bar{K}}}(1-x_{0,\lambda})^2}{-\frac{\bar{K}}{K-\bar{K}}\cdot\sqrt{\frac{K}{K-\bar{K}}}(1-x_{0,\lambda})^2}\,.
\end{align*}
Thus, combine the above two equations, we can compute that
\begin{align*}
    \lim_{\lambda\to0+}\frac{\log g_a(x_{0,\lambda})}{\log g_b(x_{0,\lambda})} = \lim_{\lambda\to0+}\frac{\log g_a(x_{0,\lambda})}{\log g_c(x_{0,\lambda})}= -(K-1)\,.
\end{align*}

\myparagraph{Connection between $\rho_{\lambda}$ and $\Theta_{\lambda}$} In previous analysis, we have shown that 
\begin{align}
    \rho_{\lambda}=\|B_{\lambda}^{\ast}A_{\lambda}^{\ast}\|_{\ast}\,,\qquad \Theta_{\lambda}=\|B_{\lambda}^{\ast}A_{\lambda}^{\ast}\|_{F}
\end{align}
It remains to relate $\rho_{\lambda}$ and $\Theta_{\lambda}$ across the regularization regimes.
Write $\bar A_{\lambda}^{\ast}\!:=\!A_{\lambda}^{\ast}X_{\mathrm{ft}}$ for the data-coordinate factor denoted by $\tilde A^{\ast}$ above.
By Lemma~\ref{applem:nc_obj}, $B_{\lambda}^{\ast}A_{\lambda}^{\ast}\!=\!B_{\lambda}^{\ast}\bar A_{\lambda}^{\ast}X_{\mathrm{ft}}^\top$.
Since $X_{\mathrm{ft}}^\top X_{\mathrm{ft}}\!=\!\boldsymbol I_{\bar Kn}$, the weight-space and data-coordinate products have the same nonzero singular values, and hence the same nuclear and Frobenius norms.

First, it is obvious in the high-penalty regime, $\rho_{\lambda}=\Theta_{\lambda}=0$.

Second, in the \textit{intermediate regime}, the data-coordinate product is
\begin{align*}
    B_{\lambda}^{\ast}\bar A_{\lambda}^{\ast}
    \;=\;
    \frac{\rho_{\lambda}}{\sqrt{\bar{K}n(\bar{K}-1)}}\,
    \begin{pmatrix}
M_{\bar{K}}\\
\mathbf 0_{(K-\bar{K})\times \bar{K}}
\end{pmatrix}
\otimes \boldsymbol 1_n^\top \,.
\end{align*}
Consequently,
\begin{align*}
    \rho_{\lambda}=\|B_{\lambda}^{\ast}\bar A_{\lambda}^{\ast}\|_{\ast}\,,\qquad \Theta_{\lambda}=\|B_{\lambda}^{\ast}\bar A_{\lambda}^{\ast}\|_{F}=\frac{\rho_{\lambda}}{\sqrt{\bar K-1}}\,.
\end{align*}
Equivalently, the data-coordinate product in the \textit{intermediate regime} is
\begin{align*}
    B_{\lambda}^{\ast}\bar A_{\lambda}^{\ast}
    \;=\;
    \frac{\Theta_{\lambda}}{\sqrt{n\bar K}}\,
    \begin{pmatrix}
M_{\bar{K}}\\
\mathbf 0_{(K-\bar{K})\times \bar{K}}
\end{pmatrix}
\otimes \boldsymbol 1_n^\top \,.
\end{align*}
Multiplying by $X_{\mathrm{ft}}^\top$ recovers the weight-space formula~\eqref{eqn:structure_lora_intermediate}.
Third, in the \textit{low-penalty regime}, the preceding asymptotic analysis gives $\lim_{\lambda\to0^+}\|B_{\lambda}^{\ast}\bar A_{\lambda}^{\ast}\|_F\!=\!\infty$ and characterizes the limiting direction.

\myparagraph{The solution of \eqref{eqn:ft_obj}} Finally, one can use Lemma~\ref{applem:nc_obj} to characterize the optimal solution of \eqref{eqn:ft_obj} based on the above results.

\end{proof}

\subsection{Proof of General Version}\label{appsec:general_version}
In this subsection we present the general version of Theorem~\ref{thm:main}: under the near-orthogonal regime in Assumption~\ref{asmp:data} (i.e., $\epsilon_1,\epsilon_2$ small but possibly nonzero), the optimal LoRA update $B_\lambda^{\ast}A_\lambda^{\ast}$ remains close, in Frobenius norm, to its closed-form orthogonal-case description, with the deviation controlled linearly in $\epsilon_1$ and $\epsilon_2$.

\myparagraph{Notation} Write $G:=X_{\mathrm{ft}}^\top X_{\mathrm{ft}}\in\R^{\bar Kn\times\bar Kn}$, $R:=G^{1/2}$, $Q:=X_{\mathrm{ft}}R^{-1}\in\R^{d\times\bar Kn}$ (so that $Q^\top Q=I_{\bar Kn}$ and $X_{\mathrm{ft}}=QR$), $S:=W_{\mathrm{pre}}X_{\mathrm{ft}}\in\R^{K\times\bar Kn}$, and let $N_{\mathrm{pre}}:=|\mathcal D_{\mathrm{pre}}|$ be the number of pre-training samples. Set $\delta:=(\bar Kn-1)\,\epsilon_1$. For any $\bar S\in\R^{K\times\bar Kn}$ and any positive-definite $\bar R\in\R^{\bar Kn\times\bar Kn}$, define
\begin{align}
    \Phi(Z;\bar S,\bar R)&:=\frac{1}{\bar Kn}\sum_{c=1}^{\bar K}\sum_{j=1}^n\mathcal{L}_{\mathrm{CE}}\bigl(y_c,(Z\bar R+\bar S)\,e_{(c-1)n+j}\bigr)+\lambda\|Z\|_*\,,\\
    Z_\lambda^{\ast}(\bar S,\bar R)&:=\arg\min_{Z\in\R^{K\times\bar Kn}}\Phi(Z;\bar S,\bar R)\,.
\end{align}
The minimizer $Z_\lambda^{\ast}$ is unique by Lemma~\ref{applem:ce_unique_solution} (for $\bar S=0,\bar R=I$; the same argument applies for general $\bar R\succ 0$).

\begin{thm}[General version of Theorem~\ref{thm:main}]\label{appthm:general}
Fix $\lambda\!>\!0$ and suppose Assumptions~\ref{asmp:data}--\ref{asmp:pretrain_w} hold with $\epsilon_1,\epsilon_2\!\in\![0,1)$. There exist constants $\delta_0\!\in\!(0,1/2]$, $s_0\!\in\!(0,1]$, and $\mu_0,C_0\!>\!0$, depending on $\lambda,K,\bar K,n,\|Z_\lambda^{\ast}(0,I)\|_*$ but not on $\epsilon_1,\epsilon_2$, such that whenever $\delta\!:=\!(\bar Kn-1)\,\epsilon_1\!\le\!\delta_0$ and $\|S\|_F\!\le\!s_0$,
\begin{align}\label{appeqn:gen_main_bound}
    \bigl\|B_\lambda^{\ast}A_\lambda^{\ast}-Z_\lambda^{\ast}(0,I)\,X_{\mathrm{ft}}^\top\bigr\|_F\le\frac{C_0}{\mu_0}\,\Bigl(\|S\|_F+\delta\,\bigl(1+\|Z_\lambda^{\ast}(0,I)\|_F\bigr)\Bigr)\,.
\end{align}
In particular, as $\epsilon_1,\epsilon_2\!\to\! 0$, $B_\lambda^{\ast}A_\lambda^{\ast}$ converges in Frobenius norm to its orthogonal-case form $Z_\lambda^{\ast}(0,I)X_{\mathrm{ft}}^\top$.
\end{thm}

\begin{rem}[Bounding $\|S\|_F$ in $\epsilon_2$]\label{apprem:S_bound}
The bound \eqref{appeqn:gen_main_bound} is stated directly in terms of $\|S\|_F=\|W_{\mathrm{pre}}X_{\mathrm{ft}}\|_F$. To translate this into a bound in $\epsilon_2$, one needs a lower bound on $\sigma_{\min}(X_{\mathrm{pre}})$. By Proposition~\ref{prop:max_margin} and Assumption~\ref{asmp:pretrain_w}, each row $w_k$ of $W_{\mathrm{pre}}$ admits the representation $w_k=X_{\mathrm{pre}}\beta_k$ with $\|\beta_k\|_2\le\|w_k\|_2/\sigma_{\min}(X_{\mathrm{pre}})$. Combined with Assumption~\ref{asmp:data}(c),
\begin{align}
    |w_k^\top\bar x_{c,j}|=|\beta_k^\top X_{\mathrm{pre}}^\top\bar x_{c,j}|\le\|\beta_k\|_2\,\sqrt{N_{\mathrm{pre}}}\,\epsilon_2\le\frac{\|w_k\|_2\,\sqrt{N_{\mathrm{pre}}}\,\epsilon_2}{\sigma_{\min}(X_{\mathrm{pre}})}\,,
\end{align}
hence $\|S\|_F\le\sqrt{\bar Kn\,N_{\mathrm{pre}}}\,\|W_{\mathrm{pre}}\|_F\,\epsilon_2/\sigma_{\min}(X_{\mathrm{pre}})$. In particular, $\|S\|_F\to 0$ as $\epsilon_2\to 0$ whenever $\sigma_{\min}(X_{\mathrm{pre}})$ is bounded below by a positive constant. The cleanest specialization is the orthonormal case $X_{\mathrm{pre}}^\top X_{\mathrm{pre}}=I_{N_{\mathrm{pre}}}$ (so $\sigma_{\min}(X_{\mathrm{pre}})=1$), which yields
\begin{align}\label{eqn:upper_bound_S}
    \|S\|_F\le\sqrt{\bar Kn\,N_{\mathrm{pre}}}\,\|W_{\mathrm{pre}}\|_F\,\epsilon_2\,.
\end{align}
The same conclusion holds, up to a multiplicative conditioning constant, whenever $\sigma_{\min}(X_{\mathrm{pre}})$ can be lower-bounded (e.g., near-orthogonal pre-training features, or unit-norm pre-training features with bounded condition number).
We highlight that we do not claim \eqref{eqn:upper_bound_S} is tight. The goal of Remark~\ref{apprem:S_bound} is to show that $\|S\|_F$ can be controlled by $\epsilon_2$ under mild assumptions, so that the right-hand side of \eqref{appeqn:gen_main_bound} goes to zero as $\epsilon_2\to 0$.
\end{rem}

\begin{proof}[Proof of Theorem~\ref{appthm:general}]
We first reformulate the optimization problem in $Z$ as it was done in the previous proof.
By the same change of variable used in the proof of Lemma~\ref{applem:nc_obj}, decompose $A=\tilde AQ^\top+A_\perp$ where $\tilde A:=AQ\in\R^{r\times\bar Kn}$ and $A_\perp:=A(I_d-QQ^\top)$. Then $Ax_{c,j}=\tilde ARe_{(c-1)n+j}$ and \eqref{eqn:ft_obj} can be rewritten as
\begin{align}
    \mathcal{L}(B,A)=\frac{1}{\bar Kn}\sum_{c,j}\mathcal{L}_{\mathrm{CE}}\bigl(y_c,(B\tilde AR+S)e_{(c-1)n+j}\bigr)+\frac{\lambda}{2}\bigl(\|B\|_F^2+\|\tilde A\|_F^2+\|A_\perp\|_F^2\bigr)\,.
\end{align}
The component $A_\perp$ contributes only to the regularizer and is therefore zero at any minimizer. Setting $Z=B\tilde A$ and applying the variational form of the nuclear norm (Lemma~\ref{applem:varitional_nuclear}), the LoRA problem becomes the rank-constrained minimization $\min_{Z\in\R^{K\times \bar Kn},\,\mathrm{rank}(Z)\le r}\Phi(Z;S,R)$, where $\Phi(Z;\bar S,\bar R):=L(Z;\bar S,\bar R)+\lambda\|Z\|_*$. Let $Z_\lambda^{\ast}(S,R)$ denote the unique unconstrained minimizer of $\Phi(\cdot;S,R)$ over $\R^{K\times \bar Kn}$ (Lemma~\ref{applem:ce_unique_solution}). We first establish a perturbation bound for this unconstrained problem, then verify that its minimizer satisfies the rank constraint.

Then, we provide a spectral bound on $R$. By Assumption~\ref{asmp:data}, $G_{ii}=\|\bar x_i\|^2=1$ and $|G_{ij}|=|\bar x_i^\top\bar x_j|\le\epsilon_1$ for $i\ne j$. By Gershgorin's circle theorem,
\begin{align}
    \|G-I\|_{\mathrm{op}}\le\max_{i}\sum_{j\ne i}|G_{ij}|\le(\bar Kn-1)\,\epsilon_1=\delta\,.
\end{align}
Hence every eigenvalue of $G$ lies in $[1-\delta,1+\delta]$, so $\sigma_i(R)=\sqrt{\sigma_i(G)}\in[\sqrt{1-\delta},\sqrt{1+\delta}]$, which gives $\sigma_{\min}(R)\ge\sqrt{1-\delta}$. Using $|\sqrt{x}-1|\le|x-1|$ for $x\ge0$,
\begin{align}
    \|R-I\|_{\mathrm{op}}=\max_{i}|\sqrt{\sigma_i(G)}-1|\le\max_{i}|\sigma_i(G)-1|=\|G-I\|_{\mathrm{op}}\le\delta\,.
\end{align}

Based on the previous spectral bound, we can now bound the deviation of $Z_\lambda^{\ast}(S,R)$ from $Z_\lambda^{\ast}(0,I)$, which will lead to the desired bound on $\|B_\lambda^{\ast}A_\lambda^{\ast}-Z_\lambda^{\ast}(0,I)X_{\mathrm{ft}}^\top\|_F$.
Let $L(Z;\bar S,\bar R):=\Phi(Z;\bar S,\bar R)-\lambda\|Z\|_*$ be the smooth (cross-entropy) part of the objective, and write $Z_1:=Z_\lambda^{\ast}(S,R)$, $Z_0:=Z_\lambda^{\ast}(0,I)$. The first-order optimality conditions read
\begin{align}
    -\nabla_Z L(Z_1;S,R)\in\lambda\,\partial\|Z_1\|_*\,,\qquad-\nabla_Z L(Z_0;0,I)\in\lambda\,\partial\|Z_0\|_*\,.
\end{align}
Subdifferentials of convex functions are monotone, so $\langle G_1-G_0,\,Z_1-Z_0\rangle\ge0$ for any $G_i\in\partial\|Z_i\|_*$. Substituting,
\begin{align}\label{appeqn:gen_mono}
    \langle\nabla L(Z_1;S,R)-\nabla L(Z_0;0,I),\,Z_1-Z_0\rangle\le 0\,.
\end{align}

We work under the preliminary bound $\|S\|_F\!\le\!1$ and will choose $s_0\!\le\!1$ below. By Lemma~\ref{applem:ce_unique_solution}, both $Z_0$ and $Z_1$ lie in the subspace $\{Z:\mathbf 1_K^\top Z=0\}$. We next establish an a priori F-norm bound to enable application of local strong convexity. The unperturbed bound $\|Z_0\|_*\le\log K/\lambda$ is already given in the proof of Lemma~\ref{applem:nc_obj} (using $\Phi(0;0,I)=\log K$). The same argument extends to the perturbed problem: by optimality of $Z_1$ and feasibility of $Z=0$,
\begin{align}\label{appeqn:gen_apriori}
    \lambda\,\|Z_1\|_F\le\lambda\,\|Z_1\|_*\le\Phi(Z_1;S,R)\le\Phi(0;S,R)=L(0;S,R)\le\log K+2\|S\|_F\,,
\end{align}
where the last inequality uses $\mathcal{L}_{\mathrm{CE}}(y_c,z)=-z_c+\log\sum_i e^{z_i}\le\log K+2\|z\|_\infty$ together with $\|Se_{(c-1)n+j}\|_\infty\le\|S\|_F$. Define
\begin{align*}
    R_0:=\frac{\log K+2}{\lambda}\,,\qquad \mathcal B:=\bigl\{Z\in\R^{K\times \bar Kn}:\|Z\|_F\le R_0,\,\mathbf 1_K^\top Z=0\bigr\}\,,
\end{align*}
so that $\{Z_0,Z_1\}\subset\mathcal B$. By Lemma~\ref{applem:ce_strong_convex} applied with radius $R_0$, $L(\cdot;0,I)$ is $\mu_0$-strongly convex on $\mathcal B$ in directions $\Delta_Z\in\{\Delta_Z:\mathbf 1_K^\top\Delta_Z=0\}$, where $\mu_0=\mu_0(R_0)>0$. Consequently
\begin{align}
    \langle\nabla L(Z_1;0,I)-\nabla L(Z_0;0,I),\,Z_1-Z_0\rangle\ge\mu_0\|Z_1-Z_0\|_F^2\,.
\end{align}
Subtracting from \eqref{appeqn:gen_mono},
\begin{align}
    \mu_0\|Z_1-Z_0\|_F^2&\le\langle\nabla L(Z_1;0,I)-\nabla L(Z_1;S,R),\,Z_1-Z_0\rangle\\
    &\le\|\nabla L(Z_1;0,I)-\nabla L(Z_1;S,R)\|_F\,\|Z_1-Z_0\|_F\,,
\end{align}
hence
\begin{align}\label{appeqn:gen_grad}
    \mu_0\,\|Z_1-Z_0\|_F\le\|\nabla L(Z_1;0,I)-\nabla L(Z_1;S,R)\|_F\,.
\end{align}

It remains to bound the right-hand side of \eqref{appeqn:gen_grad}. Direct calculation gives
\begin{align}
    \nabla L(Z;\bar S,\bar R)=\frac{1}{\bar Kn}\sum_{c,j}\sigma_{c,j}(Z;\bar S,\bar R)\,(\bar R\,e_{(c-1)n+j})^\top\,,
\end{align}
where $\sigma_{c,j}(Z;\bar S,\bar R):=\mathrm{softmax}\bigl((Z\bar R+\bar S)e_{(c-1)n+j}\bigr)-y_c\in\R^K$. Splitting,
\begin{align}
    \nabla L(Z;S,R)-\nabla L(Z;0,I)=\frac{1}{\bar Kn}\sum_{c,j}\Bigl[&\bigl(\sigma_{c,j}(Z;S,R)-\sigma_{c,j}(Z;0,I)\bigr)\,(R\,e_{(c-1)n+j})^\top\\
    &\quad+\sigma_{c,j}(Z;0,I)\,((R-I)\,e_{(c-1)n+j})^\top\Bigr]\,.
\end{align}
The softmax map is $1$-Lipschitz, so
\begin{align}
    \|\sigma_{c,j}(Z;S,R)-\sigma_{c,j}(Z;0,I)\|_2&\le\|((ZR+S)-Z)e_{(c-1)n+j}\|_2 \nonumber\\
    &\le\|Z\|_F\,\|R-I\|_{\mathrm{op}}+\|S\,e_{(c-1)n+j}\|_2\,,
\end{align}
and $\|\sigma_{c,j}(Z;0,I)\|_2\le\sqrt{2}$ since $\sigma_{c,j}$ is the difference of a probability vector and a one-hot label, both in $[0,1]^K$. Combining via the triangle inequality and Cauchy--Schwarz (using $\sum_{c,j}\|Se_{(c-1)n+j}\|_2\le\sqrt{\bar Kn}\,\|S\|_F$ and $\sum_{c,j}\|(R-I)e_{(c-1)n+j}\|_2\le\sqrt{\bar Kn}\,\|R-I\|_F\le\bar Kn\,\delta$),
\begin{align}
    \|\nabla L(Z;S,R)-\nabla L(Z;0,I)\|_F\le(1+\delta)\Bigl(\delta\|Z\|_F+\frac{\|S\|_F}{\sqrt{\bar Kn}}\Bigr)+\sqrt{2}\,\delta\,.
\end{align}
Evaluating at $Z=Z_1$, using $\|Z_1\|_F\le\|Z_0\|_F+\|Z_1-Z_0\|_F$, and substituting into \eqref{appeqn:gen_grad}:
\begin{align}
    \bigl(\mu_0-(1+\delta)\delta\bigr)\,\|Z_1-Z_0\|_F\le(1+\delta)\Bigl(\delta\,\|Z_0\|_F+\frac{\|S\|_F}{\sqrt{\bar Kn}}\Bigr)+\sqrt{2}\,\delta\,.
\end{align}
Set $\eta_0\!:=\!\min(1/2,\mu_0/3)\!\in\!(0,1/2]$. For any $\delta\!\le\!\eta_0$ we have $1+\delta\le 3/2$ and
\begin{align*}
    (1+\delta)\delta\le\tfrac{3}{2}\delta\le\tfrac{3}{2}\cdot\tfrac{\mu_0}{3}=\tfrac{\mu_0}{2}\,,
\end{align*}
so $\mu_0-(1+\delta)\delta\ge\mu_0/2>0$. Dividing through and using $1+\delta\le 3/2$, $\sqrt{2}\delta\le\sqrt{2}\delta\,(1+\|Z_0\|_F)$,
\begin{align}\label{appeqn:gen_zbound}
    \|Z_1-Z_0\|_F\le\frac{C_1}{\mu_0}\,\Bigl(\|S\|_F+\delta\,\bigl(1+\|Z_0\|_F\bigr)\Bigr)\,,
\end{align}
where $C_1$ is an absolute constant absorbing the factors $2,\,3/\sqrt{\bar Kn},\,\sqrt{2}$ from the previous display.

By Lemma~\ref{applem:rank_stability} below, there exist thresholds $\delta_0'\!\le\!\eta_0$ and $s_0\!\le\!1$ such that $\mathrm{rank}(Z_1)\!\le\!\bar K\!\le\!r$ whenever $\delta\!\le\!\delta_0'$ and $\|S\|_F\!\le\!s_0$. Choose the theorem's threshold $\delta_0\!:=\!\delta_0'$. Under the joint conditions in the statement, the rank constraint is therefore non-binding, so $B^{\ast}\tilde A^{\ast}\!=\!Z_1$ and $B^{\ast}A^{\ast}\!=\!Z_1Q^\top$.

Finally, we bound $\|B_\lambda^{\ast}A_\lambda^{\ast}-Z_0\,X_{\mathrm{ft}}^\top\|_F$.
From the reformulation above, $B_\lambda^{\ast}A_\lambda^{\ast}=Z_1\,Q^\top$. Using $X_{\mathrm{ft}}=QR$, hence $X_{\mathrm{ft}}^\top=RQ^\top$, we obtain $Z_0\,X_{\mathrm{ft}}^\top=Z_0\,R\,Q^\top$, so
\begin{align}
    \bigl\|B_\lambda^{\ast}A_\lambda^{\ast}-Z_0\,X_{\mathrm{ft}}^\top\bigr\|_F
    &=\bigl\|(Z_1-Z_0R)\,Q^\top\bigr\|_F \nonumber\\
    &=\bigl\|Z_1-Z_0R\bigr\|_F&&(\text{since }Q^\top Q=I) \nonumber\\
    &\le\|Z_1-Z_0\|_F+\bigl\|Z_0(I-R)\bigr\|_F \nonumber\\
    &\le\|Z_1-Z_0\|_F+\delta\,\|Z_0\|_F\,.
\end{align}
Combining with \eqref{appeqn:gen_zbound} and using $\delta\,\|Z_0\|_F\le\delta\,(1+\|Z_0\|_F)$,
\begin{align}
    \bigl\|B_\lambda^{\ast}A_\lambda^{\ast}-Z_0\,X_{\mathrm{ft}}^\top\bigr\|_F
    &\le\frac{C_1}{\mu_0}\bigl(\|S\|_F+\delta\,(1+\|Z_0\|_F)\bigr)+\delta\,(1+\|Z_0\|_F)\\
    &\le\frac{C_1+\mu_0}{\mu_0}\bigl(\|S\|_F+\delta\,(1+\|Z_0\|_F)\bigr)\,.
\end{align}
Setting $C_0:=C_1+\mu_0$ (which depends only on $\lambda,K,\bar K,n,\|Z_0\|_*$, as required) gives \eqref{appeqn:gen_main_bound}.
\end{proof}

\begin{lem}[Rank stability of the unconstrained minimizer]\label{applem:rank_stability}
Adopt the notation from the proof of Theorem~\ref{appthm:general}. Let $Z_\lambda^{\ast}(\bar S,\bar R)$ denote the unique unconstrained minimizer of $\Phi(\cdot;\bar S,\bar R)=L(\cdot;\bar S,\bar R)+\lambda\|\cdot\|_*$ over $\R^{K\times \bar Kn}$ (Lemma~\ref{applem:ce_unique_solution}), and write $Z_0:=Z_\lambda^{\ast}(0,I)$. There exist constants $\delta_0'\!\in\!(0,\eta_0]$ and $s_0\!\in\!(0,1]$, depending on $\lambda,K,\bar K,n,\mu_0,C_1,\|Z_0\|_*$ but not on $\epsilon_1,\epsilon_2$, such that whenever $\delta\le\delta_0'$ and $\|S\|_F\le s_0$,
\begin{align*}
    \mathrm{rank}\bigl(Z_\lambda^{\ast}(S,R)\bigr)\le\bar K\,.
\end{align*}
\end{lem}
\begin{proof}
Write $Z_1:=Z_\lambda^{\ast}(S,R)$ and $G_i:=-\nabla L(Z_i;\bar S_i,\bar R_i)/\lambda$ for $i\in\{0,1\}$, with $(\bar S_0,\bar R_0)=(0,I)$ and $(\bar S_1,\bar R_1)=(S,R)$.

By the within-class permutation argument in the proof of Lemma~\ref{applem:unconstraint_constraint_nonconvex}, $Z_0=V\otimes\mathbf1_n^\top$ for some $V\in\R^{K\times\bar K}$.

Consequently, $\sigma(Z_0\,e_{(c-1)n+j})=\sigma(V_{:,c})=:p_c$ depends only on $c$. Letting $D:=\sum_{c=1}^{\bar K}(p_c-y_c)\,e_c^\top\in\R^{K\times \bar K}$,
\begin{align}\label{appeqn:rank_G0_struct}
    \nabla L(Z_0;0,I)=\frac{1}{\bar Kn}\sum_{c=1}^{\bar K}(p_c-y_c)\sum_{j=1}^n e_{(c-1)n+j}^\top=\frac{1}{\bar Kn}\,D\otimes \mathbf 1_n^\top\,,
\end{align}
so $\mathrm{rank}\bigl(\nabla L(Z_0;0,I)\bigr)\le\mathrm{rank}(D)\le\bar K$, and
\begin{align}\label{appeqn:rank_G0}
    \sigma_i(G_0)=0 \quad\text{for all } i>\bar K\,.
\end{align}

Then, we study the first-order optimality condition for the unconstrained convex problem reads $G_1\in\partial\|Z_1\|_*$. Letting $Z_1=U_1\Sigma_1V_1^\top$ be a compact SVD of rank $r_1:=\mathrm{rank}(Z_1)$, the standard subdifferential characterization of the nuclear norm \citep{watson1992characterization} gives $G_1=U_1V_1^\top+W$ with $U_1^\top W=0$, $WV_1=0$, and $\|W\|_{\mathrm{op}}\le 1$. Since $U_1V_1^\top$ and $W$ have orthogonal column and row spaces, their singular values combine: $\sigma_i(G_1)=1$ for $i\le r_1$, and $\sigma_i(G_1)=\sigma_{i-r_1}(W)\le 1$ for $i>r_1$. Applying Weyl's inequality for singular values to $G_1$ and $G_0$, together with \eqref{appeqn:rank_G0},

\begin{align}\label{appeqn:rank_weyl}
    \sigma_{\bar K+1}(G_1)\le\sigma_{\bar K+1}(G_0)+\|G_1-G_0\|_{\mathrm{op}}=\|G_1-G_0\|_{\mathrm{op}}\le\|G_1-G_0\|_F\,.
\end{align}

Next, based on the previous analysis, we bound the difference between the gradients.
Decompose
\begin{align*}
    \nabla L(Z_1;S,R)-\nabla L(Z_0;0,I)=\underbrace{\nabla L(Z_1;S,R)-\nabla L(Z_1;0,I)}_{=:(\mathrm{I})}+\underbrace{\nabla L(Z_1;0,I)-\nabla L(Z_0;0,I)}_{=:(\mathrm{II})}\,.
\end{align*}
The bound \eqref{appeqn:rank_grad_I_eq} on $(\mathrm{I})$ follows from the gradient calculation in the proof of Theorem~\ref{appthm:general} (the display preceding \eqref{appeqn:gen_zbound}): for $\delta\!\le\!\eta_0$,
\begin{align}\label{appeqn:rank_grad_I_eq}
    \|(\mathrm{I})\|_F\le(1+\delta)\Bigl(\delta\|Z_1\|_F+\frac{\|S\|_F}{\sqrt{\bar Kn}}\Bigr)+\sqrt 2\,\delta\,.
\end{align}
For $(\mathrm{II})$, the softmax map $\sigma:\R^K\!\to\!\Delta^{K-1}$ has Jacobian $\mathrm{diag}(p)-pp^\top$ with operator norm at most $1$, and is therefore $1$-Lipschitz in $\ell_2$. Since $\{e_{(c-1)n+j}\}_{c,j}$ is an orthonormal basis of $\R^{\bar Kn}$,
\begin{align}\label{appeqn:rank_grad_II_eq}
    \|(\mathrm{II})\|_F^2=\frac{1}{(\bar Kn)^2}\sum_{c,j}\bigl\|\sigma(Z_1 e_{(c-1)n+j})-\sigma(Z_0 e_{(c-1)n+j})\bigr\|_2^2\le\frac{\|Z_1-Z_0\|_F^2}{(\bar Kn)^2}\,.
\end{align}
By \eqref{appeqn:gen_zbound}, $\|Z_1-Z_0\|_F\le\frac{C_1}{\mu_0}\bigl(\|S\|_F+\delta(1+\|Z_0\|_F)\bigr)$. Combined with $\|Z_1\|_F\le\|Z_0\|_F+\|Z_1-Z_0\|_F$ and restricting to $\delta\!\le\!\eta_0$ and $\|S\|_F\le 1$, there exists a constant
\begin{align*}
    C_2:=(1+\eta_0)\Bigl(\|Z_0\|_F+\tfrac{C_1}{\mu_0}(1+\|Z_0\|_F)+\tfrac{1}{\sqrt{\bar Kn}}\Bigr)+\sqrt 2+\tfrac{C_1(1+\|Z_0\|_F)}{\mu_0\bar Kn}\,,
\end{align*}
depending only on $\lambda,K,\bar K,n,\mu_0,C_1,\|Z_0\|_*$, such that
\begin{align}\label{appeqn:rank_grad_total}
    \|\nabla L(Z_1;S,R)-\nabla L(Z_0;0,I)\|_F\le C_2\,(\delta+\|S\|_F)\,.
\end{align}

Finally, we use the above bounds to establish a condition under which the rank of $Z_1$ is bounded. Set
\begin{align*}
    \delta_0':=\min\!\bigl(\eta_0,\,\tfrac{\lambda}{4C_2}\bigr)\,,\qquad s_0:=\min\!\bigl(1,\,\tfrac{\lambda}{4C_2}\bigr)\,.
\end{align*}
For $\delta\le\delta_0'$ and $\|S\|_F\le s_0$,
\begin{align*}
    \|G_1-G_0\|_F=\frac{1}{\lambda}\,\|\nabla L(Z_1;S,R)-\nabla L(Z_0;0,I)\|_F\le\frac{C_2(\delta_0'+s_0)}{\lambda}\le\frac{1}{2}<1\,.
\end{align*}
Combined with \eqref{appeqn:rank_weyl}, $\sigma_{\bar K+1}(G_1)<1$. Since at most $\bar K$ singular values of $G_1$ can equal $1$ by Step~2, we conclude $\mathrm{rank}(Z_1)\le\bar K$.
\end{proof}

\section{Proof of Theorem~\ref{thm:lora_forget_less} and Corollary~\ref{cor:optimal_lambda_tradeoff}}\label{appsec:proof_forget}
In this section, we first present the proofs of Theorem~\ref{thm:lora_forget_less} and Corollary~\ref{cor:optimal_lambda_tradeoff}.
Then, under additional smallness assumptions on $\epsilon_1,\epsilon_2$, we show that the same conclusions still hold, with additional error terms that depend on $\epsilon_1,\epsilon_2$.

\subsection{Proof of Theorem~\ref{thm:lora_forget_less}}
For convenience, we will first define the following notations.
Let $W_{\mathrm{LoRA}}^{\lambda}= W_{\mathrm{pre}}+B_{\lambda}^{\ast} A_{\lambda}^{\ast}$, and define the following shorthand for margins as:
$\gamma_{\mathrm{pre}}=\gamma(W_{\mathrm{pre}};\mathcal D_{\mathrm{pre}})$,
$\gamma_{\mathrm{ft},\lambda}=\gamma(B_{\lambda}^{\ast}A_{\lambda}^{\ast};\mathcal D_{\mathrm{ft}})$, and
$\gamma_{\lambda}=\gamma(W_{\mathrm{LoRA}}^{\lambda};\mathcal D_{\mathrm{pre}}\cup\mathcal D_{\mathrm{ft}})$.
Additionally, given an $\bar{K}$-class max-margin classifier $W_{\mathrm{ft}}^{\ast}\in\mathbb{R}^{\bar{K}\times d}$ on the fine-tuning data, let $\gamma_{\mathrm{ft}}^{\mathrm{pad}}=\gamma(([W_{\mathrm{ft}}^{\ast};\,\boldsymbol 0_{(K-\bar{K})\times d}]);\mathcal{D}_{\mathrm{ft}})$. When $\bar{K}<K$ this is in general a lower bound on $\max_{W\in\mathbb{R}^{K\times d}}\gamma(W;\mathcal{D}_{\mathrm{ft}})$, with equality iff $\bar{K}=K$. In the balanced case $\bar K=K$, $\gamma_{\mathrm{ft}}^{\mathrm{pad}}=\gamma_{\mathrm{ft}}^{\ast}$ as defined in the main theorem.

With these definitions in place, we restate Theorem~\ref{thm:lora_forget_less} in expanded form covering the general $\bar K\!\le\!K$ setting. The expanded version specifies the closed-form values of $\Theta_\lambda$ and $\gamma_{\mathrm{ft},\lambda}$ in three regimes when $\bar K\!<\!K$. In the balanced case $\bar K\!=\!K$, the intermediate and low-penalty regimes coincide in their ETF geometry and margin formulas, recovering the two-regime statement of Theorem~\ref{thm:lora_forget_less}. More precisely, $\lambda_{\mathrm{crit}}\!=\!0$, so the intermediate formula holds for every $0\!<\!\lambda\!<\!1/(K\sqrt n)$, and the limit $\lambda\!\to\!0^+$ retains the same ETF geometry.

\begin{thm}[Detailed version of Theorem~\ref{thm:lora_forget_less}]\label{appthm:detailed_forget}
  Adopt the setup of Theorem~\ref{thm:main} and let $\Theta_{\lambda}\!:=\!\|B^{\ast}_{\lambda}A^{\ast}_{\lambda}\|_F$. The normalized margins of $W_{\mathrm{LoRA}}^{\lambda}$ on the pre-training and fine-tuning data are
  \begin{align}\label{appeqn:margins_pre_ft}
      \gamma\bigl(W_{\mathrm{LoRA}}^{\lambda};\mathcal D_{\mathrm{pre}}\bigr)
       =\gamma_{\mathrm{pre}}\,\frac{\rho_{\mathrm{pre}}}{\sqrt{\Theta_{\lambda}^2+\rho_{\mathrm{pre}}^{2}}}\,,\quad
       \gamma\bigl(W_{\mathrm{LoRA}}^{\lambda};\mathcal D_{\mathrm{ft}}\bigr)
       = \gamma_{\mathrm{ft,\lambda}}\,\frac{\Theta_{\lambda}}{\sqrt{\Theta_{\lambda}^2+\rho_{\mathrm{pre}}^{2}}}\,,
  \end{align}
  which yield the closed-form $\mathrm{Adapt}(\lambda)$ and $\mathrm{Forget}(\lambda)$ in~\eqref{eqn:margin_combined} via $\mathrm{Adapt}(\lambda)=\gamma(W_{\mathrm{LoRA}}^{\lambda};\mathcal D_{\mathrm{ft}})$ (since $\gamma(W_{\mathrm{pre}};\mathcal D_{\mathrm{ft}})=0$ in the strict-orthogonal regime) and $\mathrm{Forget}(\lambda)=\gamma_{\mathrm{pre}}-\gamma(W_{\mathrm{LoRA}}^{\lambda};\mathcal D_{\mathrm{pre}})$. Moreover, $\Theta_\lambda$ and $\gamma_{\mathrm{ft},\lambda}$ take the following values:
  \begin{enumerate}[label=(\roman*)]
    \item \textbf{High--penalty regime}
          \(\lambda \ge \frac{1}{\bar{K}\sqrt{n}}\): $\Theta_{\lambda}=\gamma_{\mathrm{ft},\lambda}=0$, and hence $\mathrm{Adapt}(\lambda)=\mathrm{Forget}(\lambda)=0$.

    \item \textbf{Intermediate regime}
          \(\lambda_{\mathrm{crit}} < \lambda < \frac{1}{\bar{K}\sqrt{n}}\): $\Theta_{\lambda}=\frac{\rho_{\lambda}}{\sqrt{\bar{K}-1}}$, $\gamma_{\mathrm{ft,\lambda}}=\gamma_{\mathrm{ft}}^{\mathrm{pad}}$, and $\Theta_{\lambda}$ is a strictly decreasing function of $\lambda$. Both $\mathrm{Adapt}(\lambda)$ and $\mathrm{Forget}(\lambda)$ are therefore strictly decreasing in $\lambda$ on this regime.

    \item \textbf{Low--penalty regime} \(0\!<\!\lambda\!\le\!\lambda_{\mathrm{crit}}\):
    \begin{align*}
        \Theta_{\lambda}= \sqrt{n\bar{K}a_{\lambda}^{2}+n\bar{K}(\bar{K}-1)b_{\lambda}^{2}+n\bar{K}(K-\bar{K})c_{\lambda}^2}\,,\quad \gamma_{\mathrm{ft},\lambda}=\frac{a_{\lambda}+c_{\lambda}}{\Theta_{\lambda}}\,.
    \end{align*}
  \end{enumerate}
\end{thm}

\begin{rem}
\myparagraph{Full fine-tuning as an alternative method}
While our work focuses on LoRA, it is instructive to briefly contrast it with the more conventional strategy of full fine-tuning. Full fine-tuning updates all parameters of the pretrained model, which is considerably less efficient in both computation and storage, whereas LoRA achieves adaptation through a compact low-rank parameterization. In the linear classification setting, the two approaches correspond to the following regularized objectives:
\begin{align}
    &\min_{B, A} \;
    L\!\left(W_{\mathrm{pre}} + BA;\, \mathcal{D}_{\mathrm{ft}}\right)
    + \tfrac{\lambda}{2}\!\left(\|B\|_F^2 + \|A\|_F^2\right)\,,
    && \text{LoRA objective}\,, \label{eqn:lora_example}\\
    &\min_{W} \;
    L\!\left(W;\, \mathcal{D}_{\mathrm{ft}}\right)
    + \tfrac{\lambda}{2}\|W - W_{\mathrm{pre}}\|_F^2\,,
    && \text{Full fine-tuning objective}\,.
\end{align}
Beyond efficiency, the two formulations induce fundamentally different (implicit) biases. By exploiting the variational form of the nuclear norm~\citep{recht2010guaranteed}, one can show that for any solution $(B^\ast, A^\ast)$ of \eqref{eqn:lora_example}, the induced update $W_{\mathrm{LoRA}}^{\lambda}$ is equivalently the solution of
\begin{align}
    \min_{\substack{W, \mathrm{rank}(W-W_{\mathrm{pre}}) \leq r}} \;
    L\!\left(W;\, \mathcal{D}_{\mathrm{ft}}\right)
    + \lambda\|W - W_{\mathrm{pre}}\|_{\ast}\,,
    \qquad \text{Implicit bias of LoRA fine-tuning}\,,
\end{align}
which highlights LoRA's connection to nuclear-norm regularization under rank constraints. We refer the readers to \cite{jang2024lora} for the proof of the argument. Understanding the consequences of this distinction between nuclear norm regularization and Frobenius norm regularization remains an open direction for future work. Importantly, our goal is not to compare LoRA and full fine-tuning, but to rigorously analyze why LoRA exhibits reduced forgetting and to provide principled guidance for selecting regularization parameters in adapter-based fine-tuning.
\end{rem}

\begin{proof}
    Our key observation is that when one computes the margin of $W_{\mathrm{LoRA}}^{\lambda}$ on the pre-training and fine-tuning datasets, only $W_{\mathrm{pre}}$ or $B_{\lambda}^{\ast}A_{\lambda}^{\ast}$ is activated.
    This is because by Theorem~\ref{thm:main}, the product of the optimal LoRA adapters lies in the span of fine-tuning data, and by Proposition~\ref{prop:max_margin} together with Assumption~\ref{asmp:data}, $W_{\mathrm{pre}}$ lies in the span of pre-training data.
    Thus, under Assumption~\ref{asmp:data}, $B_{\lambda}^{\ast}A_{\lambda}^{\ast}x=0$ if $x\in\mathcal{D}_{\mathrm{pre}}$, and $W_{\mathrm{pre}}x = 0$ if $x\in\mathcal{D}_{\mathrm{ft}}$. In particular, $\gamma(W_{\mathrm{pre}};\mathcal D_{\mathrm{ft}})=0$, so $\mathrm{Adapt}(\lambda)=\gamma(W_{\mathrm{LoRA}}^{\lambda};\mathcal D_{\mathrm{ft}})$.
    Before proving that the margins take the structures shown in Theorem~\ref{appthm:detailed_forget}, we introduce the following notation. Let $W_{\mathrm{pre}}=[w_1,\dots,w_K]^\top$ and $B_{\lambda}^{\ast}A_{\lambda}^{\ast}=[\delta w_{\lambda, 1},\cdots, \delta w_{\lambda, K}]^\top$.

    Now, when we compute the margin of $W_{\mathrm{LoRA}}^{\lambda}$ on pre-training data, one has
    \begin{align*}
        \gamma(W_{\mathrm{LoRA}}^{\lambda}; \mathcal{D}_{\mathrm{pre}}) &= \frac{\min_{(x_{i,c},y_c)\in\mathcal{D}_{\mathrm{pre}}}(w_c+\delta_{\lambda, c})^\top x-\max_{i\not= y} (w_i+\delta_{\lambda, i})^\top x}{\|W_{\mathrm{LoRA}}^{\lambda}\|_F} \\
        &=\frac{\min_{(x_{i,c},y_c)\in\mathcal{D}_{\mathrm{pre}}}w_c^\top x-\max_{i\not= y} w_i^\top x}{\|W_{\mathrm{pre}}\|_F} \cdot \frac{\|W_{\mathrm{pre}}\|_F}{\|W_{\mathrm{LoRA}}^{\lambda}\|_F} \\
        &= \gamma_{\mathrm{pre}}\cdot \frac{\rho_{\mathrm{pre}}}{\sqrt{\rho_{\mathrm{pre}}^2 +\|B_{\lambda}^{\ast}A_{\lambda}^{\ast}\|_F^2}}\,.
    \end{align*}
    Then, for the fine-tuning data, one has
    \begin{align*}
        \gamma(W_{\mathrm{LoRA}}^{\lambda}; \mathcal{D}_{\mathrm{ft}}) &= \frac{\min_{(x_{i,c},y_c)\in\mathcal{D}_{\mathrm{ft}}}(w_c+\delta_{\lambda, c})^\top x-\max_{i\not= y} (w_i+\delta_{\lambda, i})^\top x}{\|W_{\mathrm{LoRA}}^{\lambda}\|_F} \\
        &=\frac{C(B_{\lambda}^{\ast}A_{\lambda}^{\ast};\mathcal D_{\mathrm{ft}})}{\sqrt{\rho_{\mathrm{pre}}^2+\|B_{\lambda}^{\ast}A_{\lambda}^{\ast}\|_F^2}}\,,\\
        &=\gamma_{\mathrm{ft},\lambda}\cdot \frac{\|B_{\lambda}^{\ast} A_{\lambda}^{\ast}\|_F}{\sqrt{\rho_{\mathrm{pre}}^2 +\| B_{\lambda}^{\ast}A_{\lambda}^{\ast}\|_F^2}}\,.
    \end{align*}
    The last equality also holds when $B_{\lambda}^{\ast}A_{\lambda}^{\ast}=0$ by the convention $\gamma(0;\mathcal D)=0$. These are exactly the formulas in~\eqref{appeqn:margins_pre_ft} with $\Theta_\lambda=\|B_{\lambda}^{\ast}A_{\lambda}^{\ast}\|_F$, and substituting them into the definitions of $\mathrm{Adapt}(\lambda)$ and $\mathrm{Forget}(\lambda)$ recovers~\eqref{eqn:margin_combined}. It remains to compute $\|B_{\lambda}^{\ast}A_{\lambda}^{\ast}\|_F$ in each regime.

    \myparagraph{High-penalty regime} In this regime, $B_{\lambda}^{\ast}A_{\lambda}^{\ast}=0$, so $\|B_{\lambda}^{\ast}A_{\lambda}^{\ast}\|_F=0$.

    \myparagraph{Intermediate regime} In the intermediate regime, one has
    \begin{align}
            B_{\lambda}^{\ast}A_{\lambda}^{\ast}
            \;=\;
            \frac{\rho_{\lambda}}{\sqrt{\bar{K}n(\bar{K}-1)}}\,
            \left(\begin{pmatrix}
        M_{\bar{K}}\\
        \mathbf 0_{(K-\bar{K})\times \bar{K}}
    \end{pmatrix}
    \otimes \boldsymbol 1_n^\top \right)X_{\mathrm{ft}}^\top\,,
    \end{align}
    and one can compute its norm as follows
    \begin{align*}
        \|B_{\lambda}^{\ast}A_{\lambda}^{\ast}\|_F^2&= \frac{\rho_{\lambda}^2}{\bar{K}n(\bar{K}-1)}\cdot \frac{\bar{K}}{\bar{K}-1} \bigl\|\left((\boldsymbol I_{\bar{K}}-\frac{1}{\bar{K}}\boldsymbol 1_{\bar{K}} 1_{\bar{K}}^\top )\right)\otimes \boldsymbol 1_n^\top X_{\mathrm{ft}}^\top\bigr\|_F^2 \\
        &= \frac{\rho_{\lambda}^2}{\bar{K}n(\bar{K}-1)}\cdot \frac{\bar{K}}{\bar{K}-1} \cdot \bar{K}\biggl(n\cdot (\frac{\bar{K}-1}{\bar{K}})^2+\frac{(\bar{K}-1)n}{\bar{K}^2}\biggr) \\
        &= \frac{\rho_{\lambda}^2}{\bar{K}-1}\,.
    \end{align*}
    Thus, $\Theta_{\lambda}=\|B_{\lambda}^{\ast}A_{\lambda}^{\ast}\|_F=\frac{\rho_{\lambda}}{\sqrt{\bar{K}-1}}$.

    \myparagraph{Low-penalty regime} In this regime, $B_{\lambda}^{\ast}A_{\lambda}^{\ast}$ takes the following form
    \begin{align}
    B_{\lambda}^{\ast}A_{\lambda}^{\ast}
        =
        \left(\begin{pmatrix}
            \bigl(a_{\lambda}+b_{\lambda}\bigr)\,\mathbf I_{\bar{K}}
            \;-\;
            b_{\lambda}\,\mathbf 1_{\bar{K}}\mathbf 1_{\bar{K}}^{\!\top}\\
            -c_{\lambda}\,\mathbf 1_{K-\bar{K}}\mathbf 1_{\bar{K}}^{\!\top}
        \end{pmatrix}
        \otimes 1_n^\top\right) X_{\mathrm{ft}}^\top\,,
    \end{align}
    and its Frobenius norm is
    \begin{align*}
        \|B_{\lambda}^{\ast}A_{\lambda}^{\ast}\|_F &= \sqrt{n} \biggl\|\begin{pmatrix}
            \bigl(a_{\lambda}+b_{\lambda}\bigr)\,\mathbf I_{\bar{K}}
            \;-\;
            b_{\lambda}\,\mathbf 1_{\bar{K}}\mathbf 1_{\bar{K}}^{\!\top}\\
            -c_{\lambda}\,\mathbf 1_{K-\bar{K}}\mathbf 1_{\bar{K}}^{\!\top}
        \end{pmatrix}\biggr\|_F \\
        &=\sqrt{n} \cdot\sqrt{\bar{K}a_{\lambda}^2+\bar{K}(\bar{K}-1)b_{\lambda}^2+(K-\bar{K})\bar{K} c_{\lambda}^2}\,.
    \end{align*}
    For any sample $\boldsymbol{\bar x}_{c,j}\in\mathcal D_{\mathrm{ft}}$ in class $c\le\bar K$, the score on the correct class is $a_{\lambda}$, the score on any other class $k\le\bar K, k\neq c$ is $-b_{\lambda}$, and the score on any class $k>\bar K$ is $-c_{\lambda}$. Hence the per-sample margin is $a_{\lambda}-\max(-b_{\lambda},-c_{\lambda})=a_{\lambda}+\min(b_{\lambda},c_{\lambda})$. By the defining equation of $x_0$ in Lemma~\ref{applem:sol_step_one}, $g_b(x_0)/g_c(x_0)=1/(\sqrt{K/(K-\bar K)}+1)<1$, so $\log g_b(x_0)<\log g_c(x_0)$. Combined with $b_{\lambda}=-\log g_b(x_0)$ and $c_{\lambda}=-\log g_c(x_0)$, this gives $b_{\lambda}>c_{\lambda}$, so $\min(b_{\lambda},c_{\lambda})=c_{\lambda}$ and the unnormalized fine-tuning margin equals $a_{\lambda}+c_{\lambda}$. Dividing by $\Theta_{\lambda}=\|B_{\lambda}^{\ast}A_{\lambda}^{\ast}\|_F$ yields $\gamma_{\mathrm{ft},\lambda}=(a_{\lambda}+c_{\lambda})/\Theta_{\lambda}$.
\end{proof}

\subsection{Proof of Corollary~\ref{cor:optimal_lambda_tradeoff}}

\begin{proof}
The first identity in Corollary~\ref{cor:optimal_lambda_tradeoff},
\begin{align*}
    \gamma(W_{\mathrm{LoRA}}^\lambda;\mathcal D_{\mathrm{pre}}\cup\mathcal D_{\mathrm{ft}})
    =\min\bigl\{\gamma_{\mathrm{pre}}-\mathrm{Forget}(\lambda),\,\mathrm{Adapt}(\lambda)\bigr\}\,,
\end{align*}
follows from the definition of the union margin together with $\gamma(W_{\mathrm{LoRA}}^\lambda;\mathcal D_{\mathrm{pre}})=\gamma_{\mathrm{pre}}-\mathrm{Forget}(\lambda)$ and $\gamma(W_{\mathrm{LoRA}}^\lambda;\mathcal D_{\mathrm{ft}})=\mathrm{Adapt}(\lambda)$ from Theorem~\ref{thm:lora_forget_less}. It remains to maximize this minimum over $\lambda$.

We work in the balanced setting $\bar K\!=\!K$ of Corollary~\ref{cor:optimal_lambda_tradeoff}. In the high-penalty regime ($\lambda\!\ge\!1/(K\sqrt n)$), $\Theta_\lambda\!=\!0$ and so $\mathrm{Adapt}(\lambda)\!=\!0$, which gives a union margin of $0$. Every $0\!<\!\lambda\!<\!1/(K\sqrt n)$ gives a positive union margin, so the optimum lies in this interval.

On this interval, $\gamma_{\mathrm{ft},\lambda}\!=\!\gamma_{\mathrm{ft}}^{\ast}$ is constant. By the balanced-case solution in Appendix~\ref{app:proof_main_theorem}, $\Theta_\lambda$ decreases continuously and strictly from $+\infty$ to $0$ as $\lambda$ increases from $0$ to $1/(K\sqrt n)$. From~\eqref{appeqn:margins_pre_ft}, the pre-training margin is strictly decreasing in $\Theta_\lambda$ and the fine-tuning margin is strictly increasing in $\Theta_\lambda$. Their minimum is therefore uniquely maximized where they are equal:
\begin{align*}
    \gamma_{\mathrm{pre}}\,\frac{\rho_{\mathrm{pre}}}{\sqrt{\Theta_\lambda^{2}+\rho_{\mathrm{pre}}^{2}}}=\gamma_{\mathrm{ft}}^{\ast}\,\frac{\Theta_\lambda}{\sqrt{\Theta_\lambda^{2}+\rho_{\mathrm{pre}}^{2}}}
    \;\Longleftrightarrow\;
    \Theta_\lambda=\Theta^{\ast}:=\frac{\rho_{\mathrm{pre}}\,\gamma_{\mathrm{pre}}}{\gamma_{\mathrm{ft}}^{\ast}}\,.
\end{align*}
Since $0\!<\!\Theta^{\ast}\!<\!\infty$, there is a unique $\lambda^{\ast}\!\in\!(0,1/(K\sqrt n))$ with $\Theta_{\lambda^{\ast}}\!=\!\Theta^{\ast}$. No additional balancing condition is needed. Substituting $\Theta^{\ast}$ into~\eqref{appeqn:margins_pre_ft} gives
\begin{align*}
    \gamma(W_{\mathrm{LoRA}}^{\lambda^{\ast}};\mathcal D_{\mathrm{pre}}\cup\mathcal D_{\mathrm{ft}})
    =\frac{\gamma_{\mathrm{pre}}\,\gamma_{\mathrm{ft}}^{\ast}}{\sqrt{\gamma_{\mathrm{pre}}^{2}+(\gamma_{\mathrm{ft}}^{\ast})^{2}}}
    =\left(\frac{1}{(\gamma_{\mathrm{ft}}^{\ast})^{2}}+\frac{1}{\gamma_{\mathrm{pre}}^{2}}\right)^{-1/2}\,.
\end{align*}
\end{proof}

\subsection{Proof of General Version}\label{appsec:general_version_forget}
In this subsection we present the general version of Theorem~\ref{thm:lora_forget_less}: under the near-orthogonal regime in Assumption~\ref{asmp:data}, the normalized margins of $W_{\mathrm{LoRA}}^{\lambda}$ on pre-training, fine-tuning, and combined data remain close to their strict-orthogonal closed-form values from Theorem~\ref{appthm:detailed_forget}, with deviations controlled by the perturbation error $E_\lambda+\epsilon_2$ defined below.

\myparagraph{Notation} We adopt the notation of Theorem~\ref{appthm:general}: $G,R,Q,S,\delta,\delta_0,s_0,\mu_0,C_0$, and $Z_\lambda^{\ast}(\bar S,\bar R)$, with the key bound \eqref{appeqn:gen_main_bound}. Additionally, write
\begin{align}\label{appeqn:gen_forget_theta_def}
    \Theta_\lambda^{(0)}\!:=\!\|Z_\lambda^{\ast}(0,I)\|_F\,,\qquad \gamma_{\mathrm{ft},\lambda}^{(0)}\!:=\!\bigl.\gamma_{\mathrm{ft},\lambda}\bigr|_{\epsilon_1=\epsilon_2=0}\,,
\end{align}
for the strict-orthogonal adapter Frobenius norm and the strict-orthogonal fine-tuning margin from Theorem~\ref{appthm:detailed_forget}. (When $X_{\mathrm{ft}}^\top X_{\mathrm{ft}}=I$, $\Theta_\lambda^{(0)}=\|B_\lambda^{\ast}A_\lambda^{\ast}\|_F$ and $\gamma_{\mathrm{ft},\lambda}^{(0)}=\gamma_{\mathrm{ft},\lambda}$.) Set
\begin{align}\label{appeqn:gen_forget_E_def}
    E_\lambda\!:=\!\frac{C_0}{\mu_0}\bigl(\|S\|_F+\delta\,(1+\|Z_\lambda^{\ast}(0,I)\|_F)\bigr)\quad\text{(the bound in \eqref{appeqn:gen_main_bound})}\,.
\end{align}
The construction $C_0\!=\!C_1+\mu_0$ in the proof of Theorem~\ref{appthm:general} gives $C_0/\mu_0\!>\!1$, so $\|S\|_F\!\le\!E_\lambda$ and $\delta\!\le\!E_\lambda$. Therefore, $\|S\|_F=O(\epsilon_2)$, and $E_\lambda\to 0$ as $\epsilon_1,\epsilon_2\to 0$.

\begin{thm}[General version of Theorem~\ref{thm:lora_forget_less}]\label{appthm:general_forget}
Adopt the setup of Theorem~\ref{appthm:general} (so $\delta\!\le\!\delta_0$ and $\|S\|_F\!\le\!s_0$). Assume additionally that pre-training samples are unit-norm. There exists a constant $C\!>\!0$, depending on $\lambda,K,\bar K,n,\rho_{\mathrm{pre}},\mu_0,C_0,\|Z_\lambda^{\ast}(0,I)\|_*$ but not on $\epsilon_1,\epsilon_2$, such that
\begin{align}\label{appeqn:gen_forget_pre}
    \Bigl|\gamma\bigl(W_{\mathrm{LoRA}}^{\lambda};\mathcal D_{\mathrm{pre}}\bigr)-\gamma_{\mathrm{pre}}\,\frac{\rho_{\mathrm{pre}}}{\sqrt{(\Theta_\lambda^{(0)})^2+\rho_{\mathrm{pre}}^{2}}}\Bigr|&\le C\,\bigl(E_\lambda+\epsilon_2\bigr)\,,\\
    \Bigl|\gamma\bigl(W_{\mathrm{LoRA}}^{\lambda};\mathcal D_{\mathrm{ft}}\bigr)-\gamma_{\mathrm{ft},\lambda}^{(0)}\,\frac{\Theta_\lambda^{(0)}}{\sqrt{(\Theta_\lambda^{(0)})^2+\rho_{\mathrm{pre}}^{2}}}\Bigr|&\le C\,\bigl(E_\lambda+\epsilon_2\bigr)\,. \label{appeqn:gen_forget_ft}
\end{align}
The corresponding bound on $\gamma(W_{\mathrm{LoRA}}^{\lambda};\mathcal D_{\mathrm{pre}}\cup\mathcal D_{\mathrm{ft}})=\min\{\cdot,\cdot\}$ inherits the same right-hand side. In particular, as $E_\lambda+\epsilon_2\!\to\!0$, the three margins converge to their strict-orthogonal closed-form values from Theorem~\ref{appthm:detailed_forget}.
\end{thm}

\begin{proof}
Throughout the proof, $C$ denotes a positive constant whose value may change line to line, but always depends only on the quantities listed in the theorem statement.

We first establish leakage bounds quantifying how much pre-training and fine-tuning components mix in the near-orthogonal regime. Since $S\!=\!W_{\mathrm{pre}}X_{\mathrm{ft}}$, for any fine-tuning sample $\bar x_j$ and any row $w_k$ of $W_{\mathrm{pre}}$,
\begin{align}\label{appeqn:leak_ft}
    |w_k^\top\bar x_j|=|S_{kj}|\le\|S\|_F\le E_\lambda\,.
\end{align}
For the symmetric direction, write $\delta_{\lambda,k}:=(B_\lambda^{\ast}A_\lambda^{\ast})_k^\top$ for the $k$-th row of $B_\lambda^{\ast}A_\lambda^{\ast}$. By the reformulation in the proof of Theorem~\ref{appthm:general}, $B_\lambda^{\ast}A_\lambda^{\ast}=Z_\lambda^{\ast}(S,R)\,Q^\top$, so for any unit-norm $x\in\mathcal D_{\mathrm{pre}}$,
\begin{align}\label{appeqn:leak_pre}
    |\delta_{\lambda,k}^\top x|\le\|(Z_\lambda^{\ast}(S,R))_k\|_2\,\|Q^\top x\|_2\,.
\end{align}
Since $\|Q^\top x\|_2=\|R^{-1}X_{\mathrm{ft}}^\top x\|_2\le\|R^{-1}\|_{\mathrm{op}}\,\|X_{\mathrm{ft}}^\top x\|_2$, $\|R^{-1}\|_{\mathrm{op}}\le 1/\sqrt{1-\delta}$, and $\|X_{\mathrm{ft}}^\top x\|_2^2=\sum_l|\bar x_l^\top x|^2\le\bar Kn\,\epsilon_2^2$ by Assumption~\ref{asmp:data}(c), we have $|\delta_{\lambda,k}^\top x|\le\frac{\sqrt{\bar Kn}\,\epsilon_2}{\sqrt{1-\delta}}\,\|(Z_\lambda^{\ast}(S,R))_k\|_2$. Using $\|Z_\lambda^{\ast}(S,R)\|_F\le\|Z_\lambda^{\ast}(0,I)\|_F+E_\lambda$ from \eqref{appeqn:gen_zbound},
\begin{align}\label{appeqn:leak_pre_total}
    |\delta_{\lambda,k}^\top x|\le C\,(\|Z_\lambda^{\ast}(0,I)\|_F+E_\lambda)\,\epsilon_2\le C\,\epsilon_2\,.
\end{align}

Then, we bound the perturbation of the Frobenius norm $\|W_{\mathrm{LoRA}}^{\lambda}\|_F$ from $\sqrt{\rho_{\mathrm{pre}}^2+(\Theta_\lambda^{(0)})^2}$. Expanding,
\begin{align}\label{appeqn:gen_forget_Wnorm_expand}
    \|W_{\mathrm{LoRA}}^{\lambda}\|_F^2=\|W_{\mathrm{pre}}\|_F^2+2\,\bigl\langle W_{\mathrm{pre}},B_\lambda^{\ast}A_\lambda^{\ast}\bigr\rangle+\|B_\lambda^{\ast}A_\lambda^{\ast}\|_F^2\,.
\end{align}
Each row of $B_\lambda^{\ast}A_\lambda^{\ast}$ lies in $\mathrm{span}(\mathcal D_{\mathrm{ft}})$ (the column space of $X_{\mathrm{ft}}$), so
\begin{align}\label{appeqn:gen_forget_cross_proj}
    \bigl|\bigl\langle W_{\mathrm{pre}},B_\lambda^{\ast}A_\lambda^{\ast}\bigr\rangle\bigr|=\Bigl|\sum_k w_k^\top\delta_{\lambda,k}\Bigr|\le\sum_k\|P_{X_{\mathrm{ft}}}w_k\|_2\,\|(B_\lambda^{\ast}A_\lambda^{\ast})_k\|_2\,,
\end{align}
where $P_{X_{\mathrm{ft}}}$ is the orthogonal projection onto $\mathrm{span}(\mathcal D_{\mathrm{ft}})$. The identity $W_{\mathrm{pre}}Q\!=\!SR^{-1}$ gives $\|W_{\mathrm{pre}}Q\|_F\!\le\!\|S\|_F/\sqrt{1-\delta}$. Thus, Cauchy--Schwarz and $B_\lambda^\ast A_\lambda^\ast\!=\!Z_\lambda^\ast(S,R)Q^\top$ yield
\begin{align}\label{appeqn:gen_forget_cross_bound}
    \bigl|\bigl\langle W_{\mathrm{pre}},B_\lambda^{\ast}A_\lambda^{\ast}\bigr\rangle\bigr|\le\frac{\|S\|_F\,\|B_\lambda^{\ast}A_\lambda^{\ast}\|_F}{\sqrt{1-\delta}}\le C\,(\Theta_\lambda^{(0)}+E_\lambda)\,\|S\|_F\,.
\end{align}
Write $\Delta_\lambda\!:=\!B_\lambda^\ast A_\lambda^\ast$. Equation~\eqref{appeqn:gen_zbound} and $Q^\top Q\!=\!I$ give $\bigl|\|\Delta_\lambda\|_F-\Theta_\lambda^{(0)}\bigr|\!\le\!E_\lambda$. Comparing first with $\sqrt{\rho_{\mathrm{pre}}^2+\|\Delta_\lambda\|_F^2}$ and then using this norm bound yields
\begin{align}\label{appeqn:gen_forget_norm}
    \Bigl|\,\|W_{\mathrm{LoRA}}^{\lambda}\|_F-\sqrt{\rho_{\mathrm{pre}}^2+(\Theta_\lambda^{(0)})^2}\,\Bigr|
    &\le\frac{2\bigl|\langle W_{\mathrm{pre}},\Delta_\lambda\rangle\bigr|}
    {\|W_{\mathrm{LoRA}}^\lambda\|_F+\sqrt{\rho_{\mathrm{pre}}^2+\|\Delta_\lambda\|_F^2}}+E_\lambda\nonumber\\
    &\le\frac{2\|S\|_F}{\sqrt{1-\delta}}+E_\lambda\le C\,(E_\lambda+\epsilon_2)\,.
\end{align}

Next, we bound the perturbation of the unnormalized margins $C(W_{\mathrm{LoRA}}^{\lambda};\cdot)$ on each dataset. For pre-training data $(x,y_c)\in\mathcal D_{\mathrm{pre}}$ and $k\ne c$,
\begin{align}\label{appeqn:gen_forget_pre_decomp}
    [(w_c+\delta_{\lambda,c})-(w_k+\delta_{\lambda,k})]^\top x=(w_c-w_k)^\top x+(\delta_{\lambda,c}-\delta_{\lambda,k})^\top x\,.
\end{align}
Taking the minimum over $(x,c,k)$ and combining $\min(w_c-w_k)^\top x = C(W_{\mathrm{pre}};\mathcal D_{\mathrm{pre}})=\rho_{\mathrm{pre}}\gamma_{\mathrm{pre}}$ (definition of $\gamma_{\mathrm{pre}}$) with $|\delta_{\lambda,c}^\top x-\delta_{\lambda,k}^\top x|\le 2\,C\epsilon_2$ from \eqref{appeqn:leak_pre_total},
\begin{align}\label{appeqn:gen_forget_C_pre}
    \bigl|C(W_{\mathrm{LoRA}}^{\lambda};\mathcal D_{\mathrm{pre}})-\rho_{\mathrm{pre}}\gamma_{\mathrm{pre}}\bigr|\le C\,\epsilon_2\,.
\end{align}
For fine-tuning data $(\bar x,y_c)\in\mathcal D_{\mathrm{ft}}$ and $k\ne c$,
\begin{align}\label{appeqn:gen_forget_ft_decomp}
    [(w_c+\delta_{\lambda,c})-(w_k+\delta_{\lambda,k})]^\top\bar x=(\delta_{\lambda,c}-\delta_{\lambda,k})^\top\bar x+(w_c-w_k)^\top\bar x\,.
\end{align}
The unnormalized fine-tuning margin is $2$-Lipschitz in the F-norm of the classifier matrix (since for any $W,W'\in\R^{K\times d}$ and any unit-norm $\bar x$, $|((W-W')_c-(W-W')_k)^\top\bar x|\le 2\|W-W'\|_F$), so by Theorem~\ref{appthm:general},
\begin{align}\label{appeqn:gen_forget_step3a}
    \bigl|C(B_\lambda^{\ast}A_\lambda^{\ast};\mathcal D_{\mathrm{ft}})-C(Z_\lambda^{\ast}(0,I)X_{\mathrm{ft}}^\top;\mathcal D_{\mathrm{ft}})\bigr|\le 2\,E_\lambda\,.
\end{align}
We further compare $C(Z_\lambda^{\ast}(0,I)X_{\mathrm{ft}}^\top;\mathcal D_{\mathrm{ft}})$ — the unnormalized margin of the strict-orthogonal classifier evaluated on the actual (near-orthogonal) data — with its strict-orthogonal value $\Theta_\lambda^{(0)}\gamma_{\mathrm{ft},\lambda}^{(0)}$. For any sample $\bar x_{c',j'}\in\mathcal D_{\mathrm{ft}}$,
\begin{align}\label{appeqn:gen_forget_residual_def}
    X_{\mathrm{ft}}^\top \bar x_{c',j'}=e_{(c'-1)n+j'}+r_{c',j'}\,,\qquad r_{c',j'}\!:=\!X_{\mathrm{ft}}^\top \bar x_{c',j'}-e_{(c'-1)n+j'}\,,
\end{align}
where the $((c''-1)n+j'')$-th entry of $r_{c',j'}$ is $\bar x_{c'',j''}^\top\bar x_{c',j'}$ for $(c'',j'')\ne(c',j')$ and $0$ otherwise. By Assumption~\ref{asmp:data}(b),
\begin{align}\label{appeqn:gen_forget_residual_norm}
    \|r_{c',j'}\|_2^2\le\sum_{(c'',j'')\ne(c',j')}\epsilon_1^2=(\bar Kn-1)\,\epsilon_1^2\,,
\end{align}
hence $\|r_{c',j'}\|_2\le\sqrt{\bar Kn-1}\,\epsilon_1=\delta/\sqrt{\bar Kn-1}$. For any class $k\in[K]$,
\begin{align}\label{appeqn:gen_forget_score_perturb}
    \bigl|(Z_\lambda^{\ast}(0,I)X_{\mathrm{ft}}^\top \bar x_{c',j'})_k-(Z_\lambda^{\ast}(0,I))_{k,(c'-1)n+j'}\bigr|=\bigl|(Z_\lambda^{\ast}(0,I))_k r_{c',j'}\bigr|\le\|(Z_\lambda^{\ast}(0,I))_k\|_2\,\|r_{c',j'}\|_2\,,
\end{align}
where $(Z_\lambda^{\ast}(0,I))_k$ denotes the $k$-th row. Since $\max_k\|(Z_\lambda^{\ast}(0,I))_k\|_2\le\|Z_\lambda^{\ast}(0,I)\|_F=\Theta_\lambda^{(0)}$, the difference of two-class scores satisfies
\begin{align}\label{appeqn:gen_forget_pair_perturb}
    &\bigl|[(Z_\lambda^{\ast}(0,I)X_{\mathrm{ft}}^\top \bar x_{c',j'})_c-(Z_\lambda^{\ast}(0,I)X_{\mathrm{ft}}^\top \bar x_{c',j'})_k]-[(Z_\lambda^{\ast}(0,I))_{c,(c'-1)n+j'}-(Z_\lambda^{\ast}(0,I))_{k,(c'-1)n+j'}]\bigr|\nonumber\\
    \le&\frac{2\,\Theta_\lambda^{(0)}\,\delta}{\sqrt{\bar Kn-1}}
\end{align}
for every $\bar x\in\mathcal D_{\mathrm{ft}}$ and every pair $c\ne k$. Setting $c=c'$ and taking the minimum over $c'\in[\bar K]$, $j'\in[n]$, and $k\in[K]\setminus\{c'\}$, the $1$-Lipschitz property of $\min$ gives
\begin{align}\label{appeqn:gen_forget_min_perturb}
    \bigl|C(Z_\lambda^{\ast}(0,I)X_{\mathrm{ft}}^\top;\mathcal D_{\mathrm{ft}})-\min_{c',j',k\ne c'}\bigl[(Z_\lambda^{\ast}(0,I))_{c',(c'-1)n+j'}-(Z_\lambda^{\ast}(0,I))_{k,(c'-1)n+j'}\bigr]\bigr|\le\frac{2\,\Theta_\lambda^{(0)}\,\delta}{\sqrt{\bar Kn-1}}\,.
\end{align}
The displayed minimum is precisely the strict-orthogonal unnormalized margin $\Theta_\lambda^{(0)}\,\gamma_{\mathrm{ft},\lambda}^{(0)}$ (it is the minimum-margin definition applied to $Z_\lambda^{\ast}(0,I)$ on a one-hot $\bar K n$-dimensional input set, which is what Theorem~\ref{appthm:detailed_forget} characterizes). Hence
\begin{align}\label{appeqn:gen_forget_step3b}
    \bigl|C(Z_\lambda^{\ast}(0,I)X_{\mathrm{ft}}^\top;\mathcal D_{\mathrm{ft}})-\Theta_\lambda^{(0)}\gamma_{\mathrm{ft},\lambda}^{(0)}\bigr|\le\frac{2\,\Theta_\lambda^{(0)}}{\sqrt{\bar Kn-1}}\,\delta\le C\,\delta\,.
\end{align}
The leakage from the pre-training rows on $\mathcal D_{\mathrm{ft}}$ contributes at most $2\|S\|_F$ to the per-pair score difference via \eqref{appeqn:leak_ft}, so taking the min,
\begin{align}\label{appeqn:gen_forget_pre_leak_ft}
    \bigl|C(W_{\mathrm{LoRA}}^{\lambda};\mathcal D_{\mathrm{ft}})-C(B_\lambda^{\ast}A_\lambda^{\ast};\mathcal D_{\mathrm{ft}})\bigr|\le 2\|S\|_F\le 2E_\lambda\,.
\end{align}
Combining with \eqref{appeqn:gen_forget_step3a} and \eqref{appeqn:gen_forget_step3b} via the triangle inequality (and noting $\delta\le(\mu_0/C_0)\,E_\lambda$ by definition of $E_\lambda$, so the $\delta$ from \eqref{appeqn:gen_forget_step3b} is absorbed into $C\,E_\lambda$),
\begin{align}\label{appeqn:gen_forget_C_ft}
    \bigl|C(W_{\mathrm{LoRA}}^{\lambda};\mathcal D_{\mathrm{ft}})-\Theta_\lambda^{(0)}\gamma_{\mathrm{ft},\lambda}^{(0)}\bigr|\le C\,(E_\lambda+\epsilon_2)\,.
\end{align}

Finally, we combine the bounds. For a dataset of unit-norm inputs, $|\gamma(W;\mathcal D)|\!\le\!\sqrt2$ whenever $W\!\ne\!0$. Let $a\!:=\!\|W_{\mathrm{LoRA}}^\lambda\|_F$, $b\!:=\!\sqrt{\rho_{\mathrm{pre}}^2+(\Theta_\lambda^{(0)})^2}$, $u\!:=\!C(W_{\mathrm{LoRA}}^\lambda;\mathcal D)$, and let $u_0$ be the corresponding strict-orthogonal unnormalized margin. Then
\begin{align}\label{appeqn:gen_forget_ratio}
    \left|\frac{u}{a}-\frac{u_0}{b}\right|
    =\left|\frac{u-u_0}{b}+\frac{u}{a}\frac{b-a}{b}\right|
    \le\frac{|u-u_0|+\sqrt2\,|a-b|}{b}\,.
\end{align}
Since $b\!\ge\!\rho_{\mathrm{pre}}\!>\!0$, \eqref{appeqn:gen_forget_norm} combined with \eqref{appeqn:gen_forget_C_pre} gives \eqref{appeqn:gen_forget_pre} and combined with \eqref{appeqn:gen_forget_C_ft} gives \eqref{appeqn:gen_forget_ft}. Taking the minimum of the pre-training and fine-tuning margins gives the combined-data bound by the $1$-Lipschitz property of $\min$.
\end{proof}

We now translate Theorem~\ref{appthm:general_forget} into perturbation bounds on $\mathrm{Adapt}(\lambda)$ and $\mathrm{Forget}(\lambda)$, and on the optimal-trade-off value of Corollary~\ref{cor:optimal_lambda_tradeoff}.

\begin{cor}[Near-orthogonal bounds on $\mathrm{Adapt}$ and $\mathrm{Forget}$]\label{appcor:gen_adapt_forget}
Under the hypotheses of Theorem~\ref{appthm:general_forget}, with $C$ as there,
\begin{align}\label{appeqn:gen_adapt}
    \Bigl|\mathrm{Adapt}(\lambda)-\gamma_{\mathrm{ft},\lambda}^{(0)}\,\frac{\Theta_\lambda^{(0)}}{\sqrt{(\Theta_\lambda^{(0)})^2+\rho_{\mathrm{pre}}^{2}}}\Bigr|&\le C\,(E_\lambda+\epsilon_2)\,,\\
    \Bigl|\mathrm{Forget}(\lambda)-\gamma_{\mathrm{pre}}\Bigl(1-\frac{\rho_{\mathrm{pre}}}{\sqrt{(\Theta_\lambda^{(0)})^2+\rho_{\mathrm{pre}}^{2}}}\Bigr)\Bigr|&\le C\,(E_\lambda+\epsilon_2)\,.\label{appeqn:gen_forget}
\end{align}
In particular, as $E_\lambda+\epsilon_2\!\to\!0$, both $\mathrm{Adapt}(\lambda)$ and $\mathrm{Forget}(\lambda)$ converge to their strict-orthogonal closed-form values from Theorem~\ref{appthm:detailed_forget}.
\end{cor}

\begin{proof}
For the Forget bound, note that $\gamma_{\mathrm{pre}}=\gamma(W_{\mathrm{pre}};\mathcal D_{\mathrm{pre}})$ does not depend on $\epsilon_1,\epsilon_2$ and that $\mathrm{Forget}(\lambda)=\gamma_{\mathrm{pre}}-\gamma(W_{\mathrm{LoRA}}^\lambda;\mathcal D_{\mathrm{pre}})$, so~\eqref{appeqn:gen_forget} is just~\eqref{appeqn:gen_forget_pre} with the $\gamma_{\mathrm{pre}}$ term absorbed. For the Adapt bound, use $\mathrm{Adapt}(\lambda)=\gamma(W_{\mathrm{LoRA}}^\lambda;\mathcal D_{\mathrm{ft}})-\gamma(W_{\mathrm{pre}};\mathcal D_{\mathrm{ft}})$. The leakage bound~\eqref{appeqn:leak_ft} controls the second term: for any $\bar x\in\mathcal D_{\mathrm{ft}}$ and any pair $c\ne k$, $|(w_c-w_k)^\top\bar x|\!\le\!2\|S\|_F$, so taking the min over $(\bar x,c,k)$ and dividing by $\|W_{\mathrm{pre}}\|_F=\rho_{\mathrm{pre}}$ gives $|\gamma(W_{\mathrm{pre}};\mathcal D_{\mathrm{ft}})|\!\le\!2\|S\|_F/\rho_{\mathrm{pre}}\!\le\!2E_\lambda/\rho_{\mathrm{pre}}$. Combining with~\eqref{appeqn:gen_forget_ft} via the triangle inequality yields~\eqref{appeqn:gen_adapt}.
\end{proof}

\begin{cor}[Near-orthogonal version of Corollary~\ref{cor:optimal_lambda_tradeoff}]\label{appcor:gen_tradeoff}
Let $\gamma_{\mathrm{ft}}^{\mathrm{pad},(0)}$ denote the strict-orthogonal reference value of $\gamma_{\mathrm{ft}}^{\mathrm{pad}}$.
Adopt the hypotheses of Theorem~\ref{appthm:general_forget}, and set $\Theta^\ast\!:=\!\rho_{\mathrm{pre}}\gamma_{\mathrm{pre}}/\gamma_{\mathrm{ft}}^{\mathrm{pad},(0)}$. If $\bar K\!<\!K$, assume $\Theta^\ast\!<\!\Theta_{\lambda_{\mathrm{crit}}}^{(0)}$; when $\bar K\!=\!K$, no additional balancing condition is needed. Let $\lambda^\ast$ be the unique solution of $\Theta_{\lambda^\ast}^{(0)}\!=\!\Theta^\ast$ in $(\lambda_{\mathrm{crit}},1/(\bar K\sqrt n))$, and choose a compact subinterval $\Lambda$ of this open interval containing $\lambda^\ast$. Writing $\delta_0(\lambda),s_0(\lambda)$ for the local thresholds in Theorem~\ref{appthm:general}, require
\begin{align*}
    \delta\le\inf_{\lambda\in\Lambda}\delta_0(\lambda)\,,\qquad
    \|S\|_F\le\inf_{\lambda\in\Lambda}s_0(\lambda)\,.
\end{align*}
Suppose $E_\lambda\!\le\!\bar E$ for all $\lambda\!\in\!\Lambda$, and the constant $C$ in Theorem~\ref{appthm:general_forget} can be taken uniform on $\Lambda$ with value $\bar C$. Then
\begin{align}\label{appeqn:gen_tradeoff}
    \Bigl|\max_{\lambda\in\Lambda}\gamma(W_{\mathrm{LoRA}}^\lambda;\mathcal D_{\mathrm{pre}}\cup\mathcal D_{\mathrm{ft}})-\Bigl(\frac{1}{(\gamma_{\mathrm{ft}}^{\mathrm{pad},(0)})^{2}}+\frac{1}{\gamma_{\mathrm{pre}}^{2}}\Bigr)^{-1/2}\Bigr|\le \bar C\,(\bar E+\epsilon_2)\,.
\end{align}
In particular, as $\bar E+\epsilon_2\!\to\!0$ (with $\Lambda$ fixed), the maximum converges to the corresponding strict-orthogonal intermediate-regime balancing value, which reduces to the value in Corollary~\ref{cor:optimal_lambda_tradeoff} when $\bar K=K$.
\end{cor}

\begin{proof}
By Theorem~\ref{appthm:general_forget} applied uniformly on $\Lambda$, together with the $1$-Lipschitz property of $\min$,
\begin{align*}
    \bigl|\gamma(W_{\mathrm{LoRA}}^\lambda;\mathcal D_{\mathrm{pre}}\cup\mathcal D_{\mathrm{ft}})-\gamma^{(0)}(\lambda)\bigr|\le \bar C\,(\bar E+\epsilon_2)\quad\text{for every }\lambda\in\Lambda\,,
\end{align*}
where $\gamma^{(0)}(\lambda):=\min\bigl\{\gamma_{\mathrm{pre}}\rho_{\mathrm{pre}}/\sqrt{(\Theta_\lambda^{(0)})^2+\rho_{\mathrm{pre}}^{2}},\,\gamma_{\mathrm{ft},\lambda}^{(0)}\Theta_\lambda^{(0)}/\sqrt{(\Theta_\lambda^{(0)})^2+\rho_{\mathrm{pre}}^{2}}\bigr\}$ is the strict-orthogonal union margin. Taking $\sup_{\lambda\in\Lambda}$ on both sides and using $|\sup_\Lambda f-\sup_\Lambda g|\le\sup_\Lambda|f-g|$,
\begin{align*}
    \Bigl|\max_{\lambda\in\Lambda}\gamma(W_{\mathrm{LoRA}}^\lambda;\mathcal D_{\mathrm{pre}}\cup\mathcal D_{\mathrm{ft}})-\max_{\lambda\in\Lambda}\gamma^{(0)}(\lambda)\Bigr|\le \bar C\,(\bar E+\epsilon_2)\,.
\end{align*}
On $\Lambda$, $\gamma_{\mathrm{ft},\lambda}^{(0)}\!=\!\gamma_{\mathrm{ft}}^{\mathrm{pad},(0)}$ and $\Theta_\lambda^{(0)}$ is strictly decreasing in $\lambda$. The pre-training term in $\gamma^{(0)}$ is strictly decreasing in $\Theta_\lambda^{(0)}$, while the fine-tuning term is strictly increasing. Their minimum is therefore uniquely maximized when $\Theta_\lambda^{(0)}\!=\!\Theta^\ast$, which occurs at $\lambda^\ast\!\in\!\Lambda$. Hence $\max_{\lambda\in\Lambda}\gamma^{(0)}(\lambda)=\gamma^{(0)}(\lambda^\ast)=(1/(\gamma_{\mathrm{ft}}^{\mathrm{pad},(0)})^{2}+1/\gamma_{\mathrm{pre}}^{2})^{-1/2}$, proving~\eqref{appeqn:gen_tradeoff} for both $\bar K\!=\!K$ and $\bar K\!<\!K$.
\end{proof}

\begin{rem}[Explicit bounds on $\bar E$ and $\bar C$]\label{apprem:explicit_bounds}
The constants in Corollary~\ref{appcor:gen_tradeoff} can be chosen with the following explicit upper bounds. Set $\bar E\!:=\!\sup_{\lambda\in\Lambda}E_\lambda$ and $\Theta_{\max}^{(0)}\!:=\!\max_{\lambda\in\Lambda}\Theta_\lambda^{(0)}$. Since $a\!:=\!\min\Lambda\!>\!0$, the proof of Theorem~\ref{appthm:general} uses radii bounded by $(\log K+2)/a$. Thus $\mu_0(\lambda)$ can be chosen uniformly bounded below by a positive constant and $C_0(\lambda)$ uniformly bounded above. Write $\kappa_\Lambda\!:=\!\sup_{\lambda\in\Lambda}C_0(\lambda)/\mu_0(\lambda)\!<\!\infty$.

\myparagraph{Bound on $\bar E$} For the conversion to an epsilon-only bound, assume the uniform positive singular-value lower bound discussed in Remark~\ref{apprem:S_bound}. Combining the definition of $E_\lambda$ in~\eqref{appeqn:gen_forget_E_def} with $\delta=(\bar Kn-1)\,\epsilon_1$ from Theorem~\ref{appthm:general} and the explicit bound $\|S\|_F\le\sqrt{\bar Kn\,N_{\mathrm{pre}}}\,\rho_{\mathrm{pre}}\,\epsilon_2/\sigma_{\min}(X_{\mathrm{pre}})$ from that remark,
\begin{align}\label{appeqn:explicit_E_bound}
    \bar E\le\kappa_\Lambda\biggl(\frac{\sqrt{\bar Kn\,N_{\mathrm{pre}}}\,\rho_{\mathrm{pre}}}{\sigma_{\min}(X_{\mathrm{pre}})}\,\epsilon_2+(\bar Kn-1)\bigl(1+\Theta_{\max}^{(0)}\bigr)\,\epsilon_1\biggr)\,,
\end{align}
which is linear in $\epsilon_1$ and $\epsilon_2$.

\myparagraph{Bound on $\bar C$} The constant $C$ in the proof of Theorem~\ref{appthm:general_forget} is built from three sources:
\begin{enumerate}[label=(\roman*)]
    \item the direct leakage bound~\eqref{appeqn:leak_ft} has coefficient $1$ in $\|S\|_F$, the projection bound~\eqref{appeqn:gen_forget_cross_bound} contributes $L_{\mathrm{proj}}\!:=\!\sqrt2$, and~\eqref{appeqn:leak_pre_total} contributes $L_{\mathrm{ft}}\!:=\!\sqrt{2\bar Kn}$, uniformly since $\delta\!\le\!1/2$;
    \item Cauchy--Schwarz over $K$ classes contributes a factor $\sqrt{K}$;
    \item the unit-norm input bound $|\gamma(W;\mathcal D)|\!\le\!\sqrt2$ and~\eqref{appeqn:gen_forget_ratio} contribute a factor at most $(1+\sqrt2)/\rho_{\mathrm{pre}}$ when converting numerator and norm errors to normalized-margin errors.
\end{enumerate}
Using $E_\lambda\epsilon_2\!\le\!E_\lambda$ in~\eqref{appeqn:leak_pre_total} and absorbing universal numerical factors into $C_*\!>\!0$,
\begin{align}\label{appeqn:explicit_C_bound}
    \bar C\le\frac{C_*\,\sqrt{K}\,(1+L_{\mathrm{proj}}+L_{\mathrm{ft}})}{\rho_{\mathrm{pre}}}\,\bigl(1+\rho_{\mathrm{pre}}+\Theta_{\max}^{(0)}\bigr)\,.
\end{align}
Under the stated conditioning requirement, substituting~\eqref{appeqn:explicit_E_bound} and~\eqref{appeqn:explicit_C_bound} into the right-hand side of~\eqref{appeqn:gen_tradeoff} yields a fully explicit linear-in-$(\epsilon_1,\epsilon_2)$ bound on the gap between the near-orthogonal optimum and the strict-orthogonal value.
\end{rem}


\section{Proof of Theorem~\ref{thm:lora_merge}}\label{appsec:proof_merge}
In this section, we first present the proof of Theorem~\ref{thm:lora_merge}.
Then, under additional smallness assumptions on $\epsilon_1,\epsilon_2$, we show that the same conclusions still hold, with additional error terms that depend on $\epsilon_1,\epsilon_2$.

\subsection{Proof of Theorem~\ref{thm:lora_merge}}
\begin{proof}
    For $\boldsymbol\alpha\in\mathbb R_{\geq0}^T$, following the same argument as in Appendix~\ref{appsec:proof_forget}, one can show that the margin of linear classifier $W_{\mathrm{LoRA}}$ for each task is
    \begin{align}
    \gamma\bigl(W_{\mathrm{LoRA}}(\boldsymbol\alpha);\mathcal D_{\mathrm{pre}}\bigr)
      &=\;
        \gamma_{\mathrm{pre}}\,
        \frac{\rho_{\mathrm{pre}}}
             {\sqrt{\rho_{\mathrm{pre}}^{2}
                    +\sum_{j=1}^{T}\alpha_j^2\Theta_{\lambda,j}^{2}}},\\
    \gamma\bigl(W_{\mathrm{LoRA}}(\boldsymbol\alpha);\mathcal D_{i}\bigr)
      &=\;
        \gamma_{i}\,
        \frac{\alpha_i\Theta_{\lambda,i}}
             {\sqrt{\rho_{\mathrm{pre}}^{2}
                    +\sum_{j=1}^{T}\alpha_j^2\Theta_{\lambda,j}^{2}}},
        \quad i=1,\dots ,T.
  \end{align}
    Then, our goal is to solve the following optimization problem
    \begin{align*}
        \max_{\alpha_1,\cdots,\alpha_T\geq0} \min\biggl(\gamma_{\mathrm{pre}}\,
        \frac{\rho_{\mathrm{pre}}}
             {\sqrt{\rho_{\mathrm{pre}}^{2}
                    +\sum_{j=1}^{T}\alpha_j^2\Theta_{\lambda,j}^{2}}}\,,\quad \min_{i\leq T} \gamma_{i}\,
        \frac{\alpha_i\Theta_{\lambda,i}}{\sqrt{\rho_{\mathrm{pre}}^{2} +\sum_{j=1}^{T}\alpha_j^2\Theta_{\lambda,j}^{2}}}\biggr)
    \end{align*}
    For convenience, we introduce the following notation $x_i=\alpha_i\Theta_{\lambda,i}\geq0, i\in[T]$, then the above optimization problem can be rewritten as
    \begin{align}\label{appeqn:primal}
        \max_{x_1,\cdots,x_T\geq0} \min\biggl(\frac{\rho_{\mathrm{pre}} \gamma_{\mathrm{pre}}}{\sqrt{\rho_{\mathrm{pre}}^2+\sum_{j=1}^T x_j^2}},\quad \min_{i\in[T]} \frac{\gamma_i x_i}{\sqrt{\rho_{\mathrm{pre}}^2+\sum_{j=1}^T x_j^2}}\biggl)\,.
    \end{align}
    Let $S:=\sqrt{\rho_{\mathrm{pre}}^2+\sum_{j=1}^T x_j^2}$ and $t:=\min\{\frac{\gamma_ix_i}{S}, \frac{\gamma_{\mathrm{pre}} \rho_{\mathrm{pre}}}{S}\}$, then Problem~\ref{appeqn:primal} is equivalent to
    \begin{align}\label{appeqn:dual}
        \max_{t, S, x_1,\cdots, x_T} &t \\
        \text{s.t.} \quad&x_i\geq0,\quad\gamma_ix_i\geq tS,\quad i\in [T] \nonumber\\
        & \gamma_{\mathrm{pre}} \rho_{\mathrm{pre}} \geq tS\,,\nonumber\\
        & \sum_{i=1}^T x_i^2 + \rho_{\mathrm{pre}}^2 =S^2, \quad S>0 \nonumber \,.
    \end{align}
    We first claim at optimum, the inequalities w.r.t. $x_i$ will be tight, i.e., $x_i=\frac{tS}{\gamma_i}, \forall i\in[T]$. In this case, one can see that
    \begin{align*}
        S^2 = \rho_{\mathrm{pre}}^2+\sum_{j=1}^T x_j^2 \geq S^2t^2 \cdot \biggl(\sum_{i=1}^T\frac{1}{\gamma_i^2} + \frac{1}{\gamma_{\mathrm{pre}}^2}\biggl)\,.
    \end{align*}
    Thus, the optimal value of the objective is $t\leq\frac{1}{\textstyle
            \sqrt{\frac{1}{\gamma_{\mathrm{pre}}^{2}}
                  +\sum_{j=1}^{T}\frac{1}{\gamma_{j}^{2}}}}$, and the equality is achieved under the condition $\gamma_i x_i=\gamma_{\mathrm{pre}} \rho_{\mathrm{pre}}$, which is equivalent to $\alpha_i = \frac{\rho_{\mathrm{pre}}\gamma_{\mathrm{pre}}}{\gamma_i\Theta_{\lambda,i}},
    \; i=1,\dots ,T$.

    Finally, we show that at the optimum of Problem~\ref{appeqn:dual}, both the adapter and pre-training inequalities are tight.

    \myparagraph{Case 1: some adapter inequality is slack} Assume for the optimal solution $(t^*, S^*, \boldsymbol x^*)$ there exists $j\in [T]$ such that $\gamma_j x_j^*>t^* S^*$. Decrease $x_j^*$ slightly to $x_j^*-\epsilon$ with $\epsilon>0$. The new $S$ is
    \begin{align*}
        \tilde S=\sqrt{\rho_{\mathrm{pre}}^2+\sum_{i\not=j} (x_i^*)^2 + (x_j^*-\epsilon)^2} < S^*\,.
    \end{align*}
    For $\epsilon$ sufficiently small, every other adapter inequality remains satisfied (since $\tilde S<S^*$), the pre-training inequality remains satisfied ($\gamma_{\mathrm{pre}}\rho_{\mathrm{pre}}\ge t^* S^*>t^*\tilde S$), and $\gamma_j(x_j^*-\epsilon) > t^* \tilde S$ by continuity. Therefore one can increase $t^*$ until one inequality becomes tight, raising the objective and contradicting optimality.

    \myparagraph{Case 2: all adapter inequalities are tight but the pre-training inequality is slack} Assume $\gamma_i x_i^* = t^* S^*$ for every $i\in[T]$ but $\gamma_{\mathrm{pre}}\rho_{\mathrm{pre}}>t^* S^*$. Take $\tilde x_i := c x_i^*$ for some $c>1$, leaving $\rho_{\mathrm{pre}}$ unchanged. Then
    \begin{align*}
        \tilde S=\sqrt{\rho_{\mathrm{pre}}^2+c^2\sum_i (x_i^*)^2}=\sqrt{c^2(S^*)^2-\rho_{\mathrm{pre}}^2(c^2-1)}<c\,S^*\,.
    \end{align*}
    Setting $\tilde t := c t^* S^*/\tilde S$ keeps all adapter inequalities tight (since $\gamma_i\tilde x_i=c\gamma_i x_i^*=c t^* S^*=\tilde t\tilde S$), and $\tilde t > t^*$ from $\tilde S < cS^*$. For $c$ close enough to $1$, the pre-training inequality $\gamma_{\mathrm{pre}}\rho_{\mathrm{pre}}\ge\tilde t\tilde S=ct^* S^*$ remains satisfied by continuity. Hence $(\tilde t, \tilde S, \tilde{\boldsymbol x})$ is feasible with $\tilde t > t^*$, again contradicting optimality.

    Combining the two cases, all inequalities must be tight at any optimum.
\end{proof}

\subsection{Proof of General Version}\label{appsec:general_version_merge}
In this subsection we present the general version of Theorem~\ref{thm:lora_merge}: under a near-orthogonal multi-task assumption (defined below), the margins of the merged model $W_{\mathrm{LoRA}}(\boldsymbol\alpha)$ on the pre-training data and on each fine-tuning task remain close to the strict-orthogonal closed-form values from Theorem~\ref{thm:lora_merge}, with deviations controlled by the perturbation error $E$ defined below. Consequently, the optimal mixing coefficients of Theorem~\ref{thm:lora_merge} continue to (approximately) maximize the merged margin.

\myparagraph{Near-orthogonal multi-task assumption} Let $X_i\in\R^{d\times \bar K_in_i}$ denote the data matrix of task $i\in[T]$. We replace Assumption~\ref{asmp:merge} by the following near-orthogonal version: there exist $\epsilon_1,\epsilon_2\in[0,1)$ such that
\begin{enumerate}[label=(\alph*),leftmargin=*]
    \item Every fine-tuning sample has unit norm: $\|\bar x\|_2=1$ for every $\bar x\in\bigcup_{i=1}^T\mathcal D_i$.
    \item \emph{Within-task:} for each $i\in[T]$ and each pair of distinct $\bar x,\bar x'\in\mathcal D_i$, $|\bar x^\top \bar x'|\le\epsilon_1$.
    \item \emph{Cross-task and pre-training:} for each $i\ne j$ in $[T]$, every $\bar x\in\mathcal D_i$ and $\bar x'\in\mathcal D_j$ satisfy $|\bar x^\top \bar x'|\le\epsilon_2$; and for every $\bar x\in\bigcup_i\mathcal D_i$ and every $x\in\mathcal D_{\mathrm{pre}}$, $|\bar x^\top x|\le\epsilon_2$.
\end{enumerate}
The strict-orthogonal Assumption~\ref{asmp:merge} is recovered at $\epsilon_1=\epsilon_2=0$.

\myparagraph{Notation} For each task $i\in[T]$ we adopt the per-task analogues of the notation of Theorem~\ref{appthm:general}: $G_i:=X_i^\top X_i$, $R_i:=G_i^{1/2}$, $Q_i:=X_iR_i^{-1}$, $S_i:=W_{\mathrm{pre}}X_i$, $\delta_i:=(\bar K_in_i-1)\epsilon_1$, and the unique optimum $Z_{\lambda,i}^{\ast}(\bar S,\bar R)$ of $\Phi_i(Z;\bar S,\bar R)$ defined as in Theorem~\ref{appthm:general} but with $X_i,\bar K_i,n_i$ in place of $X_{\mathrm{ft}},\bar K,n$. We take $\delta_0,s_0,\mu_0$ as the minima of their per-task counterparts and $C_0$ as the maximum. By the per-task reformulation, $B_i^{\ast}A_i^{\ast}=Z_{\lambda,i}^{\ast}(S_i,R_i)Q_i^\top$, where the rank constraint $\mathrm{rank}(Z)\le r$ is non-binding for $\delta_i,\|S_i\|_F$ sufficiently small by Lemma~\ref{applem:rank_stability} applied per task. Write
\begin{align}
    \Theta_{\lambda,i}^{(0)}\!:=\!\|Z_{\lambda,i}^{\ast}(0,I)\|_F\,,\qquad \gamma_i^{(0)}\!:=\!\bigl.\gamma_i\bigr|_{\epsilon_1=\epsilon_2=0}\,,
\end{align}
for the strict-orthogonal per-task adapter Frobenius norm and per-task margin (so $\Theta_{\lambda,i}^{(0)}=\Theta_{\lambda,i}$ and $\gamma_i^{(0)}=\gamma_i$ from Theorem~\ref{thm:lora_merge} when $\epsilon_1=\epsilon_2=0$). Set
\begin{align}
    E_{\lambda,i}\!:=\!\frac{C_0}{\mu_0}\bigl(\|S_i\|_F+\delta_i(1+\Theta_{\lambda,i}^{(0)})\bigr)\,,\qquad E\!:=\!\max_{i\in[T]}E_{\lambda,i}+\epsilon_2\,.
\end{align}
The common constants satisfy $C_0/\mu_0\!\ge\!1$, so $\|S_i\|_F\!\le\!E_{\lambda,i}\!\le\!E$. Under a uniform positive lower bound on $\sigma_{\min}(X_{\mathrm{pre}})$, $\|S_i\|_F=O(\epsilon_2)$ uniformly in $i$ (Remark~\ref{apprem:S_bound}), so $E\to 0$ as $\epsilon_1,\epsilon_2\to 0$.

\begin{thm}[General version of Theorem~\ref{thm:lora_merge}]\label{appthm:general_merge}
Adopt the near-orthogonal multi-task assumption above with $\delta_i\!\le\!\delta_0$ and $\|S_i\|_F\!\le\!s_0$ for every $i\in[T]$, using the common thresholds chosen above. Suppose each task is trained in the intermediate or low-penalty regime, and the pre-training samples have unit norm. There exist constants $E_0\in(0,1]$ and $C>0$, depending on $\lambda,K,\{\bar K_i,n_i\}_{i=1}^T,\rho_{\mathrm{pre}},\mu_0,C_0,\{\|Z_{\lambda,i}^{\ast}(0,I)\|_*\}_{i=1}^T$ but not on $\epsilon_1,\epsilon_2$ or $\boldsymbol\alpha$, such that whenever $E\leq E_0$, the following bounds hold uniformly for all nonnegative coefficient vectors $\boldsymbol\alpha\in\mathbb R_{\geq0}^T$:
\begin{align}\label{appeqn:gen_merge_pre}
    \Bigl|\gamma\bigl(W_{\mathrm{LoRA}}(\boldsymbol\alpha);\mathcal D_{\mathrm{pre}}\bigr)-\frac{\gamma_{\mathrm{pre}}\,\rho_{\mathrm{pre}}}{\sqrt{\rho_{\mathrm{pre}}^2+\sum_{i=1}^T\alpha_i^2(\Theta_{\lambda,i}^{(0)})^2}}\Bigr|&\le CE\,,\\
    \Bigl|\gamma\bigl(W_{\mathrm{LoRA}}(\boldsymbol\alpha);\mathcal D_j\bigr)-\frac{\gamma_j^{(0)}\,\alpha_j\,\Theta_{\lambda,j}^{(0)}}{\sqrt{\rho_{\mathrm{pre}}^2+\sum_{i=1}^T\alpha_i^2(\Theta_{\lambda,i}^{(0)})^2}}\Bigr|&\le CE\,,\quad j\in[T]\,.\label{appeqn:gen_merge_task}
\end{align}
Write $F(\boldsymbol\alpha):=\gamma(W_{\mathrm{LoRA}}(\boldsymbol\alpha);\mathcal D_{\mathrm{pre}}\cup\bigcup_i\mathcal D_i)$ and let $F^{(0)}(\boldsymbol\alpha)$ be the minimum of the reference margins displayed above. The $1$-Lipschitz property of $\min$ gives $|F(\boldsymbol\alpha)-F^{(0)}(\boldsymbol\alpha)|\leq CE$ uniformly over $\boldsymbol\alpha\geq0$. In particular, the strict-orthogonal optimum $\alpha_i^{\ast}=\rho_{\mathrm{pre}}\gamma_{\mathrm{pre}}/(\gamma_i^{(0)}\Theta_{\lambda,i}^{(0)})$ satisfies
\begin{align*}
    0\leq\sup_{\boldsymbol\alpha\in\mathbb R_{\geq0}^T}F(\boldsymbol\alpha)-F(\boldsymbol\alpha^\ast)\leq2CE.
\end{align*}
Thus $\boldsymbol\alpha^\ast$ remains optimal up to an additive $O(E)$ error in achieved margin, and all margins converge uniformly to their strict-orthogonal reference values as $E\to0$.
\end{thm}

\begin{proof}
Throughout, $C$ denotes a positive constant whose value may change line to line, but always depends only on the quantities listed in the theorem statement. We work with $E\leq1$ and choose $E_0\leq1$ below.

We first establish leakage bounds. Since $S_i\!=\!W_{\mathrm{pre}}X_i$, for every task $i\in[T]$, sample $\bar x_j^{(i)}\in\mathcal D_i$, and row $w_k$ of $W_{\mathrm{pre}}$,
\begin{align}\label{appeqn:gen_merge_leak_pre_to_ft}
    |w_k^\top\bar x_j^{(i)}|=|(S_i)_{kj}|\le\|S_i\|_F\le E_{\lambda,i}\,.
\end{align}
For each $i$, write $\delta_{\lambda,i,k}:=(B_i^{\ast}A_i^{\ast})_k^\top$ for the $k$-th row of $B_i^{\ast}A_i^{\ast}$. Using $B_i^{\ast}A_i^{\ast}=Z_{\lambda,i}^{\ast}(S_i,R_i)Q_i^\top$, we obtain (i) for any unit-norm $x\in\mathcal D_{\mathrm{pre}}$,
\begin{align}\label{appeqn:gen_merge_leak_ft_to_pre}
    |\delta_{\lambda,i,k}^\top x|\le\|(Z_{\lambda,i}^{\ast}(S_i,R_i))_k\|_2\,\|Q_i^\top x\|_2\le C\,\epsilon_2\,,
\end{align}
where the last step uses $\|Q_i^\top x\|_2\le\sqrt{\bar K_in_i}\,\epsilon_2/\sqrt{1-\delta_i}$ (from $|\bar x_l^{(i),\top}x|\le\epsilon_2$) and $\|Z_{\lambda,i}^{\ast}(S_i,R_i)\|_F\le\Theta_{\lambda,i}^{(0)}+E_{\lambda,i}\le C$. (ii) For any $\bar x\in\mathcal D_j$ with $j\ne i$, the same argument with $X_j$ in place of $\{x\}$ gives
\begin{align}\label{appeqn:gen_merge_leak_cross}
    |\delta_{\lambda,i,k}^\top\bar x|\le C\,\epsilon_2\,,
\end{align}
since each entry of $X_i^\top\bar x$ is bounded by $\epsilon_2$ (cross-task assumption). The adapter leakage bounds are $O(\epsilon_2)$ uniformly in $i,k$, and the pre-trained classifier leakage is controlled by $\|S_i\|_F$.

Then, we bound the perturbation of the merged-model Frobenius norm. Expanding,
\begin{align}\label{appeqn:gen_merge_Wnorm_expand}
    \|W_{\mathrm{LoRA}}(\boldsymbol\alpha)\|_F^2=\rho_{\mathrm{pre}}^2+\sum_{i=1}^T\alpha_i^2\,\|B_i^{\ast}A_i^{\ast}\|_F^2+2\sum_{i=1}^T\alpha_i\bigl\langle W_{\mathrm{pre}},B_i^{\ast}A_i^{\ast}\bigr\rangle+\sum_{i\ne j}\alpha_i\alpha_j\bigl\langle B_i^{\ast}A_i^{\ast},B_j^{\ast}A_j^{\ast}\bigr\rangle\,.
\end{align}
The identity $W_{\mathrm{pre}}Q_i\!=\!S_iR_i^{-1}$ and the same projection argument as in the proof of Theorem~\ref{appthm:general_forget} give
\begin{align}\label{appeqn:gen_merge_pre_cross}
    \bigl|\bigl\langle W_{\mathrm{pre}},B_i^{\ast}A_i^{\ast}\bigr\rangle\bigr|\le\frac{\|S_i\|_F\,\|B_i^{\ast}A_i^{\ast}\|_F}{\sqrt{1-\delta_i}}\le C\,(\Theta_{\lambda,i}^{(0)}+E_{\lambda,i})\,\|S_i\|_F\,.
\end{align}
For the cross-task term with $i\ne j$, each row $(B_i^{\ast}A_i^{\ast})_k$ lies in $\mathrm{span}(X_i)$ and $(B_j^{\ast}A_j^{\ast})_k\in\mathrm{span}(X_j)$. Writing $u\in\mathrm{span}(X_i)$ as $u=X_i\alpha$ with $\|\alpha\|_2\le\|u\|_2/\sigma_{\min}(R_i)\le\|u\|_2/\sqrt{1-\delta_i}\le C\,\|u\|_2$ (and similarly for $v=X_j\beta$), we have $u^\top v=\alpha^\top X_i^\top X_j\beta$ with each entry of $X_i^\top X_j$ bounded by $\epsilon_2$ in absolute value, hence $|u^\top v|\le\|X_i^\top X_j\|_{\mathrm{op}}\,\|\alpha\|_2\,\|\beta\|_2\le C\,\epsilon_2\,\|u\|_2\,\|v\|_2$. Summing this bound over rows $k$ and applying Cauchy--Schwarz,
\begin{align}\label{appeqn:gen_merge_task_cross}
    \bigl|\bigl\langle B_i^{\ast}A_i^{\ast},B_j^{\ast}A_j^{\ast}\bigr\rangle\bigr|\le C\,\epsilon_2\,\|B_i^{\ast}A_i^{\ast}\|_F\,\|B_j^{\ast}A_j^{\ast}\|_F\le C\,\epsilon_2\,(\Theta_{\lambda,i}^{(0)}+E_{\lambda,i})(\Theta_{\lambda,j}^{(0)}+E_{\lambda,j})\,.
\end{align}
Set $a_{\boldsymbol\alpha}:=\|W_{\mathrm{LoRA}}(\boldsymbol\alpha)\|_F$ and $b_{\boldsymbol\alpha}^2:=\rho_{\mathrm{pre}}^2+\sum_i\alpha_i^2(\Theta_{\lambda,i}^{(0)})^2$. Combining \eqref{appeqn:gen_merge_pre_cross}, \eqref{appeqn:gen_merge_task_cross} with the per-task bound $\bigl|\|B_i^{\ast}A_i^{\ast}\|_F-\Theta_{\lambda,i}^{(0)}\bigr|\le E_{\lambda,i}+O(\delta_i\Theta_{\lambda,i}^{(0)})$ from Theorem~\ref{appthm:general} and using $E\leq1$ yields, for some $D>0$ independent of $\boldsymbol\alpha,\epsilon_1,\epsilon_2$,
\begin{align*}
    |a_{\boldsymbol\alpha}^2-b_{\boldsymbol\alpha}^2|\leq D E(1+\|\boldsymbol\alpha\|_2^2).
\end{align*}
Here the linear and mixed coefficient terms are controlled using $\|\boldsymbol\alpha\|_1\leq\sqrt T\|\boldsymbol\alpha\|_2$ and $\|\boldsymbol\alpha\|_1^2\leq T\|\boldsymbol\alpha\|_2^2$. The stated regimes ensure $\Theta_{\lambda,i}^{(0)}>0$ for every $i$, so
\begin{align*}
    c:=\min\{\rho_{\mathrm{pre}},\min_i\Theta_{\lambda,i}^{(0)}\}>0,\qquad
    b_{\boldsymbol\alpha}^2\geq c^2(1+\|\boldsymbol\alpha\|_2^2).
\end{align*}
Thus, setting $\kappa:=D/c^2$, we obtain the uniform relative bound
\begin{align}\label{appeqn:gen_merge_Wnorm_bound}
    |a_{\boldsymbol\alpha}^2-b_{\boldsymbol\alpha}^2|\leq\kappa E\,b_{\boldsymbol\alpha}^2.
\end{align}
Choose $E_0:=\min\{1,(2\kappa)^{-1}\}$. For $E\leq E_0$, this gives, uniformly over $\boldsymbol\alpha\geq0$,
\begin{align*}
    a_{\boldsymbol\alpha}\geq\frac{b_{\boldsymbol\alpha}}{\sqrt2}>0,\qquad
    \frac{|a_{\boldsymbol\alpha}-b_{\boldsymbol\alpha}|}{b_{\boldsymbol\alpha}}
    =\frac{|a_{\boldsymbol\alpha}^2-b_{\boldsymbol\alpha}^2|}{b_{\boldsymbol\alpha}(a_{\boldsymbol\alpha}+b_{\boldsymbol\alpha})}
    \leq\kappa E.
\end{align*}

Next, we bound the perturbation of the unnormalized margins on each dataset. For $(x,y_c)\in\mathcal D_{\mathrm{pre}}$ and $k\ne c$, decomposing
\begin{align}\label{appeqn:gen_merge_pre_decomp}
    \Bigl[\bigl(w_c+\textstyle\sum_i\alpha_i\delta_{\lambda,i,c}\bigr)-\bigl(w_k+\textstyle\sum_i\alpha_i\delta_{\lambda,i,k}\bigr)\Bigr]^\top x=(w_c-w_k)^\top x+\sum_{i=1}^T\alpha_i(\delta_{\lambda,i,c}-\delta_{\lambda,i,k})^\top x\,,
\end{align}
and applying \eqref{appeqn:gen_merge_leak_ft_to_pre} to each adapter term,
\begin{align}\label{appeqn:gen_merge_C_pre}
    \bigl|C(W_{\mathrm{LoRA}}(\boldsymbol\alpha);\mathcal D_{\mathrm{pre}})-\rho_{\mathrm{pre}}\gamma_{\mathrm{pre}}\bigr|\le C\,\|\boldsymbol\alpha\|_1\,\epsilon_2\,.
\end{align}
For $(\bar x,y_c)\in\mathcal D_j$ and $k\ne c$,
\begin{align}\label{appeqn:gen_merge_task_decomp}
    \bigl[\cdots\bigr]^\top\bar x=\alpha_j(\delta_{\lambda,j,c}-\delta_{\lambda,j,k})^\top\bar x+\sum_{i\ne j}\alpha_i(\delta_{\lambda,i,c}-\delta_{\lambda,i,k})^\top\bar x+(w_c-w_k)^\top\bar x\,.
\end{align}
The first term in \eqref{appeqn:gen_merge_task_decomp} is the within-task contribution. Since $\alpha_j\geq0$, it can be taken outside the minimum. By the same argument as in the proof of Theorem~\ref{appthm:general_forget} (Step~3 there), $\bigl|\min(w.r.t.\,(\bar x,c,k))\,\alpha_j(\delta_{\lambda,j,c}-\delta_{\lambda,j,k})^\top\bar x-\alpha_j\,\Theta_{\lambda,j}^{(0)}\gamma_j^{(0)}\bigr|\le C\,|\alpha_j|\,(E_{\lambda,j}+\delta_j)$. The cross-task adapter terms are bounded via \eqref{appeqn:gen_merge_leak_cross}: $|\alpha_i(\delta_{\lambda,i,c}-\delta_{\lambda,i,k})^\top\bar x|\le 2\,C\,|\alpha_i|\,\epsilon_2$, and the pre-training leakage on $\mathcal D_j$ contributes at most $2\|S_j\|_F\!\le\!2E_{\lambda,j}$ via \eqref{appeqn:gen_merge_leak_pre_to_ft}. Combining via the triangle inequality,
\begin{align}\label{appeqn:gen_merge_C_task}
    \bigl|C(W_{\mathrm{LoRA}}(\boldsymbol\alpha);\mathcal D_j)-\alpha_j\,\Theta_{\lambda,j}^{(0)}\gamma_j^{(0)}\bigr|\le C\,(1+\|\boldsymbol\alpha\|_1)\,E\,,
\end{align}
where we have absorbed $\delta_j$ into $E$ via $\delta_j\le(\mu_0/C_0)E_{\lambda,j}\le(\mu_0/C_0)E$.

Finally, fix any dataset $\mathcal D\in\{\mathcal D_{\mathrm{pre}},\mathcal D_1,\dots,\mathcal D_T\}$ and set $u_{\boldsymbol\alpha}:=C(W_{\mathrm{LoRA}}(\boldsymbol\alpha);\mathcal D)$. Its reference numerator $u_{\boldsymbol\alpha}^{(0)}$ is $\rho_{\mathrm{pre}}\gamma_{\mathrm{pre}}$ for $\mathcal D_{\mathrm{pre}}$, and $\alpha_j\Theta_{\lambda,j}^{(0)}\gamma_j^{(0)}$ for $\mathcal D_j$. Equations~\eqref{appeqn:gen_merge_C_pre}--\eqref{appeqn:gen_merge_C_task} give $|u_{\boldsymbol\alpha}-u_{\boldsymbol\alpha}^{(0)}|\leq C(1+\|\boldsymbol\alpha\|_1)E$. Since
\begin{align*}
    \frac{1+\|\boldsymbol\alpha\|_1}{b_{\boldsymbol\alpha}}
    \leq\frac{\sqrt{T+1}\sqrt{1+\|\boldsymbol\alpha\|_2^2}}{b_{\boldsymbol\alpha}}
    \leq\frac{\sqrt{T+1}}{c},
\end{align*}
the numerator error divided by $b_{\boldsymbol\alpha}$ is uniformly $O(E)$. All inputs have unit norm and $a_{\boldsymbol\alpha}>0$, so $|u_{\boldsymbol\alpha}/a_{\boldsymbol\alpha}|\leq\sqrt2$. Applying the ratio estimate~\eqref{appeqn:gen_forget_ratio} and the relative norm bound above,
\begin{align*}
    \left|\frac{u_{\boldsymbol\alpha}}{a_{\boldsymbol\alpha}}-\frac{u_{\boldsymbol\alpha}^{(0)}}{b_{\boldsymbol\alpha}}\right|
    \leq\frac{|u_{\boldsymbol\alpha}-u_{\boldsymbol\alpha}^{(0)}|+\sqrt2\,|a_{\boldsymbol\alpha}-b_{\boldsymbol\alpha}|}{b_{\boldsymbol\alpha}}
    \leq CE.
\end{align*}
This proves \eqref{appeqn:gen_merge_pre}--\eqref{appeqn:gen_merge_task} uniformly in $\boldsymbol\alpha$. Taking the minimum across datasets gives $|F(\boldsymbol\alpha)-F^{(0)}(\boldsymbol\alpha)|\leq CE$ uniformly.

For the optimality claim, Theorem~\ref{thm:lora_merge} gives $F^{(0)}(\boldsymbol\alpha)\leq F^{(0)}(\boldsymbol\alpha^\ast)$ for every $\boldsymbol\alpha\geq0$. Therefore,
\begin{align*}
    F(\boldsymbol\alpha)
    \leq F^{(0)}(\boldsymbol\alpha)+CE
    \leq F^{(0)}(\boldsymbol\alpha^\ast)+CE
    \leq F(\boldsymbol\alpha^\ast)+2CE.
\end{align*}
Taking the supremum over all nonnegative coefficient vectors proves the stated $2CE$ near-optimality bound.
\end{proof}


\section{Lemmas}\label{appsec:lemmas}
This section first lists the preliminary lemmas (cited from external sources) used throughout the proofs, and then gives full proofs of the paper-specific supporting lemmas.

\subsection{Preliminary Lemmas}
In this subsection, we provide several preliminary lemmas that will be used in the proof.

\begin{lem}[Theorem 3 in \cite{riedel1992sherman}]
    Let $A\in\R^{l\times l}$ with $\mathrm{rank}(A)<l$, and $v_i, w_i, i=1,2$ be vectors in $\R^l$.
    Let $v_1\in M(A), w_1\perp M(A)$, and $v_2\in M(A^{\ast}), w_2\perp M(A^{\ast})$, where $M(A)$ denotes the range of $A$.
    Assume $w_2 \parallel w_1$ and $w_i\not= 0, i=1,2$.
    Let $\Omega=A+(v_1+w_1)(v_2+w_2)^{\ast}$, Then
    \begin{align*}
        \Omega^{\dagger}=A^{\dagger}-\frac{w_2v_2^{\ast} A^{\dagger}}{\|w_2\|^2}-\frac{A^{\dagger}v_1w_1^{\ast}}{\|w_1\|^2}+(1+v_2^{\ast}A^{\dagger}v_1)\cdot \frac{w_2w_1^{\ast}}{\|w_1\|^2\|w_2\|^2}\,.
    \end{align*}
\end{lem}
We refer the readers to \cite{riedel1992sherman} for the proof.

\begin{lem}[Variational form of the nuclear norm]\label{applem:varitional_nuclear}
    For any fixed $Z\in\R^{K\times m}$, we have
    \begin{align*}
        \|Z\|_*=\min_{Z=BA} \frac{1}{2}(\|B\|_F^2+\|A\|_F^2)\,.
    \end{align*}
\end{lem}
We refer the readers to \cite{recht2010guaranteed} for the proof.

\begin{lem}\label{applem:dual_nuclear}
    For any $Z\in\R^{K\times m}$,
    \begin{align*}
        \|Z\|_*=\max_{V\in\R^{m\times K},\,\|V\|_{\mathrm{op}}\le 1}\Tr(ZV)\,.
    \end{align*}
    For any orthogonal projection $P\in\R^{K\times K}$ and any $Z\in\R^{K\times m}$,
    \begin{align*}
        \|PZ\|_*\le \|P\|_{\mathrm{op}}\,\|Z\|_*\le\|Z\|_*\,,
    \end{align*}
    Equality $\|PZ\|_*=\|Z\|_*$ holds if and only if $PZ=Z$.
    Moreover, for any two matrices $A, B\in\R^{K\times m}$, one has
    \begin{align*}
        \sum_{i=1}^{\min(K,m)} \sigma_i(AB^\top) \leq \sum_{i=1}^{\min(K,m)} \sigma_i(A)\sigma_i(B)\,.
    \end{align*}
\end{lem}

We refer the readers to \cite{horn2012matrix} for the proof.

\begin{lem}[Berge's Maximum Theorem]\label{applem:berge}
    Let $X, \Theta$ be topological spaces, $f:X\times \Theta \to \R$ be a continuous function on the product $X\times \Theta$, and $C:\Theta \to X$ be a compact-valued correspondence such that $C(\theta)\not=\emptyset$ for all $\theta\in\Theta$.
    Define the marginal function $f^{\ast}:\Theta\to\R$ and the set of minimizers $C^{\ast}:\Theta\to X$ by
    \begin{align*}
        f^{\ast}(\theta)&=\sup\{f(x,\theta): x\in C(\theta)\}\\
        C^{\ast}(\theta) &= \argmax \{f(x,\theta):x\in C(\theta)\}=\{x\in C(\theta): f(x;\theta)=f^{\ast}(\theta)\}\,.
    \end{align*}
    If $C$ is continuous at $\theta$, then the value function $f^{\ast}(\theta)$ is continuous, and the set of maximizers $C^{\ast}(\theta)$ is upper-hemicontinuous with nonempty and compact values. Moreover, if $C^{\ast}(\theta)$ is single-valued, and thus is a continuous function rather than a correspondence. 
\end{lem}
We refer the readers to \cite{sundaram1996first} for the proof.

\begin{lem}[Invariance under Permutation]\label{applem:permutation}
Suppose $X$ satisfies $\Pi X \Pi^\top =X$ for any permutation $\Pi$, then $X=a\boldsymbol I+c \boldsymbol 1 \boldsymbol 1^\top$ for any constants $a,c\in\R$.
\end{lem}
We refer the readers to Lemma 10 in \cite{hong2023neural} for the proof.

\begin{prop}[Representer Theorem for Hard-margin SVM Problem]\label{prop:max_margin}
    Let $\{(\boldsymbol x_{c,i},\boldsymbol y_c)\}_{i=1, c=1}^{n_{c}, K}$ be the dataset with features $\boldsymbol x_{c,i}\in\R^d$ and class labels $\boldsymbol y_c\in\R^K$. Then, let $W^{\mathrm{mm}}$ be the solution of the following hard-margin SVM problem, i.e.,
\[
    \min_{W\in\mathbb R^{K\times d}} \;
        \tfrac12\|W\|_{F}^{2}
    \quad\text{s.t.}\quad
    \forall\, i\in[K],\;
    \forall\, j\in[n_i],\;
    \forall\, k\neq i:\;
    \boldsymbol w_{i}^{\top} \boldsymbol x_{i,j} \;\ge\; \boldsymbol w_{k}^{\top}\boldsymbol x_{i,j} + 1\,,
\]
where $W=[\boldsymbol w_1,\cdots, \boldsymbol w_K]^\top$. 
Then, there exists a set of scalars $\{\alpha_{i,c,j}\}_{i=1,c=1,j=1}^{K, K, n_i}$ such that each row of $W^{\mathrm{mm}}$ has the following structure
\begin{align*}
    \boldsymbol w_{i}^{\mathrm{mm}}=\sum_{c=1}^{K}\sum_{j=1}^{n_c} \alpha_{i,c,j}\boldsymbol x_{c,j}\,.
\end{align*}
\end{prop}
We refer the readers to \cite{scholkopf2018learning} for the proof.
\subsection{Proof of Lemma~\ref{applem:nc_obj}}\label{app_proof:nc_obj}
\begin{proof}
By Proposition~\ref{prop:max_margin} and Assumption~\ref{asmp:pretrain_w}, each row of $W_{\mathrm{pre}}$ lies in the span of the pre-training features. Under Assumption~\ref{asmp:data}(c) with $\epsilon_2=0$, every fine-tuning feature is orthogonal to this span, so
\begin{align}
    W_{\mathrm{pre}} X_{\mathrm{ft}}=0\,.
\end{align}

Moreover, Assumption~\ref{asmp:data}(a)--(b) with $\epsilon_1=0$ gives $X_{\mathrm{ft}}^\top X_{\mathrm{ft}}=\boldsymbol I_{\bar{K}n}$.
We apply change of variable as follows: $\tilde A= AX_{\mathrm{ft}} \in\R^{r\times \bar{K}n}$.
\begin{align*}
    &\frac{1}{\bar{K}n}\sum_{c=1}^{\bar{K}} \sum_{j=1}^n \mathcal{L}_{\mathrm{CE}}\left(y_c, (W_{\mathrm{pre}}+BA)x_{c,j}\right)+\frac{\lambda}{2}\left(\|B\|_F^2+\|A\|_F^2\right) \\
    =& \frac{1}{\bar{K}n}\sum_{c=1}^{\bar{K}} \sum_{j=1}^n \mathcal{L}_{\mathrm{CE}}\left(y_c, W_{\mathrm{pre}}X_{\mathrm{ft}} e_{(c-1)n+j}+ BA X_{\mathrm{ft}} e_{(c-1)n+j}\right)+\frac{\lambda}{2}\left(\|B\|_F^2+\|A\|_F^2\right) \\
    =& \frac{1}{\bar{K}n}\sum_{c=1}^{\bar{K}} \sum_{j=1}^n \mathcal{L}_{\mathrm{CE}}\left(y_c, W_{\mathrm{pre}}X_{\mathrm{ft}} e_{(c-1)n+j}+ BA QR e_{(c-1)n+j}\right)+\frac{\lambda}{2}\left(\|B\|_F^2+\|AQ\|_F^2+\|AQ_{\perp}\|_F^2\right) \\
    =& \frac{1}{\bar{K}n}\sum_{c=1}^{\bar{K}} \sum_{j=1}^n\mathcal{L}_{\mathrm{CE}}\left(y_c, W_{\mathrm{pre}}X_{\mathrm{ft}} e_{(c-1)n+j}+ B\tilde A R e_{(c-1)n+j}\right)+\frac{\lambda}{2}\left(\|B\|_F^2+\|\tilde A\|_F^2 + \|AQ_{\perp}\|_F^2\right)\,.
\end{align*}
Under this transformation, the objective in~\eqref{eqn:ft_obj} can be rewritten as follows
\begin{align*}
    &\frac{1}{\bar{K}n}\sum_{c=1}^{\bar{K}} \sum_{j=1}^n \mathcal{L}_{\mathrm{CE}}\left(y_c, (W_{\mathrm{pre}}+BA)x_{c,j}\right)+\frac{\lambda}{2}\left(\|B\|_F^2+\|A\|_F^2\right) \\
    =& \frac{1}{\bar{K}n}\sum_{c=1}^{\bar{K}} \sum_{j=1}^n \mathcal{L}_{\mathrm{CE}}\left(y_c, BA X_{\mathrm{ft}} e_{(c-1)n+j}\right)+\frac{\lambda}{2}\left(\|B\|_F^2+\|A\|_F^2\right) \\
    =& \frac{1}{\bar{K}n}\sum_{c=1}^{\bar{K}} \sum_{j=1}^n\mathcal{L}_{\mathrm{CE}}\left(y_c, B\tilde A e_{(c-1)n+j}\right)+\frac{\lambda}{2}\left(\|B\|_F^2+\|A X_{\mathrm{ft}}\|_F^2 + \|AX_{\mathrm{ft},\perp}\|_F^2\right) \\
    =& \frac{1}{\bar{K}n}\sum_{c=1}^{\bar{K}} \sum_{j=1}^n\mathcal{L}_{\mathrm{CE}}\left(y_c, B\tilde A e_{(c-1)n+j}\right)+\frac{\lambda}{2}\left(\|B\|_F^2+\|\tilde A\|_F^2 + \|AX_{\mathrm{ft},\perp}\|_F^2\right)\,.
\end{align*}
Notice that the component \(AX_{\mathrm{ft},\perp}\) contributes only to the \emph{regularization} term and does not affect the cross‐entropy loss.  Hence at any global minimizer \((B^{\ast},A^{\ast})\) of \eqref{eqn:ft_obj} we must have
\[
  AX_{\mathrm{ft},\perp} = 0.
\]
Moreover, if \((B^{\ast},A^{\ast})\) is a global minimizer of the original problem, then
\[
  (B^{\ast},\,A^{\ast} X_{\mathrm{ft}})
\]
is a global minimizer of the objective in \eqref{appeqn:obj_relax}.  Conversely, let \(\,(B^{\ast},\tilde A^{\ast})\) be any global minimizer of \eqref{appeqn:obj_relax}, then one can solve the following equation to obtain $(B^{\ast},A^{\ast})$ which is a global minimizer of the original objective in \eqref{eqn:ft_obj}.
\begin{align}
    A^{\ast} X_{\mathrm{ft}}=\tilde A^{\ast}, \quad A^{\ast} X_{\mathrm{ft},\perp}=0\,.
\end{align}

\end{proof}

\subsection{Proof of Lemma~\ref{applem:unconstraint_constraint_nonconvex}}\label{app_proof:unconstraint_constraint_nonconvex}
\begin{proof}
    To show that there exists a unique value $\rho_{\lambda}$ such that the \eqref{appeqn:problem_one} and \eqref{appeqn:problem_two} enjoys the same set of solutions,
    we first introduce two additional optimization problems.
    \begin{align}
        &\min_{Z}\; 
    \frac{1}{\bar{K}n}\sum_{c=1}^{\bar{K}} \sum_{j=1}^n \mathcal{L}_{\mathrm{CE}}\bigl(y_c,\,Z e_{(c-1)n+j}\bigr)
    + \lambda \|Z\|_*\,, \tag{Problem three} \label{appeqn:problem_three}\\
    &\min_{Z}\; 
    \frac{1}{\bar{K}n}\sum_{c=1}^{\bar{K}} \sum_{j=1}^n \mathcal{L}_{\mathrm{CE}}\bigl(y_c,\,Z e_{(c-1)n+j}\bigr) \quad \text{s.t.} \|Z\|_*\leq \rho_{\lambda}\,. \tag{Problem four}\label{appeqn:problem_four}
    \end{align}
    where $Z\in\R^{K\times \bar{K}n}$. Let $S_i(\lambda), i=1,2,3,4$ be the solution sets of the above four optimization problems for the same fixed $\lambda$. 
    We will show that 
    \begin{itemize}
        \item $S_1(\lambda) = S_2(\lambda), S_3(\lambda) = S_4(\lambda)$.
        \item $S_3(\lambda), S_4(\lambda)$ contains only one element for any fixed $\lambda>0$.
        \item $\forall B^{\ast},\tilde A^{\ast}\in S_1(\lambda)$, one has $B^{\ast}\tilde A^{\ast} \in S_3(\lambda)$.
        \item Let $S_3(\lambda) = S_4(\lambda)=\{{Z_{\lambda}^{\ast}}\}$, then there exists $B^{\ast}\tilde A^{\ast}=Z_{\lambda}^{\ast}$ such that $(B^{\ast},\tilde A^{\ast})\in S_1(\lambda)$.
    \end{itemize}

    We first establish that \eqref{appeqn:problem_three} has a unique solution.
    Our analysis is based on the following lemma.
    \begin{lem}[Cross entropy loss with nuclear norm has a unique minimizer]\label{applem:ce_unique_solution}
        For any $\lambda>0$, define
        \begin{align*}
            \phi(Z; \lambda)=\frac{1}{\bar{K}n}\sum_{c=1}^{\bar{K}} \sum_{j=1}^n \mathcal{L}_{\mathrm{CE}}\bigl(y_c,\,Z e_{(c-1)n+j}\bigr) + \lambda \|Z\|_*,
        \end{align*}
        where $y_c$ is a one-hot vector with $c$-th index equals one. 
        Then $\phi(Z;\lambda)$ has a unique solution.
    \end{lem}
    We refer the readers to later sections in Appendix~\ref{appsec:proof_ce_unique_solution} for the proof.

    Based on Lemma~\ref{applem:ce_unique_solution}, \eqref{appeqn:problem_three} has a unique solution, denoted by $Z_{\lambda}^{\ast}$.

    Before relating this solution to the fixed-width factorization in \eqref{appeqn:problem_one}, we verify its rank feasibility. Let $\Pi=\mathrm{diag}(\Pi_1,\dots,\Pi_{\bar K})$, where each $\Pi_c\in\R^{n\times n}$ is a permutation matrix. Permuting columns within a class leaves the mean cross entropy unchanged, and $\|Z\Pi\|_*=\|Z\|_*$ by unitary invariance. Thus $\phi(Z\Pi;\lambda)=\phi(Z;\lambda)$, and uniqueness implies $Z_\lambda^\ast\Pi=Z_\lambda^\ast$ for every such $\Pi$. Consequently, columns within each class coincide, so
    \begin{align*}
        Z_\lambda^\ast=V_\lambda\otimes\mathbf1_n^\top,\qquad
        V_\lambda\in\R^{K\times\bar K},\qquad
        \operatorname{rank}(Z_\lambda^\ast)\leq\bar K\leq r,
    \end{align*}
    where the final inequality follows from Assumption~\ref{asmp:lora_rank}.

    Set $q=\operatorname{rank}(Z_\lambda^\ast)$. If $q>0$, take the compact SVD $Z_\lambda^\ast=U_q\Sigma_q V_q^\top$, with $U_q\in\R^{K\times q}$ and $V_q\in\R^{\bar Kn\times q}$, and define
    \begin{align*}
        B_0&=\bigl[U_q\Sigma_q^{1/2}\ \ \boldsymbol0_{K\times(r-q)}\bigr]\in\R^{K\times r},\\
        \tilde A_0&=\begin{bmatrix}\Sigma_q^{1/2}V_q^\top\\ \boldsymbol0_{(r-q)\times\bar Kn}\end{bmatrix}\in\R^{r\times\bar Kn}.
    \end{align*}
    If $q=0$, take $B_0=0$ and $\tilde A_0=0$ with the same dimensions. In either case,
    \begin{align*}
        B_0\tilde A_0=Z_\lambda^\ast,\qquad
        \|B_0\|_F^2=\|\tilde A_0\|_F^2=\|Z_\lambda^\ast\|_*.
    \end{align*}
    By Lemma~\ref{applem:varitional_nuclear}, the objective of \eqref{appeqn:problem_one} at any $(B,\tilde A)$ is at least $\phi(B\tilde A;\lambda)\geq\phi(Z_\lambda^\ast;\lambda)$, and $(B_0,\tilde A_0)$ attains this lower bound. Hence the minimum values of \eqref{appeqn:problem_one} and \eqref{appeqn:problem_three} agree. Every $(B^\ast,\tilde A^\ast)\in S_1(\lambda)$ must therefore satisfy
    \begin{align*}
        B^\ast\tilde A^\ast=Z_\lambda^\ast,\qquad
        2\|Z_\lambda^\ast\|_*=\|B^\ast\|_F^2+\|\tilde A^\ast\|_F^2.
    \end{align*}
    Equality in $\|B^\ast\tilde A^\ast\|_*\leq\|B^\ast\|_F\|\tilde A^\ast\|_F\leq(\|B^\ast\|_F^2+\|\tilde A^\ast\|_F^2)/2$ also gives $\|B^\ast\|_F^2=\|\tilde A^\ast\|_F^2=\|Z_\lambda^\ast\|_*$.

    Then, one can choose $\rho_{\lambda}=\|Z_{\lambda}^{\ast}\|_*$, and based on strong duality,  \eqref{appeqn:problem_three} and \eqref{appeqn:problem_four} admits the same set of solutions, i.e., $S_3(\lambda)=S_4(\lambda)$.

    Next, we show that $\forall (B^{\ast}, \tilde A^{\ast})\in S_2(\lambda)$, it must satisfy
    \begin{align*}
        B^{\ast}\tilde A^{\ast} \in S_4(\lambda), \qquad 2 \|B^{\ast}\tilde A^{\ast}\|_* = \|B^{\ast}\|_F^2 + \|\tilde A^{\ast}\|_F^2\,.
    \end{align*}
    This is because first, the minimum objective value of \eqref{appeqn:problem_two} and \eqref{appeqn:problem_four} must be equal.
    Given $Z_{\lambda}^{\ast}\in S_4(\lambda)$, one can do balanced factorization, and obtain the corresponding $B^{\ast}, \tilde A^{\ast}$. 
    This implies that the minimum objective value of \eqref{appeqn:problem_two} is less than or equal to that of \eqref{appeqn:problem_four}.
    On the other hand, $\forall (B^{\ast}, \tilde A^{\ast})\in S_2(\lambda)$, one has $\|B^{\ast}\tilde A^{\ast}\|_* \leq \frac{\|B^{\ast}\|_F^2 + \|\tilde A^{\ast}\|_F^2}{2}\leq \rho_{\lambda}$ which implies $B^{\ast}\tilde A^{\ast}$ is a feasible solution of \eqref{appeqn:problem_four}.
    Thus, the minimum objective value of \eqref{appeqn:problem_four} is less than or equal to that of \eqref{appeqn:problem_two}.
    Combine these together, we conclude that the minimum objective value of \eqref{appeqn:problem_two} and \eqref{appeqn:problem_four} must be equal, and $\forall (B^{\ast}, \tilde A^{\ast})\in S_2(\lambda)$, $B^{\ast}\tilde A^{\ast} \in S_4(\lambda)$ must hold. Since $S_4(\lambda)=\{Z_\lambda^\ast\}$, the preceding norm inequalities are equalities, and $\|B^\ast\|_F^2=\|\tilde A^\ast\|_F^2=\rho_\lambda$.

    Finally, we show $S_1(\lambda)=S_2(\lambda)$.

    This is because, on one hand, $\forall B_1^{\ast}, \tilde A_1^{\ast} \in S_1(\lambda)$, from previous reasoning, they must satisfy
    \begin{align*}
        B_1^{\ast}\tilde A_1^{\ast} \in S_3(\lambda)=S_4(\lambda), \qquad \|B_1^{\ast}\|_F^2=\|\tilde A_1^{\ast}\|_F^2=\rho_{\lambda}\,,
    \end{align*}
    and this indicates $B_1^{\ast}, \tilde A_1^{\ast} \in S_2(\lambda)$.

    On the other hand, $\forall B_2^{\ast}, \tilde A_2^{\ast} \in S_2(\lambda)$, they must satisfy
    \begin{align*}
        B_2^{\ast}\tilde A_2^{\ast} \in S_3(\lambda)=S_4(\lambda), \qquad \|B_2^{\ast}\|_F^2=\|\tilde A_2^{\ast}\|_F^2=\rho_{\lambda}\,,
    \end{align*}
    and this indicates $(B_2^{\ast}, \tilde A_2^{\ast}) \in S_1(\lambda)$.
    Thus, we finish the proof for $S_1(\lambda)=S_2(\lambda)$.

    The last step is to show $\rho_{\lambda}$ is a continuous and non-increasing function of $\lambda$ that satisfies
    \begin{align*}
        \lim_{\rho\to0+}\rho_{\lambda}=\infty, \qquad \lim_{\rho\to\infty}\rho_{\lambda}=0\,.
    \end{align*}
    \myparagraph{Continuity of $\rho_{\lambda}$} In this part, we will use Lemma~\ref{applem:berge} to prove the solution to \eqref{appeqn:problem_three} is continuous, which implies that $\rho_{\lambda}$ is a continuous function w.r.t. $\lambda$. 
    Notice Lemma~\ref{applem:berge} is presented for maximization problem, one can replace the objective $f$ to $-f$ to extend it to minimization problem. 
    Moreover, it is obvious that the objective function is continuous, convex, and the set of minimizers is single-valued for any fixed $\lambda>0$.
    Thus, in order to apply Lemma~\ref{applem:berge} to show the continuity of $\rho_{\lambda}$, we only need to check $C(\lambda)$, which is the range of $Z$, is a compact-valued correspondence.
    At first glance, this seems wrong since $Z$ can take any values in $\R^{K\times \bar K n}$ in \eqref{appeqn:problem_three}. However, we will show that one can constrain the domain of $Z$ which leads to the same minimizer.
    Notice that when $Z\!=\!\boldsymbol 0$, the objective in \eqref{appeqn:problem_three} takes the value $\log K$. Thus, we can choose $C(\lambda)\!=\!\left\{Z\|\|Z\|_*\!\leq\!\frac{\log K}{\lambda}\right\}$, which is a compact set. Therefore, for any fix $\lambda$, one can apply Lemma~\ref{applem:berge} to show $Z_{\lambda}^{\ast}$ is continuous w.r.t. $\lambda$, which implies $\rho_{\lambda}\!:=\!\|Z_{\lambda}^{\ast}\|_*$ is also continuous w.r.t. $\lambda$.
     
    \myparagraph{$\rho_{\lambda}$ is a non-increasing function} For any $0<\lambda_1<\lambda_2$, let $Z_{\lambda_1}, Z_{\lambda_2}$ be the solutions for \eqref{appeqn:problem_three} with the corresponding regularization penalty.
    Then, our first observation is
    \begin{align*}
        &\phi(Z_{\lambda_2};\lambda_2) \leq \phi(Z_{\lambda_1};\lambda_2) \\
        \iff& \phi(Z_{\lambda_2};\lambda_1) + (\lambda_2-\lambda_1)\|Z_{\lambda_2}\|_* \leq \phi(Z_{\lambda_1};\lambda_1)+ (\lambda_2-\lambda_1)\|Z_{\lambda_1}\|_* \\
        \iff & \phi(Z_{\lambda_2};\lambda_1)- \phi(Z_{\lambda_1};\lambda_1)\leq(\lambda_2-\lambda_1)(\|Z_{\lambda_1}\|_*-\|Z_{\lambda_2}\|_*) \\
        \iff & \phi(Z_{\lambda_2};\lambda_1)- \phi(Z_{\lambda_1};\lambda_1)\leq(\lambda_2-\lambda_1)(\rho(\lambda_1)-\rho(\lambda_2))\,.
    \end{align*}
    Notice $Z_{\lambda_1}\in\argmin \phi(Z;\lambda_1)$, thus, we have $\phi(Z_{\lambda_2};\lambda_1)- \phi(Z_{\lambda_1};\lambda_1)\geq0$, and it leads to $\rho(\lambda_1)\geq\rho(\lambda_2)$.
    Therefore, $\rho_{\lambda}$ is non-increasing in $\lambda$.
    If $\rho(\lambda_1)=\rho(\lambda_2)$, the preceding comparison and uniqueness imply $Z_{\lambda_1}=Z_{\lambda_2}=:Z$.
    Write $\ell(Z):=\phi(Z;0)$. By first-order optimality, choose $G_i\in\partial\|Z\|_*$ such that
    \begin{align*}
        \nabla\ell(Z)+\lambda_iG_i=0,\qquad i=1,2.
    \end{align*}
    Subtracting these optimality conditions gives
    \begin{align}\label{appeqn:subg_z}
        \lambda_1G_1=\lambda_2G_2.
    \end{align}
    Every nuclear-norm subgradient satisfies $\langle G_i,Z\rangle=\|Z\|_*$.
    Taking Frobenius inner products of \eqref{appeqn:subg_z} with $Z$ therefore yields
    \begin{align*}
        (\lambda_1-\lambda_2)\|Z\|_*=0.
    \end{align*}
    Since $0<\lambda_1<\lambda_2$, this forces $\|Z\|_*=0$, hence $Z=0$.
    Thus, when $\rho(\lambda_1)=\rho(\lambda_2)$ holds, one must have $Z_{\lambda_1}=Z_{\lambda_2}=0$.

    \myparagraph{Asymptotic behaviour of $\rho_{\lambda}$}
    In this part, we aim to show
    \begin{align*}
        \lim_{\lambda\to0}\rho_{\lambda}=\infty, \lim_{\lambda\to\infty}\rho_{\lambda}=0\,.
    \end{align*}
    We first use proof by contradiction to show that $\lim_{\lambda\to\infty}\rho_{\lambda}=0$. 
    Assume there exists $M>0$ such that $\forall \lambda>0$, one has $\rho_{\lambda} \geq M$.
    Then, we take a series $\lambda_i$ such that $\lim_{i\to\infty}\lambda_i=\infty$, and let $S=\{\rho(\lambda_i)\}_{i=1}^{\infty}$.
    Based on the assumption, we know all the elements in $S$ has a lower bound $M$. 
    Moreover, we know $\forall \lambda_i$, $\rho(\lambda_i)\leq\frac{\log K}{\lambda_i}$.
    This is because $\phi(0;\lambda)=\log K$ which is independent of the choices of $\lambda_i$.
    Thus, there exists $M'$ such that $\forall \rho_{\lambda} \in S$, we have $\rho_{\lambda} \leq M'$.
    Based on Bolzano–Weierstrass theorem, there exists a subsequence $\lambda_{i_k}$ such that $\lim_{k\to\infty}\rho(\lambda_{i_k})=M^{\ast}$ where $M\leq M^{\ast}\leq M'$.
    However, in this case, $\lim_{k\to\infty}\phi(Z_{\lambda_{i_k}}^{\ast};\lambda_{i_k}) \geq \lim_{k\to\infty}\lambda_{i_k} \rho(\lambda_{i_k})=\infty$.
    This is in contradictory to the assumption that $Z_{\lambda_{i_k}}^{\ast}$ minimizes $\phi(Z;\lambda_{i_k})$ and $\phi(0;\lambda)=\log K$.
    Thus, one must have $\lim_{\lambda\to\infty}\rho_{\lambda}=0$. 

    Next, we first use proof by contradiction to show that $\lim_{\lambda\to0}\rho_{\lambda}=\infty$. 
    Assume there exists $N>0$ such that $\forall \lambda>0$, one has $\rho_{\lambda} \leq N$.
    For every finite $R>0$, $\phi(Z;0)$ is continuous and strictly positive on the compact nuclear-norm ball $\{Z:\|Z\|_*\leq R\}$. Consequently,
    \begin{align*}
        \epsilon_R:=\min_{\|Z\|_*\leq R}\phi(Z;0)>0\,.
    \end{align*}
    Then, we pick a series $\lambda_k=\frac{1}{k^2}, R_k=k$, and choose 
    \begin{align*}
        \tilde Z_k=k  \begin{pmatrix}
        \boldsymbol 1_n^\top& \boldsymbol 0_n^\top & \boldsymbol 0_n^\top &\cdots, &\boldsymbol 0_n^\top \\
        \boldsymbol 0_n^\top & \boldsymbol 1_n^\top & \boldsymbol 0_n^\top&\cdots & \boldsymbol 0_n^\top\\
        \boldsymbol 0_n^\top  & \boldsymbol 0_n^\top&\cdots & \boldsymbol 0_n^\top & \boldsymbol 1_n^\top\\
        &&\boldsymbol 0_{(K-\bar{K})\times \bar{K}n}&&
    \end{pmatrix}
    \end{align*}
    Based on the optimality of $Z_{\lambda_k}^{\ast}$, one must have
    \begin{align*}
        \phi(Z_{\lambda_k}^{\ast}; \lambda_k) \leq \phi(\tilde Z_k; \lambda_k) = \log\bigl(1+(K-1)\exp(-k)\bigr) + \frac{\bar{K}\sqrt{n}}{k}\,.
    \end{align*}
    On the other hand, 
    \begin{align*}
        \phi(Z_{\lambda_k}^{\ast}; \lambda_k) \geq \phi(Z_{\lambda_k}^{\ast}; 0) \geq \epsilon_N > 0\,.
    \end{align*}
    Combine these two inequalities together, one has
    \begin{align*}
        \epsilon_N \leq \log\bigl(1+(K-1)\exp(-k)\bigr) + \frac{\bar{K}\sqrt{n}}{k}\,.
    \end{align*}
    However, one can choose $k$ sufficiently large that the above inequality breaks. 
    Thus, there cannot exist a $N>0$ such that $\rho_{\lambda} \leq N$ holds $\forall \lambda>0$.
    Therefore, one must have $\lim_{\lambda\to0}\rho_{\lambda}=\infty$.
\end{proof}

\subsection{Proof of Lemma~\ref{applem:ce_unique_solution}}\label{appsec:proof_ce_unique_solution}
\begin{proof}
    Our starting point is the following lemma which is developed in \cite{hong2023neural}.
    \begin{lem}[Cross entropy loss is locally strongly convex in restricted direction]\label{applem:ce_strong_convex}
        Define
        \begin{align*}
            \phi(Z)=\frac{1}{\bar{K}n}\sum_{c=1}^{\bar{K}} \sum_{j=1}^n \mathcal{L}_{\mathrm{CE}}\bigl(y_c,\,Z e_{(c-1)n+j}\bigr)\,,
        \end{align*}
        where $y_c$ is a one-hot vector with $c$-th entry equal to one. For any $R>0$, there exists $\mu(R)>0$ such that on the compact set $\{Z\in\R^{K\times\bar{K}n}:\|Z\|_F\le R\}$, $\phi(Z)$ is $\mu(R)$-strongly convex in any direction $\Delta_Z\in\{\Delta_Z: \mathbf 1_K^\top \Delta_Z=0\}$.
    \end{lem}
    This bounded-set statement follows from the pointwise restricted Hessian bound in the proof of Lemma 5.1 in \cite{hong2023neural}.

    Fix $\lambda>0$ and write $\phi(Z;\lambda):=\phi(Z)+\lambda\|Z\|_*$. Since cross entropy is nonnegative,
    \begin{align*}
        \phi(Z;\lambda)\geq\lambda\|Z\|_*\geq\lambda\|Z\|_F.
    \end{align*}
    Hence $\phi(\cdot;\lambda)$ is continuous and coercive, so it attains a global minimum. Every global minimizer $Z^\ast$ satisfies
    \begin{align*}
        \lambda\|Z^\ast\|_F\leq\phi(Z^\ast;\lambda)\leq\phi(0;\lambda)=\log K,
    \end{align*}
    and therefore $\|Z^\ast\|_F\leq\log K/\lambda$.

    We next show that every global minimizer is centered. Let $Z^\ast$ be any global minimizer and decompose it as
    \begin{align*}
        Z^{\ast}=\frac{1}{K}1_K 1_K^\top Z^{\ast}+(I_K-\frac{1}{K}1_K 1_K^\top) Z^{\ast}\,.
    \end{align*}
    For simplicity, let $P=I_K-\frac{1}{K}1_K 1_K^\top$ be the projection matrix onto the space orthogonal to $1_K$.
    The removed component $\frac{1}{K}1_K1_K^\top Z^\ast$ has constant entries within each column. Since cross entropy is invariant under such logit shifts, $\phi(Z^\ast)=\phi(PZ^\ast)$.
    By the contraction property in Lemma~\ref{applem:dual_nuclear} applied to the orthogonal projection $P$,
    \begin{align*}
        \|PZ^{\ast}\|_* \le \|Z^{\ast}\|_*\,,
    \end{align*}
    with equality if and only if $PZ^{\ast}=Z^{\ast}$. Combined with $\phi(Z^{\ast})=\phi(PZ^{\ast})$, we have $\phi(PZ^{\ast};\lambda)\le\phi(Z^{\ast};\lambda)$, and optimality of $Z^{\ast}$ forces equality, hence $PZ^{\ast}=Z^{\ast}$ and
    \begin{align*}\label{appeqn:optimal_z}
        \mathbf 1_K^\top Z^{\ast}=0\,.
    \end{align*}
    Finally, choose $R>\log K/\lambda$. All global minimizers belong to the convex set
    \begin{align*}
        \mathcal B_R:=\{Z\in\R^{K\times\bar Kn}:\mathbf1_K^\top Z=0,\ \|Z\|_F\leq R\}.
    \end{align*}
    By Lemma~\ref{applem:ce_strong_convex}, $\phi(Z)$ is $\mu(R)$-strongly convex on $\mathcal B_R$. Since the nuclear norm is convex, the regularized objective $\phi(Z;\lambda)$ is also $\mu(R)$-strongly convex on this set. Thus, for any two global minimizers $Z_1,Z_2$,
    \begin{align*}
        \phi\!\left(\frac{Z_1+Z_2}{2};\lambda\right)
        \leq\frac{\phi(Z_1;\lambda)+\phi(Z_2;\lambda)}{2}
        -\frac{\mu(R)}{8}\|Z_1-Z_2\|_F^2.
    \end{align*}
    If $Z_1\ne Z_2$, their midpoint would have a strictly smaller objective value, contradicting global optimality. Therefore the global minimizer is unique.
\end{proof}

\section{Experiment}\label{appsec:experiment}
In this section, we present the detailed setup of the experiments in \S\ref{sec:experiment}.

\myparagraph{Code availability}
Code is available at \url{https://github.com/simonxu97/MaLoRA}.

\myparagraph{Hardware}
All experiments are run on a single NVIDIA RTXA5000 GPU (24GB) per
(benchmark, seed), allocated through an academic SLURM cluster. The
backbones---ViT-Base/16 pre-trained on ImageNet-21k and XLM-RoBERTa
Base---are loaded directly from public checkpoints; we do not pretrain
any model from scratch. Per-domain LoRA finetuning runs for $20$ fixed
epochs with AdamW and takes 5--30 minutes per domain depending on the
dataset, totalling roughly 0.5--3GPU-hours per (benchmark, seed) across
all per-domain adapters and the pretrain-task adapter.
Computational overhead and merge-selection runtime are reported in
Appendix~\ref{appsec:computational_overhead}, using a dedicated matched-budget timing protocol.

\subsection{Details for experiments in \S\ref{sec:margin} and \S\ref{sec:adapter_merging}}

\myparagraph{Definition of superclasses of CIFAR-100} CIFAR-100 groups its $100$ fine categories into
\textbf{20 coarse \emph{superclasses}} as follows

\begin{table}[ht]
\centering
\caption{CIFAR-100 coarse superclasses and their five fine labels.}
\setlength{\tabcolsep}{6pt}
\renewcommand{\arraystretch}{1.15}
\begin{tabular}{@{}p{0.32\linewidth}|p{0.64\linewidth}@{}}
\toprule
\textbf{Superclass} & \textbf{Fine classes (5 per superclass)} \\ \midrule
aquatic mammals                    & beaver, dolphin, otter, seal, whale \\
fish                               & aquarium fish, flatfish, ray, shark, trout \\
flowers                            & orchid, poppy, rose, sunflower, tulip \\
food containers                    & bottle, bowl, can, cup, plate \\
fruit \& vegetables                & apple, mushroom, orange, pear, sweet pepper \\
household electrical devices       & clock, computer keyboard, lamp, telephone, television \\
household furniture                & bed, chair, couch, table, wardrobe \\
insects                            & bee, beetle, butterfly, caterpillar, cockroach \\
large carnivores                   & bear, leopard, lion, tiger, wolf \\
large man-made outdoor things      & bridge, castle, house, road, skyscraper \\
large natural outdoor scenes       & cloud, forest, mountain, plain, sea \\
large omnivores \& herbivores      & camel, cattle, chimpanzee, elephant, kangaroo \\
medium-sized mammals               & fox, porcupine, possum, raccoon, skunk \\
non-insect invertebrates           & crab, lobster, snail, spider, worm \\
people                             & baby, boy, girl, man, woman \\
reptiles                           & crocodile, dinosaur, lizard, snake, turtle \\
small mammals                      & hamster, mouse, rabbit, shrew, squirrel \\
trees                              & maple, oak, palm, pine, willow \\
vehicles 1                         & bicycle, bus, motorcycle, pickup truck, train \\
vehicles 2                         & lawn-mower, rocket, streetcar, tank, tractor \\ \bottomrule
\end{tabular}
\end{table}

\myparagraph{Construction of pre-training and fine-tuning tasks} For every superclass $S$, let $[c_1^S, \ldots, c_5^S]$ be a fixed random permutation of its five fine labels.
We build three disjoint labelled sets
\[
  \mathcal{D}_{\mathrm{pre}}    \;=\;\bigcup_{S}  \{c_1^{S},c_2^{S},c_3^{S}\},\quad
  \mathcal{D}_{1}     \;=\;\bigcup_{S}  \{c_4^{S}\},\quad
  \mathcal{D}_{2}    \;=\;\bigcup_{S}  \{c_5^{S}\}.
\]
Hence $\mathcal{D}_{\mathrm{pre}}$ contains $60$ fine classes (three per super-class) while
each fine-tuning task $\mathcal{D}_i$ contains exactly one \emph{new} class per
superclass, preserving maximal diversity yet zero overlap with $\mathcal{D}_{\mathrm{pre}}$.

\paragraph{Frozen feature extractors.}
We evaluate four widely used backbones, noting their different pre-training dataset:

\begin{itemize}[leftmargin=*]
\item \textbf{ResNet-50} \citep{He_2016_CVPR}  
      (torchvision, \textit{supervised ImageNet-1K}).
\item \textbf{ViT-Base/16} \citep{dosovitskiy2021vit}  
      (timm: supervised pre-training on \textit{ImageNet-21K}, then
      fine-tuned on ImageNet-1K).
\item \textbf{ConvNeXt-Tiny} \citep{liu2022convnet} pre-trained on ImageNet-12K, then fine-tuned on ImageNet-1K.
\item \textbf{CLIP ViT-B/32} \citep{radford2021learning}  
      (OpenAI’s contrastive pre-training on \emph{web-scale image–text}
      pairs; no ImageNet supervision).
\end{itemize}

All models are kept frozen; we extract the CLS token (ViT/CLIP) or the
global-average-pooled penultimate tensor (CNNs) at 224\,$\times$\,224
resolution for every CIFAR-100 image.

\paragraph{Pre-training stage.}
A linear classifier
$W_{\mathrm{pre}}\in\R^{20\times d}$ ($d$ = embedding dimension)
is trained from scratch on $\mathcal{D}_{\mathrm{pre}}$
for $600$ epochs with Adam ($\eta=0.2$).  
The $20$ rows correspond to the super-classes, not the $60$ fine labels; this
matches our theoretical model where each task’s labels share a common output.
\subsubsection{Empirical Motivation for the Task Geometry Assumption}\label{appsec:validation_asmp_orthogonal_data}
\begin{figure}[t]
    \centering
    \includegraphics[width=\textwidth]{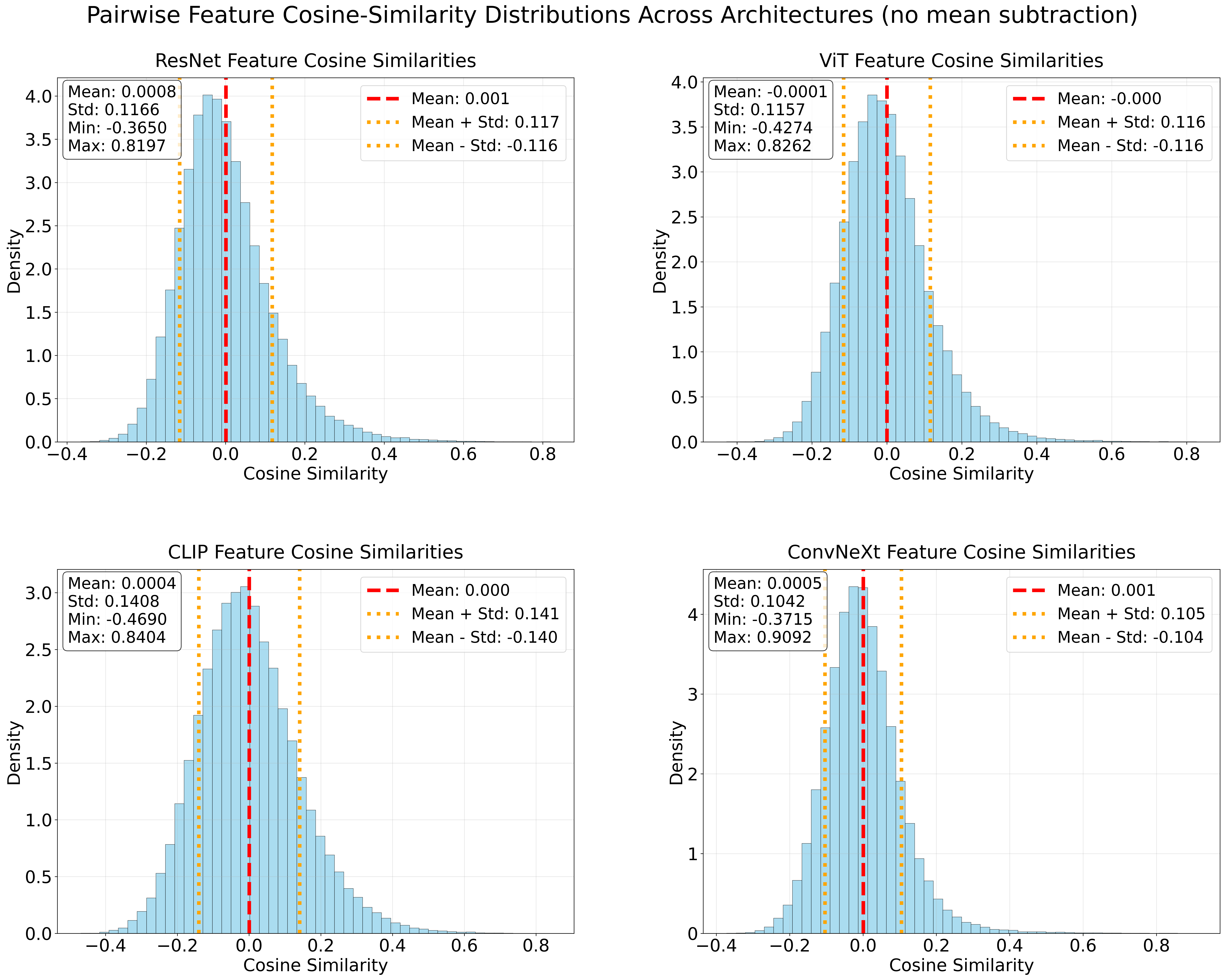}
    \caption{Distributions of pairwise feature correlation across four pre-trained models. The red dashed line represent the mean value of correlations, and the dashed lines represent mean plus and minus one std.}
    \label{fig:feature_correlation_histogram}
\end{figure}
In this section, we conduct a comprehensive feature correlation analysis on CIFAR-100.
This analysis quantifies both intra-class and inter-class correlations in the feature space to provide empirical motivation for the near-orthogonal task geometry in Assumption~\ref{asmp:data}.

Let $\mathcal{F}_{\theta}(x)$ denote the feature extraction function of a neural network with parameters $\theta$, where $x \in \mathbb{R}^{3 \times 32 \times 32}$ represents an input image from CIFAR-100. For each architecture (ResNet-50, ViT-Base/16, CLIP ViT-B/32, and ConvNeXt-Tiny), we extract features from the penultimate layer, obtaining feature vectors $f_i = \mathcal{F}_{\theta}(x_i) \in \mathbb{R}^d$ for each training sample $x_i$.

We organize the extracted features by class, creating sets $\mathcal{S}_c = \{f_i : y_i = c\}$ for each class $c \in \{0, 1, \ldots, 99\}$, where $y_i$ is the ground truth label of sample $x_i$.

\myparagraph{Cosine-Similarity Computation} For any two feature vectors $f_i, f_j \in \mathbb{R}^d$, we compute the cosine similarity:
\begin{equation}
s(f_i, f_j) = \frac{\langle f_i, f_j \rangle}{\|f_i\|_2 \, \|f_j\|_2} = \frac{\sum_{k=1}^{d} f_i^{(k)} f_j^{(k)}}{\sqrt{\sum_{k=1}^{d} \big(f_i^{(k)}\big)^2}\;\sqrt{\sum_{k=1}^{d} \big(f_j^{(k)}\big)^2}},
\end{equation}
where $f_i^{(k)}$ denotes the $k$-th coordinate of $f_i$. 

\myparagraph{Intra-class Similarity} For each class $c$, we compute the average intra-class cosine similarity:
\begin{equation}
s_{\text{intra}}(c) = \frac{1}{|\mathcal{S}_c|(|\mathcal{S}_c|-1)/2} \sum_{f_i, f_j \in \mathcal{S}_c,\; i < j} s(f_i, f_j).
\end{equation}
This measures how aligned features are within the same class, indicating the consistency of learned representations for semantically similar samples.

\myparagraph{Inter-class Similarity} For any two distinct classes $c_1 \neq c_2$, we compute the average inter-class cosine similarity:
\begin{equation}
s_{\text{inter}}(c_1, c_2) = \frac{1}{|\mathcal{S}_{c_1}| \cdot |\mathcal{S}_{c_2}|} \sum_{f_i \in \mathcal{S}_{c_1}} \sum_{f_j \in \mathcal{S}_{c_2}} s(f_i, f_j).
\end{equation}
This quantifies the alignment between features from different classes, with values closer to zero indicating more orthogonal (i.e., better separated) classes.

\myparagraph{Similarity Matrix Construction} We construct a $100 \times 100$ similarity matrix $\mathbf{S}$ where:
\begin{equation}
S_{ij} = \begin{cases}
s_{\text{intra}}(i) & \text{if } i = j \\
s_{\text{inter}}(i, j) & \text{if } i \neq j.
\end{cases}
\end{equation}

\begin{figure}[t]
    \centering
    \includegraphics[width=\textwidth]{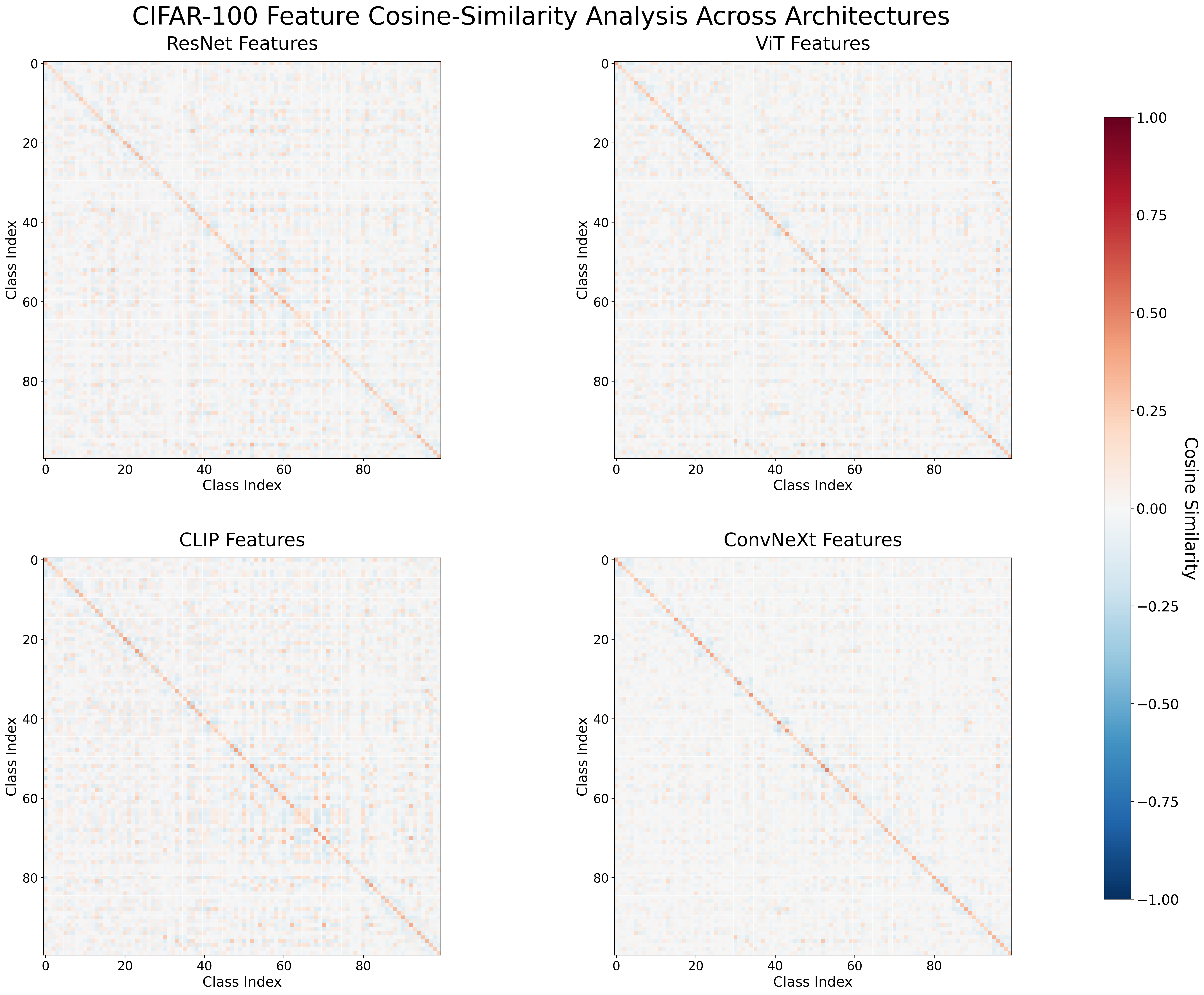}
    \caption{Feature correlation analysis across different pretrained models.}
    \label{fig:feature_correlation_heatmap}
\end{figure}

\myparagraph{Implementation Details}
To ensure computational efficiency while maintaining statistical reliability, we subsample feature pairs for similarity computation. 
Specifically, for each class we randomly sample 100 points (10{,}000 in total). 
We first report the distribution of pairwise cosine similarities across all sampled points in Figure~\ref{fig:feature_correlation_histogram}. 
To further distinguish intra-class and inter-class relationships, we visualize the corresponding similarities as a heatmap in Figure~\ref{fig:feature_correlation_heatmap}. 
For intra-class similarities, we randomly sample up to 1{,}000 distinct sample pairs per class (drawn without replacement to exclude trivial self-pairs whose cosine would be $1$); for inter-class similarities, we sample up to 1{,}000 pairs for each class pair. 
This subsampling strategy yields robust similarity estimates while keeping the computation tractable.

Figure~\ref{fig:feature_correlation_histogram} highlights two key observations: (1) most pairwise cosine similarities are concentrated around zero; and (2) the mean similarity $\pm$ one standard deviation remains below $0.15$ for every architecture.
These descriptive statistics provide qualitative motivation for studying a near-orthogonal reference geometry.
\begin{table}[t]
    \centering
    \begin{tabular}{ccccc}
    \hline
      & CLIP  & RESNET & VIT & CONVNEXT \\
    \hline
    Intra-class mean (std) &  0.2713(0.0769)  & 0.2312(0.0689) & 0.2536(0.0652) & 0.2731(0.0784)\\
    \hline
    Inter-class mean (std) &  -0.0024(0.0670)  & -0.0021(0.0621) & -0.0021(0.0563) & -0.0027(0.0439)\\
    \hline
    \end{tabular}
    \caption{Mean and standard deviation of intra-class and inter-class feature cosine similarities. Values closer to zero indicate greater orthogonality between features.}
    \label{table:correlation_mean_std}
\end{table}
%

Figure~\ref{fig:feature_correlation_heatmap} highlights three observations:
(1) higher diagonal values indicate stronger within-class alignment;
(2) near-zero off-diagonal values indicate weak average inter-class alignment;
and (3) Table~\ref{table:correlation_mean_std} quantifies this contrast, with inter-class means close to zero and intra-class means around $0.23$--$0.27$.
The weak average inter-class and within-class alignment provides qualitative motivation for studying a near-orthogonal reference geometry.
Extending the analysis to richer within-class structure and stronger correlations remains an important direction for future work.
%
\subsubsection{Effect of regularization}\label{sec:margin}
In this section, we numerically verify Theorem~\ref{thm:lora_forget_less}.

\begin{figure}[t]
    \centering
    \includegraphics[width=0.9\textwidth]{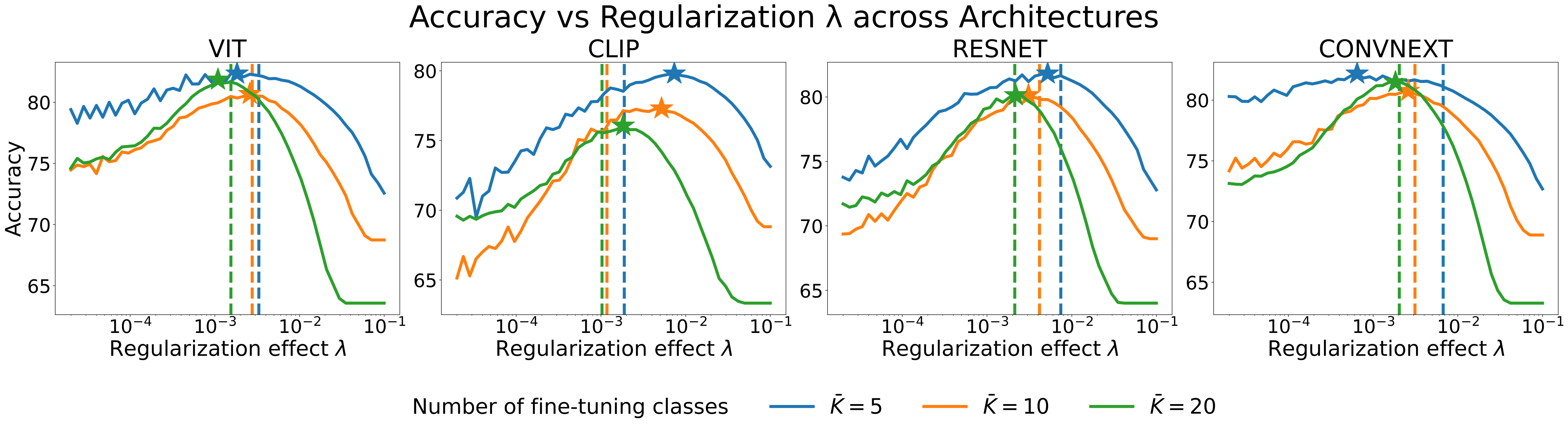}
    \caption{Accuracy of the fine-tuned model across varying regularization levels for four different pre-trained models. Each line represents a fine-tuning task, indicated by color. Stars highlight the best-performing regularization value for each task. Vertical dashed lines indicate the theoretically predicted optimal regularization parameter for each setting. Alignment between the star and the dashed line of the same color reflects how well our theory predicts the empirically optimal regularization level.
}
    \label{fig:reg_accuracy}
\end{figure}

We first pre-train a linear classifier on the pretraining task, and then fine-tune it using LoRA across 50 logarithmically spaced regularization strengths, with $\lambda \in [10^{-5}, 10^{-1}]$.
Figure~\ref{fig:reg_accuracy} shows that LoRA's performance exhibits a non-monotonic relationship with $\lambda$, peaking at a moderate value. 
For ViT and ResNet, the theoretically predicted optimal $\lambda$ is within a factor of $1.8$ of the best empirical choice for every $\bar K$, and for ConvNeXt within a factor of $1.2$ when $\bar K\in\{10,20\}$; for CLIP, the empirical optimum is $1.8$--$4.4$ times larger than predicted.
Detailed comparisons are reported in Appendix~\ref{appsec:experiment}.

For each fine-tuning task, we freeze the pre-trained weights $W_{\mathrm{pre}}$ and attach a rank-$r$ LoRA block, where $r$ is set to the number of classes in the fine-tuning task. 
This satisfies Assumption~\ref{asmp:lora_rank}.
We evaluate performance across three fine-tuning tasks with varying class counts: $\bar{K} = 5, 10, 20$. Each task is constructed by selecting the top $\bar{K}$ superclasses from $\mathcal{D}_2$.

We fine-tune the model on each task using SGD for 3000 steps with a learning rate of 0.5, sweeping over regularization values $\lambda \in [10^{-5}, 10^{-1}]$ with 50 logarithmically spaced regularization strengths. 
For each $\bar{K}$ and each frozen feature extractor, we record both the empirically optimal regularization parameter and the theoretically predicted one from Theorem~\ref{thm:lora_forget_less}. 
The detailed figures are given in Table~\ref{tab:lambda_comparison}.
\begin{table}[ht]
\centering
\caption{Empirical (Emp) and theoretical (Thm) optimal regularization parameters $\lambda$ for different pre-trained models and fine-tuning task sizes ($\bar{K}$).}
\label{tab:lambda_comparison}
\begin{tabular}{l@{\hskip 20pt}c@{\hskip 20pt}c@{\hskip 20pt}c}
\toprule
\textbf{Model}  & $\bar{K} = 5$ & $\bar{K} = 10$ & $\bar{K} = 20$ \\
\midrule
\texttt{ResNet-50} &
\begin{tabular}{@{}c@{}}Emp: $0.005208$\\Thm: $0.007424$\end{tabular} &
\begin{tabular}{@{}c@{}}Emp: $0.003092$\\Thm: $0.004158$\end{tabular} &
\begin{tabular}{@{}c@{}}Emp: $0.002184$\\Thm: $0.002121$\end{tabular} \\
\texttt{ViT-B/16} &
\begin{tabular}{@{}c@{}}Emp: $0.001835$\\Thm: $0.003304$\end{tabular} &
\begin{tabular}{@{}c@{}}Emp: $0.002599$\\Thm: $0.002763$\end{tabular} &
\begin{tabular}{@{}c@{}}Emp: $0.001090$\\Thm: $0.001551$\end{tabular} \\
\texttt{ConvNeXt} &
\begin{tabular}{@{}c@{}}Emp: $0.000647$\\Thm: $0.006709$\end{tabular} &
\begin{tabular}{@{}c@{}}Emp: $0.002599$\\Thm: $0.003114$\end{tabular} &
\begin{tabular}{@{}c@{}}Emp: $0.001835$\\Thm: $0.002039$\end{tabular} \\
\texttt{CLIP} &
\begin{tabular}{@{}c@{}}Emp: $0.007373$\\Thm: $0.001889$\end{tabular} &
\begin{tabular}{@{}c@{}}Emp: $0.005208$\\Thm: $0.001178$\end{tabular} &
\begin{tabular}{@{}c@{}}Emp: $0.001835$\\Thm: $0.001030$\end{tabular} \\
\bottomrule
\end{tabular}
\end{table}

\subsubsection{Details for experiments in \S\ref{sec:adapter_merging}}
For the experiments in \S\ref{sec:adapter_merging}, we fine-tune $\mathcal{D}_1, \mathcal{D}_2$ using different LoRA adapters $(B_i, A_i), i=1,2$ with LoRA rank 20. We fine-tune the LoRA adapters using Adam ($\eta=0.1$) with regularization parameter $\lambda=5\times 10^{-7}$. Let $(B_i^{\ast}, A_i^{\ast}), i=1,2$ be the optimal LoRA adapters we achieve at the end of fine-tuning, we merge them together as follows
\begin{align*}
    W_{\mathrm{LoRA}}^{\lambda}(\alpha_1,\alpha_2)=W_{\mathrm{pre}}+\alpha_1 B_1^{\ast}A_1^{\ast}+\alpha_2 B_2^{\ast}A_2^{\ast}
\end{align*}
To seek for the optimal mixing coefficients $(\alpha_1,\alpha_2)$, we run grid search over a $50\times50$ lattice on $(0,1)^2$, and compare it with the optimal theoretical mixing coefficients presented in Theorem~\ref{thm:lora_merge}. We test different number of classes in the fine-tuning datasets, i.e., $\bar K=5,10, 20$, and reports their results here.
\begin{figure}[ht]
    \centering
    \includegraphics[width=\textwidth]{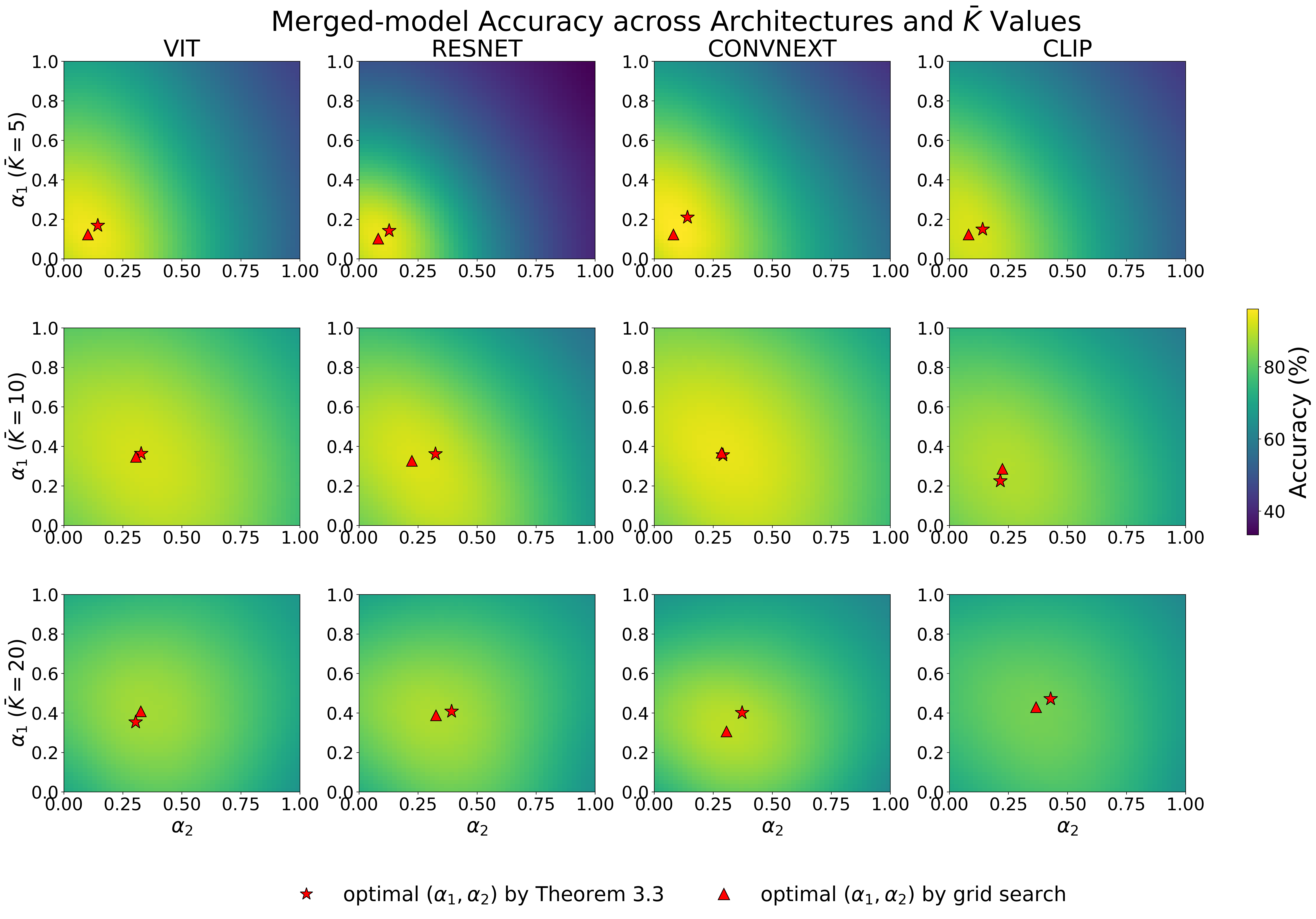}
    \caption{Merged-model accuracy across architectures and different number of classes in the fine-tuning tasks. Each panel shows the accuracy of the merged model evaluated on the combined dataset, across a $50 \times 50$ grid of mixing coefficients $(\alpha_1, \alpha_2) \in (0,1)^2$. For each architecture, the red star indicates the theoretically predicted optimal coefficients from Theorem~\ref{thm:lora_merge}, while the red triangle marks the empirically optimal coefficients.
}
\end{figure}
\subsection{Details for experiments in \S\ref{sec:lora_merging_comparison}}
In \S\ref{sec:lora_merging_comparison}, the datasets tested are all multi-domain benchmarks (See Table~\ref{tab:appendix-datasets} for details).
\begin{table}[H]
\centering
\caption{Average normalized accuracy (\%) following~\cite{ilharco2022editing}: each task's merged-model accuracy is divided by the per-domain finetuned model's accuracy on that task, then averaged across domains within each benchmark. Mean $\pm$ std over 10 seeds. \textbf{Bold} = best per column. Values $>$100 indicate the merged model outperforms the per-task finetuned model on that benchmark.}
\label{tab:norm_acc_with_std}
\footnotesize
\setlength{\tabcolsep}{2.5pt}
\begin{tabular}{lccccccc}
\toprule
Method & PACS & VLCS & Terra Inc. & MNLI & Amazon & CivilComm. & Avg \\
\midrule
Soup        & 96.38 $\pm$ 0.77 & 96.89 $\pm$ 0.47 & 64.31 $\pm$ 2.53 & 101.16 $\pm$ 0.19 & 100.64 $\pm$ 0.54 & \textbf{98.27 $\pm$ 1.66} & 92.94 $\pm$ 0.53 \\
TA          & 97.52 $\pm$ 0.97 & 97.19 $\pm$ 0.96 & 64.46 $\pm$ 2.65 & \textbf{101.27 $\pm$ 0.24} & 100.68 $\pm$ 0.64 & 97.80 $\pm$ 1.69 & 93.15 $\pm$ 0.58 \\
TIES        & 97.52 $\pm$ 0.86 & 96.95 $\pm$ 0.72 & 63.78 $\pm$ 2.36 & 100.87 $\pm$ 0.21 & 100.64 $\pm$ 0.54 & 98.00 $\pm$ 1.98 & 92.96 $\pm$ 0.55 \\
DARE-TIES   & \textbf{97.57 $\pm$ 0.85} & 97.23 $\pm$ 0.96 & 64.23 $\pm$ 2.27 & 100.90 $\pm$ 0.23 & 100.45 $\pm$ 0.65 & 97.92 $\pm$ 1.94 & 93.05 $\pm$ 0.55 \\
DARE-Linear & 96.48 $\pm$ 0.77 & 96.77 $\pm$ 0.51 & 64.73 $\pm$ 2.34 & 101.12 $\pm$ 0.21 & 100.50 $\pm$ 0.57 & 97.61 $\pm$ 1.88 & 92.87 $\pm$ 0.53 \\
\rowcolor{gray!15}
\textbf{MaLoRA} & 96.69 $\pm$ 0.83 & \textbf{97.54 $\pm$ 0.62} & \textbf{65.75 $\pm$ 2.31} & 101.23 $\pm$ 0.22 & \textbf{100.69 $\pm$ 0.56} & 98.00 $\pm$ 1.99 & \textbf{93.32 $\pm$ 0.55} \\
\bottomrule
\end{tabular}
\end{table}
\begin{table}[H]
\centering
\caption{Average relative accuracy drop (\%), the complement of normalized accuracy ($1 - \text{merged}/\text{oracle}$, averaged across domains within each benchmark and across benchmarks). Mean $\pm$ std over 10 seeds. \textbf{Bold} = smallest per column (least forgetting). Negative values indicate the merged model outperforms the per-task finetuned model on average. Avg is the equally weighted mean across benchmarks; its std is $\sqrt{\sum_{b}\sigma_b^2}/6$, where $\sigma_b$ is the per-benchmark std, assuming independent benchmarks.}
\label{tab:rel_drop_with_std}
\footnotesize
\setlength{\tabcolsep}{2.5pt}
\begin{tabular}{lccccccc}
\toprule
Method & PACS & VLCS & Terra Inc. & MNLI & Amazon & CivilComm. & Avg \\
\midrule
Soup        & 3.62 $\pm$ 0.77 & 3.11 $\pm$ 0.47 & 35.69 $\pm$ 2.53 & $-1.16$ $\pm$ 0.19 & $-0.64$ $\pm$ 0.54 & \textbf{1.73 $\pm$ 1.66} & 7.06 $\pm$ 0.53 \\
TA          & 2.48 $\pm$ 0.97 & 2.81 $\pm$ 0.96 & 35.54 $\pm$ 2.65 & \textbf{$-1.27$ $\pm$ 0.24} & $-0.68$ $\pm$ 0.64 & 2.20 $\pm$ 1.69 & 6.85 $\pm$ 0.58 \\
TIES        & 2.48 $\pm$ 0.86 & 3.05 $\pm$ 0.72 & 36.22 $\pm$ 2.36 & $-0.87$ $\pm$ 0.21 & $-0.64$ $\pm$ 0.54 & 2.00 $\pm$ 1.98 & 7.04 $\pm$ 0.55 \\
DARE-TIES   & \textbf{2.43 $\pm$ 0.85} & 2.77 $\pm$ 0.96 & 35.77 $\pm$ 2.27 & $-0.90$ $\pm$ 0.23 & $-0.45$ $\pm$ 0.65 & 2.08 $\pm$ 1.94 & 6.95 $\pm$ 0.55 \\
DARE-Linear & 3.52 $\pm$ 0.77 & 3.23 $\pm$ 0.51 & 35.27 $\pm$ 2.34 & $-1.12$ $\pm$ 0.21 & $-0.50$ $\pm$ 0.57 & 2.39 $\pm$ 1.88 & 7.13 $\pm$ 0.53 \\
\rowcolor{gray!15}
\textbf{MaLoRA} & 3.31 $\pm$ 0.83 & \textbf{2.46 $\pm$ 0.62} & \textbf{34.25 $\pm$ 2.31} & $-1.23$ $\pm$ 0.22 & \textbf{$-0.69$ $\pm$ 0.56} & 2.00 $\pm$ 1.99 & \textbf{6.68 $\pm$ 0.55} \\
\bottomrule
\end{tabular}
\end{table}

\begin{table}[ht]
\centering
\small
\caption{Benchmarks used in the main paper. ``Domains'' lists every domain in
the dataset; the \emph{pretrain} column marks the one folded into
\texttt{base\_with\_pretrain}. ``\#~merged'' is the number of LoRA adapters
that participate in the dense task-arithmetic merge (i.e. all non-pretrain
adapters).}
\label{tab:appendix-datasets}
\setlength{\tabcolsep}{4pt}
\begin{tabularx}{\textwidth}{llcccX}
\toprule
Dataset & Modality & \# domains & \# classes & \# merged & Domains (pretrain in \textbf{bold}) \\
\midrule
PACS               & vision & 4 & 7  & 3 & \textbf{art painting}, cartoon, photo, sketch \\
VLCS               & vision & 4 & 5  & 3 & Caltech101, LabelMe, SUN09, \textbf{VOC2007} \\
Terra Incognita    & vision & 4 & 10 & 3 & location\_38, location\_43, location\_46, \textbf{location\_100} \\
\midrule
MNLI                & language & 5 & 3 & 4 & \textbf{fiction}, government, slate, telephone, travel \\
Amazon Reviews     & language & 4 & 2 & 3 & \textbf{books}, electronics, kitchen, movies \\
CivilComments      & language & 5 & 2 & 4 & identity\_attack, \textbf{insult}, obscene, sexual\_explicit, threat \\
\bottomrule
\end{tabularx}
\end{table}


\paragraph{Validation/test splits.}
For every domain, we hold out a \textbf{stratified $10\%$ validation split} and a \textbf{stratified $10\%$ test split} from the per-domain training set, with the remaining $80\%$ used for fine-tuning. 
The split indices are deterministic in the configured random seed and are reused across all methods so that Soup, Task Arithmetic, and MaLoRA are evaluated on the same per-domain val/test sets within a seed. 
Validation accuracy averaged over all merged domains is the criterion for selecting the merge coefficient $\lambda^\ast$.

\myparagraph{LoRA configuration} We use LoRA~\cite{hu2022lora} with the following layout, applied identically to every adapter (per-domain, per-seed):

\begin{itemize}[leftmargin=*]
\item \textbf{Target modules:} \texttt{query} and \texttt{value} projections
  only (no key, no MLP, no embeddings, no normalization).
\item \textbf{Layers transformed:} the last four transformer layers, indexed
  $\{8, 9, 10, 11\}$ (out of 12). This was identified as the
  best layer placement in our preliminary layer-sweep ablation.
\item \textbf{Scaling factor:} $\alpha_{\text{LoRA}} = 16$ for all adapters. 
\item \textbf{LoRA dropout:} $0.05$.
\item \textbf{Classifier head:} a fresh classification head is initialized for each adapter and trained jointly with the LoRA parameters. At merge time, each merged adapter contributes \emph{both} its classifier head and its LoRA delta to the weighted sum, so that the merged model has a coherent classifier without an external head-aggregation step.
\end{itemize}

\paragraph{Training and per-dataset hyperparameters.}
All adapters are trained with AdamW for $20$ epochs (no early stopping; \texttt{lora\_patience = 0}). 
Before enabling the LoRA branch, we warm up the classifier head for $5$ epochs at learning rate $3 \times 10^{-4}$ (\texttt{head\_epochs = 5}, \texttt{head\_lr = 3e-4}); each domain trains its own head (\texttt{train\_shared\_head = false}). 
We use default PyTorch gradient clipping. 
The four hyperparameters that vary across datasets, i.e., LoRA rank $r$, learning rate, batch size, and weight decay, are summarized in Table~\ref{tab:appendix-hyperparameters}.

\begin{table}[h]
\centering
\small
\caption{Per-dataset training hyperparameters. All adapters use AdamW with LoRA $\alpha\!=\!16$, dropout $0.05$, query and value projections of the final four transformer layers, and $20$ training epochs. The \emph{pretrain domain} is the one folded into the \texttt{base\_with\_pretrain} model, with the remaining domains fine-tuned on top.}
\label{tab:appendix-hyperparameters}
\begin{tabular}{llcccc}
\toprule
Dataset           & Pretrain domain         & LoRA rank $r$  & Learning rate              & Batch size      & Weight decay        \\
\midrule
PACS              & art painting            & $4$           & $5 \times 10^{-4}$         & $32$            & $5 \times 10^{-3}$           \\
VLCS              & VOC2007                 & $4$            & $5 \times 10^{-4}$         & $32$            & $5 \times 10^{-2}$           \\
Terra Incognita   & location\_100           & $4$            & $5 \times 10^{-4}$         & $32$            & $5 \times 10^{-3}$           \\
MNLI              & fiction                 & $8$            & $5 \times 10^{-4}$         & $32$            & $5 \times 10^{-3}$           \\
Amazon Reviews    & books                   & $8$            & $5 \times 10^{-4}$         & $128$            & $5 \times 10^{-3}$           \\
CivilComments     & insult                  & $32$           & $5 \times 10^{-4}$         & $32$            & $5 \times 10^{-3}$           \\
\bottomrule
\end{tabular}
\end{table}

\myparagraph{Baseline methods.} We compare the following methods for merging the LoRA adapters,
\begin{itemize}[leftmargin=*]
    \item \textbf{Uniform Soup} \cite{wortsman2022model}: $\alpha_i = 1/N$ (equal weight across all merged adapters). Equivalent to Task Arithmetic at $\lambda = 1/N$.
    \item \textbf{Task Arithmetic (TA)} \cite{ilharco2022editing}: $\alpha_i = \lambda$ for every $i$, with $\lambda$ tuned on validation.
    \item \textbf{DARE-Linear}~\cite{yu2024language}: each task vector $\Delta W_i$ is randomly sparsified by a Bernoulli mask with keep rate $d$ and rescaled by $1/d$, then linearly averaged with global scale $\lambda$, i.e.\ $\Delta W_{\text{merged}} = (\lambda/N)\sum_i \tilde{\Delta W_i}$ where $\tilde{\Delta W_i}$ is the dropped-and-rescaled task vector. Both $d$ and $\lambda$ are tuned on validation.
    \item \textbf{DARE-TIES}~\cite{yu2024language}: applies the DARE drop-and-rescale step above to each task vector, then aggregates via the TIES sign-election and disjoint-merge procedure with global scale $\lambda$. Both $d$ and $\lambda$ are tuned on validation.
    \item \textbf{TIES}~\cite{yadav2023resolving}: each task vector is first trimmed to its top-$d$ fraction of parameters by magnitude; per parameter, the dominant sign is elected across adapters by total signed magnitude; the merged delta averages only the sign-aligned values and is scaled by $\lambda$. Both $d$ and $\lambda$ are tuned on validation.
\end{itemize}
All baselines that involve density and $\lambda$ are tuned over $d \in \{0.2, 0.3, \ldots, 0.9\}$ and $\lambda \in \{0.1, 0.2, \ldots, 1.0\}$ (80 configurations); Task Arithmetic uses the same $\lambda$ grid (10 configurations).

\subsubsection{Feature correlations across domains}\label{appsec:multidomain_feature_correlations}
We extend the CIFAR-100 analysis in Appendix~\ref{appsec:validation_asmp_orthogonal_data} to the six multi-domain benchmarks in \S\ref{sec:lora_merging_comparison}.
For this analysis, we extract features using the stock pre-trained checkpoints \texttt{google/vit-base-patch16-224-in21k} for PACS, VLCS, and Terra Incognita, and \texttt{xlm-roberta-base} for MNLI, Amazon Reviews, and CivilComments.
The backbones are kept fixed and used for inference only.
Each domain is treated as a task, and the domain counts include both the pre-training domain and all fine-tuning domains.

\begin{table}[htbp]
\centering
\footnotesize
\setlength{\tabcolsep}{4pt}
\caption{Feature correlations within and across classes and domains. Entries report mean correlations, with $\pm$ denoting spread across class--domain cell pairs in each group. A cell is a class--domain combination.}
\label{tab:multidomain_feature_correlations}
\begin{tabular}{lcccccc}
\toprule
Dataset & \# classes & \# domains & \multicolumn{2}{c}{Same class} & \multicolumn{2}{c}{Different classes} \\
\cmidrule(lr){4-5}\cmidrule(lr){6-7}
 & & & Same domain & Different domains & Same domain & Different domains \\
\midrule
PACS & 7 & 4 & $+0.376\!\pm\!0.155$ & $+0.020\!\pm\!0.157$ & $+0.139\!\pm\!0.100$ & $-0.060\!\pm\!0.090$ \\
VLCS & 5 & 4 & $+0.183\!\pm\!0.169$ & $+0.029\!\pm\!0.047$ & $+0.008\!\pm\!0.042$ & $-0.019\!\pm\!0.029$ \\
Terra Incognita & 10 & 4 & $+0.312\!\pm\!0.143$ & $+0.060\!\pm\!0.085$ & $+0.059\!\pm\!0.152$ & $-0.033\!\pm\!0.095$ \\
MNLI & 3 & 5 & $+0.145\!\pm\!0.132$ & $-0.018\!\pm\!0.048$ & $+0.143\!\pm\!0.131$ & $-0.019\!\pm\!0.048$ \\
Amazon Reviews & 2 & 4 & $+0.073\!\pm\!0.024$ & $-0.003\!\pm\!0.018$ & $+0.039\!\pm\!0.023$ & $-0.026\!\pm\!0.020$ \\
CivilComments & 2 & 5 & $+0.053\!\pm\!0.024$ & $+0.018\!\pm\!0.020$ & $+0.044\!\pm\!0.020$ & $+0.011\!\pm\!0.018$ \\
\midrule
\textbf{Mean} & & & $\mathbf{+0.191}$ & $\mathbf{+0.018}$ & $\mathbf{+0.072}$ & $\mathbf{-0.025}$ \\
\bottomrule
\end{tabular}
\end{table}

In Table~\ref{tab:multidomain_feature_correlations}, cross-domain mean correlations range from $-0.060$ to $0.060$, providing qualitative motivation for weak average cross-task alignment.
Within domains, small mean correlations for different-class pairs on VLCS ($0.008$) and for both same-class and different-class pairs on Amazon Reviews and CivilComments ($0.039$--$0.073$) also motivate a near-orthogonal reference geometry.
Stronger same-class alignment on PACS ($0.376$) and Terra Incognita ($0.312$) highlights structured departures from this geometry, showing that agreement with the reference geometry varies across benchmarks and pair categories.

\subsubsection{Computational overhead}\label{appsec:computational_overhead}
We measure end-to-end merge-selection runtime on all six benchmarks over ten random seeds, using one NVIDIA RTX A5000 GPU and a fixed budget of four CPU threads.
For the six methods compared in Table~\ref{tab:merging_runtime}, we use a matched budget of $49$ configurations per method, separately from the accuracy-selection grids described above.
The timings measure merging and coefficient selection from trained adapters; adapter-training costs are reported separately in the hardware description.

MaLoRA estimates margins with one forward pass over each domain's validation split, costing $1.05$--$1.48$ times a single validation pass.
The resulting margins are computed once and reused throughout the global-scale search.
For fixed adapters and validation sets, this additional margin-computation cost is independent of the number of candidate scales.

Table~\ref{tab:merging_runtime} reports the mean runtime for each benchmark and the arithmetic mean across the six benchmarks.
MaLoRA's aggregate mean runtime is $972.6$~s, compared with $950.3$~s for TA, an increase of $22.3$~s or approximately $2.35\%$.
Its aggregate mean runtime is lower than those of DARE-Linear, DARE-TIES, and TIES ($980.2$--$991.0$~s).

\begin{table}[htbp]
\centering
\caption{Mean end-to-end merge-selection runtime (seconds) over ten seeds per benchmark under the matched-budget timing protocol. Avg. is the arithmetic mean across the six benchmarks; the final column divides each method's Avg. by MaLoRA's.}
\label{tab:merging_runtime}
\footnotesize
\setlength{\tabcolsep}{4pt}
\begin{tabular}{lrrrrrrrr}
\toprule
Method & PACS & VLCS & Terra Inc. & MNLI & Amazon & CivilComm. & Avg. & Ratio \\
\midrule
Soup        & 287.6 & 752.2 & 904.1 & 3207.7 & 253.6 & 297.4 & 950.4 & $0.98\times$ \\
TA          & 287.6 & 752.8 & 903.8 & 3206.7 & 253.5 & 297.2 & 950.3 & $0.98\times$ \\
DARE-Linear & 316.4 & 784.1 & 937.1 & 3230.9 & 280.4 & 332.4 & 980.2 & $1.01\times$ \\
DARE-TIES   & 323.2 & 787.3 & 940.8 & 3242.7 & 285.2 & 340.0 & 986.5 & $1.01\times$ \\
TIES        & 324.9 & 790.6 & 939.9 & 3254.4 & 290.0 & 346.2 & 991.0 & $1.02\times$ \\
\rowcolor{gray!15}
\textbf{MaLoRA} & 295.4 & 769.6 & 924.0 & 3277.4 & 261.7 & 307.3 & 972.6 & $1.00\times$ \\
\bottomrule
\end{tabular}
\end{table}


\newpage
\section*{NeurIPS Paper Checklist}

\begin{enumerate}

\item {\bf Claims}
    \item[] Question: Do the main claims made in the abstract and introduction accurately reflect the paper's contributions and scope?
    \item[] Answer: \answerYes{} 
    \item[] Justification: The claim made in the abstract can be found in Theorem~\ref{thm:main}, Theorem~\ref{thm:lora_forget_less}, Theorem~\ref{thm:lora_merge}, and Algorithm~\ref{alg:margin_lora_merge}.

\item {\bf Limitations}
    \item[] Question: Does the paper discuss the limitations of the work performed by the authors?
    \item[] Answer: \answerYes{} 
    \item[] Justification: we discuss the limitations of the work in \S\ref{sec:conclusion}.

\item {\bf Theory assumptions and proofs}
    \item[] Question: For each theoretical result, does the paper provide the full set of assumptions and a complete (and correct) proof?
    \item[] Answer: \answerYes{} 
    \item[] Justification: we list all the assumptions in \S\ref{sec:background}, and provide proofs in Appendix~\ref{app:proof_main_theorem}--Appendix~\ref{appsec:proof_merge}.

    \item {\bf Experimental result reproducibility}
    \item[] Question: Does the paper fully disclose all the information needed to reproduce the main experimental results of the paper to the extent that it affects the main claims and/or conclusions of the paper (regardless of whether the code and data are provided or not)?
    \item[] Answer: \answerYes{} 
    \item[] Justification: we list all the model architectures, datasets, and training setups in Appendix\ref{appsec:experiment}.

\item {\bf Open access to data and code}
    \item[] Question: Does the paper provide open access to the data and code, with sufficient instructions to faithfully reproduce the main experimental results, as described in supplemental material?
    \item[] Answer: \answerYes{} 
    \item[] Justification: all the experiments conducted in this paper use open source datasets, and we provide citations to them.

\item {\bf Experimental setting/details}
    \item[] Question: Does the paper specify all the training and test details (e.g., data splits, hyperparameters, how they were chosen, type of optimizer) necessary to understand the results?
    \item[] Answer: \answerYes{} 
    \item[] Justification: in Appendix~\ref{appsec:experiment}, we provide details on the experiments.

\item {\bf Experiment statistical significance}
    \item[] Question: Does the paper report error bars suitably and correctly defined or other appropriate information about the statistical significance of the experiments?
    \item[] Answer: \answerYes{} 
    \item[] Justification: in Appendix~\ref{appsec:experiment}, we report the mean normalized accuracy and drop with std computed based on 10 repetitions.

\item {\bf Experiments compute resources}
    \item[] Question: For each experiment, does the paper provide sufficient information on the computer resources (type of compute workers, memory, time of execution) needed to reproduce the experiments?
    \item[] Answer: \answerYes{} 
    \item[] Justification: we report the usage of computational resources in Appendix~\ref{appsec:experiment}.
    
\item {\bf Code of ethics}
    \item[] Question: Does the research conducted in the paper conform, in every respect, with the NeurIPS Code of Ethics \url{https://neurips.cc/public/EthicsGuidelines}?
    \item[] Answer: \answerYes{} 
    \item[] Justification: we have reviewed the NeurIPS Code of Ethics and followed it.

\item {\bf Broader impacts}
    \item[] Question: Does the paper discuss both potential positive societal impacts and negative societal impacts of the work performed?
    \item[] Answer: \answerNA{} 
    \item[] Justification: there is no societal impact of the work performed.
    
\item {\bf Safeguards}
    \item[] Question: Does the paper describe safeguards that have been put in place for responsible release of data or models that have a high risk for misuse (e.g., pre-trained language models, image generators, or scraped datasets)?
    \item[] Answer: \answerNA{} 
    \item[] Justification: the paper poses no such risks.

\item {\bf Licenses for existing assets}
    \item[] Question: Are the creators or original owners of assets (e.g., code, data, models), used in the paper, properly credited and are the license and terms of use explicitly mentioned and properly respected?
    \item[] Answer: \answerYes{} 
    \item[] Justification: in \S\ref{sec:experiment}, we cite the original paper that produces the datasets used in the experiment.

\item {\bf New assets}
    \item[] Question: Are new assets introduced in the paper well documented and is the documentation provided alongside the assets?
    \item[] Answer: \answerNA{} 
    \item[] Justification: the paper does not release new assets.

\item {\bf Crowdsourcing and research with human subjects}
    \item[] Question: For crowdsourcing experiments and research with human subjects, does the paper include the full text of instructions given to participants and screenshots, if applicable, as well as details about compensation (if any)? 
    \item[] Answer: \answerNA{} 
    \item[] Justification: the paper does not involve crowdsourcing nor research with human subjects.

\item {\bf Institutional review board (IRB) approvals or equivalent for research with human subjects}
    \item[] Question: Does the paper describe potential risks incurred by study participants, whether such risks were disclosed to the subjects, and whether Institutional Review Board (IRB) approvals (or an equivalent approval/review based on the requirements of your country or institution) were obtained?
    \item[] Answer: \answerNA{} 
    \item[] Justification: the paper does not involve crowdsourcing nor research with human subjects.

\item {\bf Declaration of LLM usage}
    \item[] Question: Does the paper describe the usage of LLMs if it is an important, original, or non-standard component of the core methods in this research? Note that if the LLM is used only for writing, editing, or formatting purposes and does \emph{not} impact the core methodology, scientific rigor, or originality of the research, declaration is not required.
    \item[] Answer: \answerNA{} 
    \item[] Justification: we document how we use LLM in Appendix~\ref{sec:llm}.

\end{enumerate}

\end{document}